\documentclass{article}

\usepackage{amsmath}
\usepackage{amsfonts}
\usepackage{amssymb}
\usepackage{hyperref}
\usepackage{xcolor}
\definecolor{darkblue}{rgb}{0, 0, 0.5}
\hypersetup{colorlinks=true,citecolor=darkblue, linkcolor=darkblue, urlcolor=darkblue}

\usepackage[a4paper, total={6in, 8in}]{geometry}

\newtheorem{theorem}{Theorem}

\newtheorem{axiom}[theorem]{Axiom}
\newtheorem{assumption}[theorem]{Assumption}

\newtheorem{conjecture}[theorem]{Conjecture}
\newtheorem{corollary}[theorem]{Corollary}

\newtheorem{definition}[theorem]{Definition}

\begin{document}

\title{A New Semantic Theory of Natural Language}

\author{Kun Xing  \\   kun.xing.ac@gmail.com}

\date{}

\maketitle

\begin{abstract}
Formal Semantics and Distributional Semantics are two important semantic frameworks in Natural Language Processing (NLP). Cognitive Semantics belongs to the movement of Cognitive Linguistics, which is based on contemporary cognitive science. Each framework could deal with some meaning phenomena, but none of them fulfills all requirements proposed by applications. A unified semantic theory characterizing all important language phenomena has both theoretical and practical significance; however, although many attempts have been made in recent years, no existing theory has achieved this goal yet.

This article introduces a new semantic theory that has the potential to characterize most of the important meaning phenomena of natural language and to fulfill most of the necessary requirements for philosophical analysis and for NLP applications. The theory is based on a unified representation of information, and constructs a kind of mathematical model called cognitive model to interpret natural language expressions in a compositional manner. It accepts the empirical assumption of Cognitive Semantics, and overcomes most shortcomings of Formal Semantics and of Distributional Semantics. The theory, however, is not a simple combination of existing theories, but an extensive generalization of classic logic and Formal Semantics. It inherits nearly all advantages of Formal Semantics, and also provides descriptive contents for objects and events as fine-gram as possible, descriptive contents which represent the results of human cognition. 



\vspace{5pt}


\end{abstract}


\section{Introduction}

Serious discussion of the meaning of natural language could date back to ancient Greece \cite{Plato1997, Aristotle1991}. In fact, a large amount of work in philosophy could be viewed as clarifying the meaning of some important words such as entity, event, truth, knowledge, causality, possibility, morality, beauty. It was Gottlob Frege, however, who had first formulated a mathematical theory on this topic \cite{Frege1960}. Frege's theory had been developed into the logic-based approach of meaning representation, including various logic theories on formal language \cite{Russell2009, Church1956, Kripke1959, Blackburn2006, Gabbay2007-7}, various semantic theories on natural language \cite{sep-meaning}, and Formal Semantics \cite{Montague1974, Partee2008}. 

Frege's theory was incomplete: He had introduced two necessary components of meaning---reference and sense---but only reference had been formally discussed \cite{Frege1960}. This incompleteness had been inherited by all subsequent theories in the logic-based approach, including Formal Semantics. Although it caused no problem in the study of formal language, this incompleteness had introduced many difficulties when interpreting natural language \cite{Boleda2016, Beltagy2016}. The absence of sense makes objects abstract points without any descriptive content, objects which constitute the domain of most semantic models. The lack of descriptive content makes the semantic model hard to distinguish subtle difference of meaning, and restricts the use of various reasoning methods adopted by human practice, such as reasoning by analogy \cite{Boleda2016}.

Distributional Semantics could overcome many of these shortcomings in Formal Semantics \cite{Boleda2016}. It belongs to the vector-based approach of meaning representation to which powerful machine learning techniques could be easily applied. This approach, however, is difficult to capture the meaning of function words, and is therefore hard to characterize the compositionality of natural language or accurate reasoning methods such as deduction \cite{Boleda2016}. In fact, there is a more fundamental problem. Distributional models generally use statistics of surrounding words in corpora to represent meaning. Corpora are collections of language instances produced by human, collections which should be constantly changed as more instances have been produced. The continuous change of corpora would result in a continuous change of word meaning, no matter what statistical methods are used. This consequence obviously contradicts the common sense: The meaning of many words are regarded to be invariant, no matter how people use them. For example, in most contexts, the word `Socrates' is supposed to mean a unique person in the real world, who has never changed after his death. In a distributional model, however, the meaning of `Socrates' would be constantly changed as more sentences with the word are added to the corpus. This almost insuperable problem of Distributional Semantics makes it unsuitable to be an ultimate meaning theory of natural language.  

Lexical Semantics has also been frequently used in NLP practice. It defines or clarifies the meaning of words and phrases by other words and phrases, and constructs dictionaries such as WordNet \cite{Allen1995, Jurafsky2008}. A dictionary often contains various kinds of useful information about a language, summarized from corpora by lexicographers; however, a dictionary could not provide an independent meaning representation for those words or phrases it defines. To understand a dictionary, the user should have already understood the meaning of expressions in its definition entries. This is why a novice or a computer could hardly grasp a language by only looking through a dictionary written in the same language. For this reason, Lexical Semantics could only provide supplementary resources to other semantic representation. 



Cognitive Semantics believes that language is part of human cognitive ability, and could therefore only describe the world as people conceive it; it regards language as an instrument for organizing, processing and conveying information \cite{Geeraerts2007, Evans2006, Croft2004}. This assumption follows the empirical tradition, which is the basis of modern science, and hence provides the approach a solid foundation; however, Cognitive Semantics lacks a unified formal theory, which restricts its application to  NLP practice. 

The new semantic theory presented in this article is grounded in the empirical assumptions adopted by Cognitive Semantics and modern science, but is also constructed to be general and formal. It inherits those key advantages of Formal Semantics and of Distributional Semantics, and overcomes most of their difficulties. The content of this article is divided into three sections.

The principal function of natural language is to exchange information; so the first section focuses on a general, formal and unified representation of information. This task is not as difficult as it appears. We start from a scientific analysis of the procedure through which a piece of information is obtained, then seek out the necessary components to fix a minimal piece of information, and finally represent each minimal piece of information by a sequence of parameter values called a {\bf primitive observation}. Primitive observations are the fundamental meaning elements of the new semantic theory, and the main part of the first section is to clarify those components and parameters from which primitive observations are formed.

The second section constructs a kind of mathematical model called {\bf cognitive model}, under which natural language expressions could be recursively interpreted. Like other semantic models, worlds, objects and events are the most important components of a cognitive model; unlike Formal Semantics, a cognitive model provides a world, an object or an event with descriptive contents, using primitive observations. This work overcomes the major problem of Formal Semantics, and considerably extends the range of language phenomena that could be characterized by a logic-based semantic model.

The last and longest section seeks out the common interpretation of natural language under the framework of cognitive models. The interpretation starts from words and idioms, whose meanings are then composed to form the meaning of phrases and sentences. We will define three kinds of meaning for natural language expressions---denotations, senses and explanations, with the purpose of solving many of those problems or confusions in other semantic theories \cite{sep-meaning}. Sentences are interpreted from simple to complicated types, including a discussion of quantifiers, propositional attitude reports, connectives and modality. Each of those topics is large, so this article only provides some general frameworks for detailed studies in the future.

\section{Observations: A Unified Representation of Information}

The primary and fundamental function of natural language is for communication: to exchange information between the users of this language \cite{Geeraerts2007, Evans2006, Croft2004}. Therefore, the information conveyed by a natural language expression is the major part of what is called the meaning of this expression. Then, to understand the meaning of natural language, the first and most important thing is to understand information. Information theory has provided a quantitative characterization of the amount of information, using entropies of probability distributions \cite{McELIECE2004}; however, information theory is of little help to the construction of a meaning theory. For example, what is the information precisely conveyed by the sentence `Tom was at school yesterday'? To answer such a question, the crucial task is to give information a unified mathematical representation, which is the topic of this whole section. 

In scientific research, people often use data to represent information, such as recordings of scientific apparatus; however, this common practice has several disadvantages, especially to the construction of a meaning theory. Firstly, a different type of information often use a different kind of data with a totally different representation method, which makes them hard to be compared or coherently analyzed. This specialization is not a shortcoming if people focus on a particular area; nonetheless, when constructing a meaning theory, the representation should be unified because every kind of information is talked about by natural language in almost the same way.

The second disadvantage is that the data people commonly use often does not contain enough information for people to completely distinguish a unique piece of information. For example, suppose there is a camera taking photos of an object. Assume that the light, the object, and the position of the camera do not change. Then two photos taken at different time moments by this camera are probably the same: having the same recording in each pixel of the photo. Each photo, however, records a different piece of information, because they are taken in different time moments and hence formed by different strings of light. To make the representation complete so that people could distinguish every piece of information, data often needs to be supplemented by some necessary context information, such as the time moment and camera position of taking the photo. Then, what context information is sufficient in general?  

Thirdly, a piece of commonly used data often contains a large amount of information, not fine enough for theoretical analysis. For example, a photo is a piece of data that could be divided into millions of smaller pieces: the recordings in pixels. To construct a general representation of information, it needs to represent all those minimal pieces of information first, and then other pieces of information are just collections or combinations of them.

The last disadvantage is that some data is more original that other data. For example, some data is obtained from other data through some processing procedure. A data processing procedure could reduce, add or distort the information stored in the original data, and such a change could result in unreal or false information often not easy to detect. Therefore, it is important to distinguish those non-original pieces of data from original pieces. 



To find an information representation method overcoming all those disadvantages, it is better to carefully examine the whole procedure of obtaining a piece of information. From a scientific perspective, any piece of information is obtained through the following {\bf information-obtaining procedure}: 


\begin{enumerate}

\item An {\bf information source} in a world generates and transmits some signal carrying some information; for example, a book reflecting some light carrying some information. The source has its unique state at the moment of generating the information; for example, the book has its unique coordinates, orientation, reflectance at the moment of reflecting light.

\item Then the information is transfered through a {\bf medium}; for example, the light travels through the air, through the glasses of a camera, and finally reached the CMOS or CCD of a digital camera. The medium has its unique state at each moment during the transference.   

\item Then the information is received by an {\bf observer}; for example, the information carried by the light is received by the the CMOS or CCD of a camera. The observer has its unique state at the moment of receiving the information; for example, the camera has its unique coordinates, orientation, resolution power, speed (ISO) at the receiving moment. The information received by an observer at this stage is called directly obtained information, abbreviated as {\bf direct information}. 


\item Then the direct information is transfered, transformed, processed and recorded by the observer; for example, the information carried by the light is transformed into electric signal by the CMOS or CCD of the camera, filtered by some processor, transfered to the memory, recorded as a RAW file, or further processed into and stored as a JPEG file. The recordings in the memory of the observer constitute the {\bf data} one could use, such as these RAW and JPEG files. Old data could be further processed to obtain new data. If the direct information is only transfered, transformed and recorded without any addition or distortion in content (probably with a reduction), then the data is called {\bf original data}, such as the RAW files in a digital camera. Original data could be viewed as carrying direct information. All other data is called {\bf non-original data}, which carries {\bf indirect information}. 

\end{enumerate}


Direct information is directly received from the world outside the observer, with no processing procedure that will distort the information or add new information. In contrast, indirect information is often obtained through such a procedure. The distinction between direct and indirect information is important for both scientific research and everyday life. Scientific apparatus obtains some original data after experiments, and then these data is processed by researchers to get interpretations that often carry some indirect information. The scientific practice generally believes that the original data reveals the truth of the world, but the interpretation of these data might contain some subjective opinions that could be false or less reliable. In a court, the evidence of what people directly see or hear is generally regarded to be more persuasive than evidence obtained after some inference or long-term memory of the observer (The problem would not be so simple if considering whether the evidence is faithfully reported or whether the inference is truth-preserving). From a cognitive perspective, therefore, direct information and indirect information have different truth status. Such a difference should be represented in a meaning theory. 

Direct and indirect information are all information obtained by some observer. Then a natural question is how about those information never obtained by any observer? Even if such information exists, it does not need to be considered by any meaning theory of natural language. A natural language expression could only convey information known by some human being \cite{Geeraerts2007, Evans2006, Croft2004}. If a piece of information could never be obtained by any observer, it could never be obtained by any human being who is also an observer, i.e., no human will know it. If a piece of information is never known by any human, it could never be conveyed by any natural language expression, and hence could never be a part of the meaning of any expression. Therefore, the following discussion only considers information that could be obtained by some observer. Moreover, the representation of direct and indirect information are considered separately, and direct information is discussed first.

\subsection{Representation of Direct Information}

All information considered in this and the next two sections is direct information, so the word `direct' is often omitted there for simplicity. As showed before, an information-obtaining procedure uniquely determines a piece of information received at a particular moment. Such a procedure, however, is extremely complicated; its description requires a large amount of extra and redundant information that usually could not be easily obtained. Therefore, it seems impractical to represent a piece of information by a complete description of the whole procedure that obtains it. Luckily, it is also not necessary to do so: The main task of representing a piece of information is to make it fixed and distinguishable from all other pieces. For this purpose, the representation does not need to cover all details of the obtaining procedure. Instead, it only has to contain some necessary context information that could determine the unique procedure of obtaining the piece of information. Then, what context information is sufficient to determine such a procedure and how to represent the piece of received information? The representation consists of several components. 



The first component is the world where the information source exists, i.e., where the information is from. To use worlds instead of information sources makes it avoid distinguishing between different sources, a task which is far more complicated than just distinguishing between different worlds. Speaking of a world, it does not require any concrete description of this world, but just a label to distinguish it from other worlds. Thus, the first component is in fact a world label. If one wants to make the information source more specific, he could add subworld and sub-subworld labels, although these labels are not necessary. 

The second component is the observer receiving the piece of information. This component should include a description of the state of the observer at the receiving moment, including all necessary state information that could influence the form of the information it receives and recorded. A digital camera taking a photo, for example, has state information such as the shooting moment of this photo, the camera's coordinates, orientation, resolution power, speed (ISO) and some other necessary settings at this shooting moment. 

A world and an observer with one of its states together determine what  source from which the information is generated and what medium through which the information is transfered, and therefore determine a unique information-obtaining procedure during which a piece of information is received by the observer at the moment specified by the observer's state. For example, if we set a camera in the world we live, fix its position, orientation and all other settings, and select a shooting moment, then only one photo will be taken. Moreover, this photo is formed from a unique information-obtaining procedure with a unique medium and a unique information source, although we might not know the detail of the procedure exactly. Therefore, a world and an observer with one of its states are sufficient context information that uniquely fix an information-obtaining procedure, and hence uniquely fix a piece of information. 


Although the state information of an observer often appears complicated, it is generally not hard to obtain. The resolution power of a digital camera is provided by the manufacturer; the speed (ISO) belongs to its inner setting; the shooting moment could be recorded by a clock inside or outside the camera; the coordinates and orientations are easy to be measured, given a fixed measurement system. To simplify the discussion, we assume that the state information could be obtained by the observer whenever it records a piece of information. In other word, an observer is supposed to know  its own states. 

As mentioned before, to make a representation fine enough, it requires to decompose each piece of information into minimal pieces, a piece which could not be divided into smaller pieces; however, whether a piece of information is minimal or not depends on the observer that receives the information. For example, suppose that two black and white cameras take photos in the same world, in the same position and at the same moment; the only difference between these two cameras is their different resolution power: one has $1000 \times 1000$ pixels and the other has $2000 \times 2000$ pixels. Then a photo of the first camera could be divided into one million minimal pieces of information, with respect to one million pixels. In contrast, a photo of the second camera consists of four millions minimal pieces, with respect to four million pixels. When it is a color camera, each piece of information received by a single pixel is not minimal now, because it could be further divided into three or more pieces recording the strength of different colors. 

To uniformly represent all minimal pieces of information, we generalize the concept of a camera's resolution power and include it as an indispensable part of each observer's state. Informally speaking, the {\bf resolution power} of an observer is its maximal power to differentiate its state information or the received information into minimal pieces, and to determine how to turn the information into its recordings. Mathematically speaking, the resolution power of an observer defines a set of parameters, where each parameter is a mathematical variable having values in a domain with some special structure. For example, if a digital camera could record at most three colors in each pixel, then its resolution power defines a parameter having values in domain $\{1,2,3\}$ to show which color is recorded in this pixel. The resolution power of an observer also defines a measurement system for each parameter to map the received information to parameter values. Different observer often receives (recognizes) different kind of information, with different resolution power. The resolution power of an observer could also change through time, however, it is assumed to be fixed at any particular moment. 

The set of parameters defined by a resolution power could be divided into three categories: (1) {\bf state parameters}, which characterize the state of the observer; (2) {\bf resolution parameters}, which characterize how the received information is divided into minimal pieces; (3) the {\bf result parameter}, which characterizes the final result of a minimal piece of the received information. A complete sequence of values of all state parameters is called a {\bf state} of the observer; a complete sequence of values of all resolution parameters is called a {\bf resolution point}; a value of the result parameter is called an {\bf obtaining result}, usually just called a result if there is no confusion. For example, the resolution power of a digital camera defines these three subsets of parameters: state parameters including the shooting time, the space coordinates and orientation, the speed, the focus of lens, etc; resolution parameters including the pixel position, the color, etc; the result parameter has values indicating the strength of light with respect to a particular pixel and a particular color. 


There are three especially important groups of parameters in the resolution power of every observer: time parameters $\boldsymbol{t}$ belonging to state parameters, space parameters $\boldsymbol{s^1}$ also belonging to state parameters, and space parameters $\boldsymbol{s^0}$ belonging to resolution parameters. Their values are called {\bf time moments}, {\bf state space points} and {\bf space points} respectively. Each group could contain more than one parameter, where the number is decided by the dimension of the corresponding time or space. All observers are assumed to have a group of time parameters and two groups of space parameters. When an observer could not distinguish time or space, then the corresponding parameters have some predefined default values.

Resolution point and obtaining result are not included in the world component or the observer component, but they are necessary components in the representation of a minimal piece of information. Therefore, there are totally four necessary components to represent a minimal piece of information, and such a representation is called a {\bf primitive observation}.


\begin{definition}[Primitive Observation]\label{PO}
A primitive observation is a sequence of four components:
\begin{equation}
<world,\ observer,\ resolution\ point,\ obtaining\ result>
\end{equation}
where, as defined before, the $resolution\ point$ is a complete sequence of values $r[0], r[1], ...$ of all resolution parameters, the $obtaining\ result$ is a value $re[0]$ of the result parameter, the $world$ and the $observer$ components will be defined and clarified later. 

\end{definition}


When there is no confusion, a primitive observation is simply called an observation. Sometimes an observer obtains no result in an information-obtaining procedure; for example, a camera takes a photo but there is no light. In this special case, the obtaining result is assumed to be $\varnothing$ or some default value set by the observer. Moreover, as the first step of development, this article does not deal with uncertainty, vagueness or incorrectness of information; so that all values in an observation are assumed to be definite and accurate, and are faithful recordings of the observer. 

Observations are the fundamental building blocks of the new semantic theory. Such a representation of information is obviously unified: Any minimal piece of information that could be obtained by an observer could be represented by a unique observation. Does this representation make any minimal piece of information fixed and distinguishable? Summarized from all above discussions, the following axiom guarantees the representation to be fixed and distinguishable: 

\begin{axiom}[Observation Axiom]
Given a world, an observer with its resolution power and with a state it could have, and given a resolution point, there is one and only one obtaining result. 
\end{axiom}



The observation axiom is the most fundamental assumption in the new semantic theory. It guarantees that the representation of any minimal piece of information is invariant of context: The representation includes all necessary context information to fix this piece of information, and the representation does not change given any additional information. This immutability in representation is crucial for observations to competently serve as the basic meaning elements of a semantic theory, just like objects in classic logics and Formal Semantics. To make the foundation more solid, some further arguments for this axiom is presented here. 


This axiom is supported by our common sense and scientific practice. Imagine a person in a world looking at a neon lamp. The color of the neon lamp is changing through time and space position. If the person is asked to tell what the color of the lamp is, there would be no definite answer; if the person is asked what is the color at a special moment and at a space position on the lamp small enough (to the limit of the person's resolution power), however, he is expected to give a unique answer. If he say, for example, at time 23 o'clock and at position $x$ on the lamp small enough, he saw pure red and pure blue simultaneously, people would think this person is crazy or there is something wrong in his eyes. If the lamp or other phenomenon is observed by a scientific instrument, then the same conclusion would be expected. 

In fact, the observation axiom is logically equivalent to the Law of Non-contradiction, which claims that, for any proposition $p$, $p$ and $\lnot p$ could not be true simultaneously when all necessary context information is given. Suppose that the Law of Non-contradiction is false. Then using the given context information, it is easy to construct observations violating the observation axiom. For example, suppose that $p$ and $\lnot p$ hold simultaneously where $p$ is the proposition expressed by the sentence `This flower is red in the world $W$ at the moment $t$'. Then for any observer who could recognize colors and who is observing the flower at the moment $t$ in world $W$, he must obtain two different results: red and a color different to red. Conversely, suppose that the observation axiom is false, i.e., there are two different results $X$ and $Y$ observed by the same observer $A$, in the same world $W$ and at the same moment $t$. Then the two contradictory propositions expressed by `$A$ observes $X$' and `$A$ do not observes $X$' hold simultaneously in world $W$ at moment $t$, i.e., the Law of Non-contradiction does not hold in this world at this moment.   

Just like the Law of Non-contradiction, which is supposed to hold in all serious discussion but could not prevent people having inconsistent beliefs or making contradictory assertions, the observation axiom could not prevent people constructing imaginary worlds violating this axiom in everyday life. In fact, if it is checked seriously and thoroughly, many contradictions could be found in fiction such as novels, animations and movies, which construct imaginary worlds sometimes violating the observation axiom. Therefore, similar to the Law of Non-contradiction, the observation axiom is just a normative law of human rationality (reasonable thinking), rather than an empirical rule  summarized from human experience.  

\subsection{Worlds and Subworlds}

This section defines and clarifies the first component $world$ in the definition \ref{PO} of observations. 

\begin{assumption}
Each world talked about in natural language has a unique label, and different world has different label; so does each special part of a world.
\end{assumption}

\begin{definition}
The $world$ component in the definition \ref{PO} of an observation is a sequence of labels:
\begin{equation}
<w[0], w[1], w[2], ...>
\end{equation}
where $w[0]$ is the label of the world being observed, $w[1]$ is the label of the subworld being observed, $w[2]$ is the label of the sub-subworld being observed, etc. Only $w[0]$ is a necessary element for this component, other labels are selective. 
\end{definition}


Labels of worlds (sub-worlds, etc) are used to distinguish between what they refer to, and hence could be any symbols or symbol strings. They provide important context information showing where the information source belongs to or where the information comes from. For simplicity, it is common to use natural language words or phrases to be these labels, such as using $the\ real\ world$ to be the label of the world we live. Then a natural question is what worlds (sub-world, etc) could be recognized and distinguished? In other word, what world (sub-world, etc) labels are available or could be introduced? The answer starts with the following assumption: 


\begin{assumption}
There is one and only one world labeled as $the\ real\ world$, which refers to the world where we live and whose information could be obtained by our perceptions.  
\end{assumption}


All worlds different to the real world are called {\bf unreal worlds}. There are many unreal worlds named by natural language words or phrases, which could not be completely listed here; however, some typical and important classes are to be discussed and clarified. 

The first class of unreal worlds are those ones that exist in fictitious stories told by novels, operas, movies, TV shows, animations, games, etc. The title of such a literary work is often used to be the label of the world it describes. There is little controversy that all these worlds are conceived by human imaginations. 

The second class are those ones described by mathematical, scientific or other kind of theories, especially by their definitions and axioms. These worlds are usually called mathematical (scientific, etc) worlds or mathematical models. A typical mathematical world is the standard model of natural numbers. There has been a long-run dispute over the question of whether these worlds exist metaphysically; however, the answer to this question has no relation to the new semantic theory, and hence is not discussed in this article. The fact that human could learn information from these worlds and distinguish them from other worlds implies there is no essential difference between them and other worlds. Another important question is whether these worlds are parts of the real world. The answer would be no because human could never actually perceive these worlds. This conclusion is supported by following arguments. 


In geometries, the axioms define mathematical objects like points without volume, flat planes or curved surfaces whose thickness is zero. In algebra, the axioms define various kinds of numbers. In set theory, the axioms define the empty set and various other sets. All these abstract objects, however, could not be perceived by a real person. No one actually see in the real world a point without volume, the number $10$, or the empty set. In other word, any information known about these mathematical objects is not obtained from the real world. Therefore, mathematical worlds are not parts of the real world. 

The argument for scientific theories and other theories is similar. Mechanics defines objects like mass points (points that have no volume but have mass) and rigid bodies (objects that do not change in shape under any strength); quantum physics defines objects like electron clouds; thermodynamics and chemistry define ideal gas and many other ideal substances; economics defines various ideal function curves like supply curves and demand curves. These objects could not be perceived by any real person either, and hence worlds described by scientific theories could not be parts of the real world. 

Therefore, mathematical theories, scientific theories, and other theories talking about objects that could never be perceived by real persons all describe unreal worlds. These worlds are constructed by people to approximate the real world or its sub-worlds, but they are never a part of the real world. People learn these worlds (obtain information from them) not by their perceptions, but by their imaginations (including their intuitions). 


The third class of unreal worlds are constructed by assumptions,  hypotheses and suppositions, which are often not included in any theory but made by special kinds of statements such as conditionals. For example, a counterfactual conditional generally conceives a world different to the real world. Such a world usually has no special name but is imagined and talked about. A normal conditional also constructs a new unreal world when it talks about the future of the real world. For example, the conditional `if tomorrow is sunny' constructs an unreal world that has the same past of the real world and a sunny tomorrow. Since tomorrow has not been actually perceived at the moment of speaking out this conditional, it could only be imagined. 


Dreams and many other mental activities could also conceive unreal worlds. If thinking about the origin of all unreal worlds from a psychological perspective, however, it could say that these worlds are all created by human imaginations. Dreams, fictions, intuitions, theoretical definitions, assumptions and hypotheses could all be viewed as results of special kinds of imaginations. For this reason, unreal worlds are also called {\bf imaginary worlds} in this article. 

Imagination is the most powerful ability of human being that makes them far more creative than other animals. It is the basis of human intelligence and free will. Imaginary worlds could be very similar to the real world, or totally different. They could be consistent with modern science, or follow different and strange rules. Some of them have no space or time, or their space and time are very different to the real world. Some actually conceived imaginary worlds could even be inconsistent, violating the observation axiom that is supposed to hold in the new semantic theory. (Many worlds with a time travel setting could be found to be inconsistent when all world rules are thoroughly explored and all their consequences are clearly inferred. Nonetheless, implicit inconsistencies usually cause no problem if there is no need to do serious discussion as in maths or science). 

Then we turn to discuss special parts of a world, i.e., its subworlds, sub-subworlds, etc. It is a common practice for people to divide a world into many subworlds (sub-subworlds, etc) by different kinds of observers, different time segments, different space regions, or some other means, since the total amount of information that could be obtained from a world, especially the real world, is often too large to talk about. Labels of subworlds (sub-subworlds, etc) could make the information source more specific. For example, Europe is a subworld of the real world within a particular region of space, the Middle Ages is a sub-world of Europe within a particular segment of time. 

The real world is commonly divided into one outside world (including human bodies and brains) and many minds (human consciousness), where each person is supposed to have one unique mind. There has been a long-time controversy over the question of whether the outside world and human minds are totally different worlds or just different sub-worlds of the real world. The author does not want to join this metaphysical debate; however, it needs to decide whether the labels of the outside world and each person's mind are world labels or subworld labels in observations. The new semantic theory regards them as subworld labels since this choice is more coherent with contemporary science and its future development. Neural science and psychology has been studying the interaction between the outside world and mental activities in a more and more sophisticated way \cite{Kandel2013, Coon2010}. Many medicines that could strongly influence human consciousness has been invented. Many devices that could detect mental activities have also been invented. It would be strange to think that brains and minds belong to totally different worlds or parallel worlds that just happen to appear to have so deeper interactions. Therefore, the outside world and minds are supposed to be different subworlds of the real world. 

The outside world and a person's mind, however, are commonly thought to be essentially different. Then what is the essential difference between them? From the perspective of obtaining information from them, this problem could be reduced to: What is the essential difference between observations from the outside world and from a mind? Reflecting on how a person directly obtains information from the outside world and from his mind (i.e., how a person directly knows the outside world and his mental activities), one could find the answer: A person directly knows the outside world by his {\bf senses}, which could be assisted by some physical devices, but one directly knows his mental activities only by his {\bf self-consciousness}. The self-consciousness of a person could be viewed as an observer that monitors the activities of and obtains information from his mind (One could also learn his mental activities by his memory, but what memory obtains is indirectly information because there is a forgetting procedure). In other words, the information of the outside world and of one's mind are obtained by totally different observers, which results in an essential difference between them. Even if obtaining information from the same information source, different kinds of observers often have different resolution power, receive different kind of information, and generate different obtaining result. For instance, human vision could detect brightness and colors in different space point, but one's self-consciousness could only detect non-spatial mental states or activities such as feelings, desires, beliefs and reasoning. The different resolution power of time and space between human sense (especially vision, audition, kinesthesia and the sense of balance) and one's self-consciousness could be an explanation of why people generally think that the outside world changes in space but one's mind does not. 

The outside world of the real world, also called {\bf the universe} or {\bf the physical world}, is further divided into many sub-subworlds. Different scientific disciplines usually study different subworlds of the universe. Astronomers divide the universe into many galaxy clusters, galaxies, stellar systems, planets, etc; Earth scientists divide the earth into biosphere, lithosphere, hydrosphere, atmosphere, aerosphere, and further divide the lithosphere into many plates; Geographers divide the earth into many continent, many countries, many districts, etc. These subworlds of the universe are divided mainly according to different space regions. In contrast, some subworlds are divided by time segments; for example, geologists divide earth into many geological ages, and historians divides human history into many eras.

Finally, it is worth a mention that there is a clear difference between experimental science and theoretical science. Experimental science uses apparatus to obtain information from the universe, so it is studying subworlds of the universe. In contrast, as discussed before, theoretical science defines and describes objects that could never be perceived in the real world, so it constructs imaginary worlds different to the real world.

\subsection{Observer Components}\label{OSO}

The formal definition of the $observer$ component in observations is as follows. 

\begin{definition}
The $observer$ component of an observation is a sequence of four subcomponents:
\begin{equation}
<o[0], o[1], o[2], o[3]>
\end{equation}
\begin{itemize}

\item $o[0]$ is a sequence of labels that uniquely specifies the observer; $o[0][i]$ denotes the $i$th label in the sequence. 

\item $o[1]$ is the resolution power of the observer, which is a set of triples $o[1][i]=(p_i, d_i, m_i)$ where $p_i$ is a parameter, $d_i$ is the domain of $p_i$, and $m_i$ is a measurement system associated with $p_i$. Each parameter $p_i$ is a mathematical variable having values in $d_i$ and representing the change of a kind of information that could be distinguished and measured by the observer. Each domain $d_i$ is a mathematical structure: a set with some relations or operations on it \cite{Hrbacek1999}. Each measurement system $m_i$ is a function from a kind of received information or signals to $d_i$; \linebreak $m_i$ tells how the observer decides the value of $p_i$ in each observation. 
 
\item $o[2]$ is the state of the observer; $o[2][i]$ denotes the $i$th value in the sequence. 

\item $o[3]$ is either $actual$ or $imaginary$, abbreviated as $ac$ and $im$ respectively. $o[3]$ is called the $ac/im$ label. 



\end{itemize}

\end{definition}

The resolution power $o[1]$ and the state $o[2]$ of the observer have been clarified before, so it only needs to discuss the subcomponents $o[0]$ and $o[3]$. The $o[0]$ subcomponent consists of a sequence of labels to completely specify the observer. For simplicity, we make the following assumption. 

\begin{assumption}
Each observer has a unique observer label-sequence.
\end{assumption}

Why it needs a sequence of labels instead of a single label to refer to an observer? When a person is watching a movie, it is in fact this person's eyes watching the movie, not his ears or other sense organ. Therefore, in this simple example it needs at least two labels to specify the observer: the person's name and `eyes'. Similar to worlds label, it needs to explore what observers could be distinguished and recognized, or in other words, what observer label sequences are available or could be introduced.


The first label $o[0][0]$, in principle, is always supposed to be the name of a real person (or the set of names of a group of persons) who finally obtains the minimal piece of information, because any information that could be discussed in natural language must be learned by some real person. Then what other labels $o[0][1], o[0][2],...$ could be? Firstly, each person has two main collections of abilities to directly obtain information from a world: perception and imagination. 


\begin{definition}
The perception of a person is the collection of his abilities to directly obtain information from the real world. The imagination of a person is the collection of his abilities to conceive imaginary (unreal) worlds and to directly obtain information from them. 
\end{definition}

Perception is discussed first. The real world is divided into the outside world and many minds; however, a person could only directly obtain information from his own mind by his self-consciousness. Thus, the perception of a person is divided into two kinds of abilities: sensation and self-consciousness. 

\begin{definition}
The sensation of a person is the collection of his abilities to directly obtain information from the outside world. The self-consciousness of a person is the collection of his abilities to directly obtain information from his mind. 
\end{definition}



The sensation of a person consists of many sub-abilities that obtain different kinds of information from the outside world: sight (vision), hearing (audition), touch, taste, smell, kinesthetic sense, feelings of temperature, pain, balance, vibration, etc \cite{Coon2010, Kandel2013}. The self-consciousness of a person could detect many kinds of mental activities in his consciousness, for example, one's emotions (feelings), beliefs, knowledge, will, desires, intentions, expectation. Moreover, a person's sensation, imagination, reasoning, memorizing, forgetting and self-consciousness itself are all his mental activities in his consciousness, and hence they could also be detected by his self-consciousness (An interesting phenomenon of one's mind is his self-consciousness of self-consciousness). Finally, according to modern science \cite{Coon2010, Kandel2013}, one's mental activities, including his self-consciousness, are supposed to be functions of his brain. Thus, the state space point of one's self-consciousness is supposed to be within the space region occupied by his brain, which implies that the self-consciousness of different person has different state space point, and hence different state.  



Human sensation could be assisted by various kinds of scientific instruments and man-made sensors, which decide the kind of information being obtained; so that their names are often labels in $o[0]$. Most instruments and sensors are made to detect information that human could not not directly perceive, then converting it into information that human could. Since there is often a common set of rules for each kinds of instruments and sensors to detect and convert information, and to interpret the recordings, any person who knows these rules could understand the recorded results. For this reason, the person's name and his sensation label are often omitted in an observation assisted by some instrument or sensor.

Besides human beings, most other living creatures, especially animals, have abilities to perceive various kinds of information from the real world; however, just like one could not directly know other person's mental activities, the obtaining results of animal perception could not directly known by human. Instead, human learn the results of animal perception from scientific experiments, scientific theories and reasoning, where only scientific experiments obtain direct information. From a scientific perspective, animal perception is like a man-made instruments or sensors whose output is hard to learn and whose rules are left to be explored. In fact, the perception of many animals have not been completely studied or understood. Therefore, if only direct information is considered, it is rare to let animal names and abilities be labels in $o[0]$ when the world label is $the\ real\ world$. 


Human imagination is commonly regarded as the ability to conceive imaginary worlds; however, with some reflection, it is not hard to see that human know (obtain information from) these imaginary worlds at the same time they are conceived. Therefore, imagination is also a kind of observers of human being. In fact, it is the only kind of observers to directly obtain information from imaginary worlds. Dreams, intuitions, guesses, conjectures, theoretical definitions, assumptions, suppositions, hypothesis are all special kinds of human imagination. The problem of whether other animals have the ability of imagination still needs scientific investigations. 

Imaginations include two categories: One is to conceive totally new unreal worlds and obtain information from them, the other is to conceive imagined observers to perceive the real world or some unreal worlds already conceived by others. Such a distinction would influence the values of subcomponents $o[3]$, which will be discussed later. 

Names of imagined observers could be labels in $o[0]$. For example, suppose a real writer $x$ constructs an unreal world $w_1$ in which there is an imagined writer $y$ who perceives the world $w_1$ and imagines another world $w_2$. Let $imag$ be an abbreviation of $imagination$. Then when $w_1$ is observed by $x$ (i.e., the world label is $w_1$ in the observation), $o[0]=(x, imag)$; when $w_1$ is seen by $y$, $o[0]=(x, imag, y, see)$; when $w_2$ is observed by $y$ (i.e., the world label is $w_2$), then $o[0]=(x, imag, y, imag)$. 

Imagined observers could be almost all kinds analogous to real observers. There could be imagined persons, imagined instruments and sensors, imagined living creatures. Imagined observers could also be totally different to any real observer; for example, there could be imagined observers that obtain a kind of information no real observer could. In principle, an imagined observer could be conceived to obtain any kind of information from any world. Therefore, human imagination greatly expands the variety of worlds, the variety of information, and the variety of information obtaining abilities of observers. 


Of course, imagination is different to real perception, and the ability to correctly distinguish imagination from real perception is crucial to human practice. For example, there is a significant difference between real observers and imagined observers when they are observing the real world simultaneously. The results of a real person's perceptions could be used in a court as evidence, but the result of an imagined observer obviously could not. The recordings of a real instrument could be evidence to support a scientific theory, but the recordings of an imagined instrument could not. In other words, when observing the real world, the obtaining results of real observers and imagined observers have different truth status: The former ones could be regarded as true information if they are faithfully recorded and reported; in contrast, the later ones could not support themselves, and their truth have to be verified by the former ones. 

How about observing an unreal world? Is there any difference in truth status between the obtaining results of different observers? At first glance, it seems not since the whole world is imagined; however, the following example suggests that there is still a difference in truth status. Consider this problem: How does common sense decide the truth of `Hamlet does not die until he is eighty years old'? Hamlet is a fictitious person living in an unreal world conceived by Shakespeare. When he conceived and observed this world, Shakespeare claimed in his book that Hamlet had died when he was very young. For this reason, the statement `Hamlet does not die until he is eighty years old' is regarded to be false.  

The above example shows that statements about an unreal world could also be true or false, and the truth of such an statement is decided by the observations obtained and claimed by the constructors of this world. Every unreal world is directly or indirectly constructed by some real persons. For example, the writer of a novel constructs the world described by the novel, a group of mathematicians including Euclid constructed the space (a class of unreal worlds) of Euclidean geometry. If a real writer $a$ writes a novel describing an unreal world $w_1$, in which an imagined writer $b$ writes a novel describing another world $w_2$, $w_2$ is directly constructed by $b$ and indirectly constructed by $a$. When there are more than one constructor of an unreal world, human practice tells that: There should be an agreement among all those constructors of the world, otherwise the world might contain contradictions or some statements about the world are undecided; if an imaginary world is not completely constructed or not clearly told by its constructors, some statements about this world would also be undecided. 

Besides the constructors of an unreal world, other observers could also imagine observing this world. For example, a reader reading a novel imagines the world narrated by this novel; a fictitious person or animal in a movie perceives the world pictured by this movie; a learner learns a theory; a scholar interprets a special book or theory written by others; and so on. All those observations of some observers from a world not constructed by them, however, are not regarded as actual information about this world: Their truth must be verified by the observations obtained and claimed by its constructors. 

To represent the different truth status of information obtained by different observers from a world, the subcomponent $o[3]$ is introduced into the $observer$ component of all observations. In fact, $o[3]$ is a part of the state of the observer, but is listed separately because of its importance. Suppose that the world label is $w$ in an observation $a$. If $o[3] = actual$ or $o[3] = ac$, the observer is called an {\bf actual observer} of the world $w$, and $a$ is called an {\bf actual observation} of $w$; further if $w=the\ real\ world$, then the observer is called a real observer and $a$ is called a real observation. In contrast, if $o[3] = imaginary$ or $o[3] = im$, the observer is called an {\bf imaginary observer} of the world $w$, and $a$ is called an {\bf imaginary observation} of $w$. The formal definition of actual observers and imaginary observers are as follows. (Imaginary observers should not be confused with imagined observers. An imagined observer is just an observer being imagined, and it could be an actual observer or an imaginary observer.) %

\begin{definition} \label{AIO}
Observers are divided into two categories: actual observers and imaginary observers.

\begin{itemize}
\item An actual observer of the real world is an observer that exists in the real world. They are called real observers.

\item An imaginary observer of the real world is an observer that is imagined by some person and obtains information from the real world. 

\item An actual observer of an unreal world is an observer that constructs this world. 

\item An imaginary observer of an unreal world is an observer that obtains information of this world but does not construct this world. 

\end{itemize}
\end{definition}

\begin{axiom}
Actual observations of a world $w$ represent the true information of $w$. The truth of imaginary observations are decided by actual observations.
\end{axiom}



In some cases not studied in this article, another relation between an observer and a world is important: Whether or not the observer is a part of the world it observes in an observation. When the observer is a part of the world it observes, it is called an {\bf internal observer} of this world; otherwise, it is called an {\bf external observer} of this world. Internal observers of a world are assumed to follow the rules of this world, but external observers are not. Instead, external observers of a world usually have a God's perspective to this world. It is easy to see that real observers must be internal observers of the real world; actual observers of an unreal world are external observers of this world; imaginary observers of a world could be either internal or external of this world. 


Observations performed by different observers might contradict to each other, and such an inconsistency would cause problems when constructing models to interpret natural language. For example, an inconsistency occurs when one observer actually perceives that Marry wears a white dress in a world, and another observer {\bf in the same state} actually perceives that in the same world the same person Marry wears a black dress. This kind of inconsistency has not been precluded by the observation axiom; however, since different observer has an equivalent status in truth, such an inconsistency is not allowed to exist. To avoid such problems, observations are supposed to satisfy some {\bf observer consistencies}.


\begin{definition}[Weak Observer Consistency]\label{WOC}
A set $V$ of observations are {\bf weakly observer consistent} if and only if there is no $x,y \in V$ such that $x,y$ are only different in the observer label sequence $o[0]$ and the obtaining result $re[0]$ (see Definition \ref{PO}).  
\end{definition}


Weak observer consistency implies observation axiom but not vice versa. For actual observations, there is an even stronger consistency: Two actual observers could never be in the same state. This holds in the real world because two real observers, such as two real persons, could never be in the same position simultaneously to observe the same thing. For an unreal world, actual observers are its final constructors, which are supposed to be real persons' imaginations. Since a person's mental activities are generally supposed to happen within his mind and brain, different person's imagination has different state space point, and hence has different state. 




\begin{definition}[Strong Observer Consistency]\label{SOC}
A set $V$ of observations are {\bf strongly observer consistent} if and only if there is no observations $x,y \in V$ such that $x,y$ have different observer label sequence $o[0]$ but have the same world label $w[0]$, the same state $o[2]$ and the same $ac/im$ label $o[3]$.  
\end{definition}

The strong observer consistency obviously implies the weak observer consistency. 

\begin{assumption}[Consistency of Actual Observations] \label{AOC}
Every collection of actual observations satisfy the strong observer consistency.  
\end{assumption}

\subsection{Representation of Indirect Information}



Direct information is directly received by an observer from some world, while indirect information is obtained from some old information through an {\bf information processing procedure}. Informally speaking, an information processing procedure is a procedure that processes some old information and obtains some new information. Thus, such a procedure is logically equivalent to an algorithm whose inputs and outputs are information or its representations. For example, human reasoning is a special kind of information processing procedure.

Although indirect information is not directly received from a world, it is still information that is about some world, is obtained by some agents, could be divided into minimal pieces, and has definite results. Therefore, indirect information could also be represented by primitive observations, except that these observations are determined by the old information and the processing procedure, not by the observer's perception or imagination. 



Then, how to determine the observations that represent some indirect information, using the old information and the processing procedure? There is no general and simple solution to this problem, because the solution would vary according to different processing procedure and different old information. A complete study is too long and too complicated to be presented in this article. Here we just classify all processing procedures into two categories, and determine the $ac/im$ label $o[3]$ in observations that represent indirect information. The $ac/im$ label $o[3]$ is related to the truth status of the information (observation), and hence is especially important for practice. A common mistake is to label an imaginary observation as an actual observation.



\begin{definition}
A {\bf truth-preserving procedure} is an information processing procedure that preserves the truth: If the new information is obtained through a truth-preserving procedure using only true old information, then the new information is true. 
\end{definition}

\begin{assumption}
If an observation could be obtained by a set of actual observations through a truth-preserving procedure, then this observation is labeled as an actual observation; otherwise, it is labeled as an imaginary observation. 
\end{assumption}





The above assumption tells when an observation that represents indirect information could be labeled as an actual observation. A typical kind of truth-preserving procedure is deductive reasoning. All other kinds of human reasoning are not truth-preserving in general, such as induction, abduction and reasoning by analogy. 


There is a kind of information processing procedure having a close relation with a semantic theory: decoding. Interpretation of natural language on some semantic model is in fact a decoding procedure. Decoding obtains new information from various kinds of informative media, including textual materials, meaningful symbols, sounds, images and videos. When a person is reading a sentence `Hitler was dead in 1945' in a book, the visual information of this sentence is a piece of direct information obtained by the reader, the meaning of this sentence is a piece of indirect information, and the procedure of obtaining the meaning from the vision of the sentence is a decoding procedure, which is called an interpretation. The objective of this article is to seek out and clarify those semantic models and decoding procedures (interpretations).

To represent indirect information by observations, however, we should distinguish between the observer of the observation and the obtainer of the observation. The observer of an observation is the agent referred by the observer label sequence in the observation, but the {\bf obtainer} of an observer is the agent who have obtained this observation. The observer is identical to the obtainer when the observation represents a piece of direct information; but when it represents a piece of indirect information, the observer is often not the obtainer of the observation. For example, Tom hears `Marry saw a falling star yesterday', and then interprets this sentence. The meaning of this sentence is the event described by this sentence, and this event could be represented by some observations performed by Marry. The observer of those observations is Marry's eyes, because it is Marry's eyes that saw the falling star; however, the obtainer of these observations is Tom. Finally, we make the following normative assumption for the obtainer of an observation, an assumption which could be viewed as part of the assumption of rationality. 


\begin{assumption}[Obtainer Assumption]\label{OAA}
For any minimal piece of information $x$, its obtainer knows whether $x$ is a piece of direct information or indirect information, and 

\begin{itemize}

 \item when $x$ is a piece of direct information, the obtainer is supposed to know the observation that represents $x$; in other words, the obtainer knows the world where $x$ is from,  the observer who receives $x$, the state and resolution power of the observer, and so on.
 
 \item when $x$ is a piece of indirect information, the obtainer is supposed to know the old information and the processing procedure that obtains $x$, and hence it also knows the observation that represents $x$.  
 
 \end{itemize} 
\end{assumption}

\subsection{Composite Observations}\label{SCO}

All observations discussed before are primitive observations. Primitive observations are clear and fundamental; however, they are not very suitable for interpreting natural language expressions: Each primitive observation contains too less information, while a natural language expression often describes a large amount of information. 

There is an example: A person looks at a cube and says `I see a cube'. What does this sentence mean? The cube is an object with six surfaces, which could be observed from different position at different moment. When a person looking at this cube from a special position at a special moment, he could only directly perceive two images by his eyes from at most three surfaces of the cube. These two images, as discussed before, carry some direct information that is obviously part of the meaning of the sentence. The sentence `I see a cube', however, does not only mean the person sees two images; it also describes some indirect information after his reasoning (object recognition): These two images are from an object's three surfaces; the object has other three surfaces that could be seen from a different position; these six surfaces form a cube that could be seen in the past and in the future; etc. All these indirect information has also to be part of the meaning of the sentence. Therefore, this simple sentence in fact describes a large amount of information that has to be represented by a large set of primitive observations. 

Natural language is powerful to describe a large amount of information by a few sentences, probably with the help of contexts or conventions. When one says some counterfactual conditionals or other hypothetical sentences like `Imagine a world where ....', a whole imaginary world could be constructed or described by a few sentences. For this reason, when interpreting natural language expressions, suitable units are often not primitive observations, but composite observations. 

\begin{definition}
A {\bf composite observation} is a set of primitive observations.
\end{definition}

In principle, a composition observation could be any collection of primitive observations; in practice, a composite observation often consists of primitive observations in a consecutive space region, in a consecutive time segment, by the same observer, or with other naturally related parameter values. For example, recordings of a camera in all pixels at a single time moment form a picture, which could be represented by a composite observation. Composite observations will be more common in later sections.

\section{Cognitive Models}

After the definition and clarification of observations, we could construct models based on them to interpret natural language expressions. Indeed, we could do better: We could construct a kind of mathematical models called {\bf cognitive models} to interpret any meaningful string of symbols. To be clear and general, we distinguish between a cognitive model and an interpretation on this model for a set of symbol strings. This section defines and clarifies cognitive models, while interpretations are left to the next section. 

Before the discussion of cognitive models, some notations and operations related to observations need to be defined. All those operations are called {\bf extraction operations}.

\begin{definition}[Extraction Operation I]\label{EO}
Suppose that $a$ is a primitive observation and $x$ is a parameter in the construction of $a$. Then $x_a$ denote the value of $x$ in $a$. It is also written as $ET(a,x)$ when the notation $x_a$ causes confusions.
\end{definition}

The operation $x_a$ extracts the value of parameter $x$ in primitive observation $a$. Many commonly used parameters of observations have been defined before, such as the world label $w[0]$, the observer label sequence $o[0]$, the $actual$/$imaginary$ label $o[3]$, time point $t$, state space point $s^1$, space point $s^2$, and obtaining result $re[0]$. Then, $w[0]_a$ is the world label in the observation $a$, $t_a$ is the time moment of $a$, and $s^2_a$ is the space point of $a$, etc. When $x$ is not a parameter in the construction of $a$, $x_a$ is undefined.

\begin{definition}[Extraction Operation II]
Suppose that $A$ is a composite observation and $x$ is a parameter in the construction of some primitive observations. Then 
\begin{equation}
x_A = \{x_a : a \in A\}
\end{equation}
\end{definition}

For example, $t_A$ is the set of time points such that for every $t\in t_A$, there is a primitive observation $a$, $t_a=t$, i.e., $a$ happens at $t$. The next class of extraction operations extract those primitive observations in a composite observation having special parameter values. 

\begin{definition}[Extraction Operation III]
Suppose that $A$ is a composite observation. Assume that $x$ is a parameter in the construction of some primitive observations and $D^x$ is the domain of $x$. Let $D \subseteq D^x$. Then 
\begin{equation}
A_{x_a \in D} = \{a\in A :   x_a \in D\} 
\end{equation}
$A_{x_a \in D}$ is also written as $A_{x_a = d}$ when $D$ contains only one element $d$.
\end{definition}

For example, suppose that $t_0$ is a time moment, then $A_{t_a = t_0}$ is the subset of $A$ containing all and only those primitive observations that happens at $t_0$. The restrictions in the above definition could be combined with logic connectives and quantifiers. For example, 

\begin{equation}
A_{x_a \in D \thinspace \wedge \thinspace y_a \in D'} = \{a\in A :   x_a \in D \wedge y_a \in D'\}
\end{equation}

Other kind of extraction operations could be defined if necessary. Some important notations related to time and space are defined as follows. 

\begin{definition}[Segment]
Suppose that the structure $M=\langle V, \leq \rangle$ is a strict linear ordering (asymmetric, transitive and linear). Then the segment $[v_1, v_2]$ of $M$ between two elements $v_1, v_2 \in V$ is defined as follows
\begin{equation}
[v_1, v_2] = \{v : v\in V, v_1 \leq v \leq v_2\} 
\end{equation}
where $v_1, v_2$ are called the {\bf start point} and the {\bf end point} of this segment respectively.
\end{definition}

\begin{definition}[Region]
Suppose that the structure $M=\langle V, \tau \rangle$ is a topological space. Then a region $S \subseteq V$ of $M$ is a connected subspace of $M$. Moreover, the {\bf boundary}, {\bf interior} and {\bf exterior} of a region $S$ are defined as usual, and they are denoted by $\partial S$, $int(S)$ and $ext(S)$ respectively. 
\end{definition}

\begin{theorem}
The start point and end point of a segment are unique. The boundary, interior and exterior of a region are unique. 
\end{theorem}

These notations are very intuitive, though their definitions appear complicated. In fact, they just describe some simple properties of time and space of a world in scientific research or in everyday life. Their extensive study could be found in books such as \cite{Hrbacek1999, Munkres2000}. Moreover, we use $\mathbb{P}(A)$ to denote the power set of $A$ in the following discussion.  

\subsection{Definition and Clarification of Cognitive Models}

\begin{definition}[Cognitive Model]\label{CM}
A cognitive model $\mathfrak{M}$ is a multi-domain mathematical structure
\begin{equation}
\mathfrak{M}= \langle  \mathbb{D}^0_0,  \mathbb{D}^0_1, ..., \mathbb{D}^i_0,  \mathbb{D}^i_1, ...,  \mathbb{R}_0,  \mathbb{R}_1, ..., ...,  \mathbb{O}_0,  \mathbb{O}_1,...\rangle
\end{equation}
where  
\begin{itemize}
\item Each $ \mathbb{D}^i_j$ is a set and is called a domain of $\mathfrak{M}$. Each $e\in  \mathbb{D}^i_j$ is called a domain element. For each $i,j>0$, $ \mathbb{D}^i_j \subseteq  \mathbb{D}^i_0$, and $ \mathbb{D}^{i+1}_j \subseteq \mathbb{P} ( \mathbb{D}^{i}_j)$. 


\item $ \mathbb{D}^0_0$ is a set of primitive observations. 


\item $ \mathbb{D}^1_0$ is a set of composite observations, each of which is a subset of $ \mathbb{D}^0_0$. 

\item $ \mathbb{D}^1_1$ is the set of all worlds. A world $\hat{u} \in  \mathbb{D}^1_1$ is defined as follows:
\begin{equation}
\hat{u} = \{a\in  \mathbb{D}^0_0 : w[0]_a = u \}
\end{equation}

Every world is assumed to have a unified measurement system of time and space, which defines the time and space of this world. For each world, the time is assumed to be strict linear ordering and the space is assumed to be a topological space. 

\item $ \mathbb{D}^1_2$ is the set of all subworlds. A subworld $(\hat{u}, \hat{v})$ of a world $\hat{u}$ is defined as:
\begin{equation}
(\hat{u}, \hat{v}) = \{a\in  \mathbb{D}^0_0 : w[0]_a = u \wedge w[1]_a = v \}
\end{equation}

\item $ \mathbb{D}^1_3$ is the set of all processes. A process is a composite observation that consists of all and only those primitive observations in a world, in a time segment, and within a space region at each time moment. Formally speaking, $P$ is a process if and only if there is a world $ \hat{u} $ and a time segment $[t_{P,min}, t_{P,max}]$ where for every time moment  $t \in [t_{P, min}, t_{P, max}]$ there is a a space region $S_{P,t}$ such that
\begin{equation}
P = \{ a \in \hat{u} : t_a \in [t_{P, min}, t_{P, max}] \wedge s_a \in S_{P,t_a} \}
\end{equation}

$P$ is called a process of $ \hat{u} $. If all primitive observations in a process have the same subworld labels, it is called a process of this subworld. The time moments $t_{P, min}, t_{P, max}$ are called the {\bf start moment} and {\bf end moment} of $P$ respectively. Moreover, we say $P$ exists in or goes through the time segment $[t_{P, min}, t_{P, max}] $, and $P$ exists in or occupies the space region $S_{P,t}$ at each moment $t \in [t_{P, min}, t_{P, max}] $. 




\item $ \mathbb{D}^1_4$ is the set of all objects, each of which is a process that satisfies some further conditions.  

\item $ \mathbb{D}^1_5$ is the set of all actions, each of which is both a process and a subset of an object. 

\item $ \mathbb{D}^1_6$ is the set of all states. A state is a process happening at only one moment, i.e., whose start moment and end moment are the same. Moreover, $ P_{t_0} = \{ a\in P : t_a =t_0 \} $ is called the state of the process $P$ at the time moment $t_0$.

\vspace{6pt}

$ \cdots \  \cdots$

\item $ \mathbb{D}^2_0 \subseteq \mathbb{P} ( \mathbb{D}^1_0)$ is a set of composite observation classes.

\item $ \mathbb{D}^2_1 \subseteq \mathbb{P} ( \mathbb{D}^1_1)$ is a set of world classes.

\item $ \mathbb{D}^2_2 \subseteq \mathbb{P} ( \mathbb{D}^1_2)$ is a set of sub-world classes.

\item $ \mathbb{D}^2_3 \subseteq \mathbb{P} ( \mathbb{D}^1_3)$ is a set of process classes. 

\item $ \mathbb{D}^2_4 \subseteq \mathbb{P} ( \mathbb{D}^1_4)$ is a set of object classes. 

\item $ \mathbb{D}^2_5 \subseteq \mathbb{P} ( \mathbb{D}^1_5)$ is a set of action classes. 

\item $ \mathbb{D}^2_6 \subseteq \mathbb{P} ( \mathbb{D}^1_6)$ is a set of state classes. 

\vspace{8pt}

$ \cdots \  \cdots$



\item $\mathbb{R}_0$ is a set of relations, each of which is a set of sequences of elements in $\bigcup_{i,j} \mathbb{D}^{i}_j$. Since a set is regarded as an unary sequence, so for every $i,j$, a subset of $\mathbb{D}^i_j$ is a relation too.

\item $\mathbb{R}_1$ is a set of relations, each of which is a set of sequences of elements in $\bigcup_{i,j} \mathbb{D}^{i}_j \cup \mathbb{R}_0$.

\item For each $i>0$, $\mathbb{R}_{i+1}$ is a set of relations, each of which is a set of sequences of elements in $\bigcup_{i,j} \mathbb{D}^{i}_j \cup \mathbb{R}_0 \cup \dots \cup \mathbb{R}_i $. 

\item $\mathbb{R}  = \bigcup_{i} \mathbb{R}_i$ is the set of all relations. If all sequences in a relation has the same length $n$, then it is called an $n$-ary relation. When $n=1$, it is called a property as usually.

\item $ \mathbb{O}_0$ is the set of all extraction operations defined before.

\item $ \mathbb{O}_1$ is a set of operations, each of which is an $n$-ary function for some $n>0$: $A_1 \times \dots \times A_n \rightarrow A_{n+1}$, where for every $k\leq n+1$, $A_k\subseteq \bigcup_{i,j} \mathbb{D}^{i}_j \cup \mathbb{R} \cup  \mathbb{O}_0$.

\item For each $i>0$, $\mathbb{O}_{i+1}$ is a set of operations, each of which is an $n$-ary function for some $n>0$: $A_1 \times \dots \times A_n \rightarrow A_{n+1}$, where for every $k\leq n+1$, $A_k\subseteq \bigcup_{i,j} \mathbb{D}^{i}_j \cup \mathbb{R} \cup  \mathbb{O}_0 \cup \dots \cup \mathbb{O}_i $.

\item $\mathbb{O}  = \bigcup_{i} \mathbb{O}_i$ is the set of all operations. If an operation is an $n$-ary function, it is called an $n$-ary operation.



\end{itemize}



\end{definition}


It follows from Assumption \ref{AOC} that the set of actual observations contained in $\mathbb{D}^0_0$ in a cognitive model $\mathfrak{M}$ satisfies the strong observer consistency \ref{SOC}. Imaginary observations in a cognitive model are supposed to satisfy the weak observer consistency. 

\begin{assumption}[Observer Consistency of Cognitive Models] \label{SOCCM}
For every cognitive model $\mathfrak{M}$, its domain $\mathbb{D}^0_0$ satisfies the weak observer consistency \ref{WOC}. 
\end{assumption}


Informally speaking, a cognitive model is an organization of information acquired by some agent. It could also be viewed as a belief base of its constructors, with a special organizational strategy. It is constructed to reflect the constancy and similarities in the information, and is expected to be efficient and convenient for use. There could be many applications with cognitive models, but this article only considers one of them: interpreting symbol strings, especially natural language expressions. To interpret natural language expressions, a semantic model has to follow some common practice. A cognitive model is constructed to reflect this common practice as much as possible, and also to be as clear and general as possible. Some complicated clauses in the definition will be clarified and justified later, where the motivation behind them will also be mentioned.



Although this article interprets natural language based on cognitive models, the empirical problem of whether a real person or a real human community actually uses a cognitive model stored in their brains to organize information and interpret natural language will not be discussed here, but left to cognitive scientists who are interested in. Moreover, as the first step of a long-term research, also for simplicity, cognitive models studied in this article are totally certain and precise, without any representation of incomplete, inaccurate, vague, wrong or uncertain information; for this reason, related language phenomena will not be discussed either. How to deal with incompleteness, fallacies, uncertainty or vagueness is a large topic left to future study.

\subsubsection{Multi-Domain Structure}


Traditional meaning models, such as those ones used in Formal Semantics, are commonly single-domain models \cite{Allen1995, Jurafsky2008}; however, there could be many high-order quantifiers in natural language, just like those ones in mathematical language. To avoid technical complexity of high-order interpretation but retain the strong expressivity, mathematicians generally interpret mathematical language on the universe of sets \cite{Hrbacek1999}. A similar approach is adopted in this article: Whenever there is a quantifier with a range that is not an existing domain, this range is added as a new domain to the cognitive model. Then all high-order quantifiers could be reduced to first-order quantifiers \cite{Hrbacek1999}, although the semantic model has to be a multi-domain structure \cite{Enderton2001}. 
 
The domains in a cognitive model are divided into many levels: The first and the bottom level is $D^0_0$, which is a set of primitive observations representing the set of all minimal pieces of information obtained by the model's constructors. The second level is $D^1_j$ ($j \geq 0$), where each domain is a set of composite observations. The third level is $D^2_j$ ($j \geq 0$), where each domain is a set of sets of composite observations. To avoid repetition of notation, the words `class' is regarded as having the same meaning of `set' in this article. Then each domain $D^2_j$ is a set of composite observation classes. $D^1_j$ and $D^2_j$ are the most commonly used levels of domains when interpreting natural language words.

Domains $ \mathbb{D}^0_j$ ($j>0$) are left undefined. They could be  some relations or operations defined afterwards, because relations and operations could also be ranges of some quantifiers in natural language. In fact, a domain in a cognitive model could be any combination of primitive observations, only if such a combination could be a set constructed by ZFC set theory from $D^0_0$ \cite{Hrbacek1999}. 

\subsubsection{World, Time and Space}

The definition of a world (subworld) is easy to understand: It just gathers together all those primitive observations labeled by this world (sub-world). (We should distinguish between worlds and world labels.) Since the domain $D^0_0$ is supposed to contain all information obtained by the constructors, a world contains all information obtained by the constructors from this world. This exemplifies the fundamental idea behind the construction of a cognitive model: Every thing that human could obtain information from is represented by the collection of all information obtained by the model's constructors from this thing. Besides worlds and subworlds, for example, an event (process) is represented by the collection of all information obtained from this event (process); an object is represented by the collection of all information obtained from this object; etc. 

The representation of the same thing, however, could be different in detail to different constructors. This is either because the constructors have different opinions on whether a piece of information is from the thing being considered, or because they have obtained different collection of information from the thing. Whether a piece of information is obtained from a world is easy to know, since this is generally assumed to be an inner ability of the obtainer of the information (\ref{OAA}). In contrast, whether an observation is obtained from an event or object is not so easy to decide. Usually, the constructors of a cognitive model have to follow some common practice or establish some conventions.  

In the construction of primitive observations, every observer has its own measurement system to turn information into parameter values. In the definition of worlds in a cognitive model, however, it requires that each world has a unified measurement system of time and space. Such a system for a world makes all time information and space information from this world mapped to elements in the same domain, no matter which observer obtains the observation; so that observations from the same world but performed by different observers could be compared in time and space. Sometimes different worlds could share a common measurement system of time and space, which makes the observations in different world could also be compared in time and space. However, this is a choice, not a requirement. 

In the world we live, for example, people use a measurement system---a calendar and a clock---to measure and record time. In different eras or areas, people had invented many different calendars such as the Hebrew calendar and the Gregorian calendar; new calendars could be invented if necessary. Different calendar or clock measures and records time in a different way, often with different zero point and different unit. Some relative or inaccurate time measurements could also be used, such as the everyday usage of today, tomorrow, morning and evening. With all these varieties, however, it is generally believed that there is a unified and most accurate method to measure and record time of the real world; all other measurements could then be mapped into this unified system. Such a unified measurement system of time makes all observations from the real world could be compared in time, which is crucial to define and talk about various kinds of composite observations such as processes and objects. 

The space of a world is similar. There could be many different coordinate systems to measure positions and orientations in a world, each of which could have different zero point, length units, coordinates directions, etc. For the same space position or orientation, different coordinate system usually generates different results; however, it is also generally believed that there is a unified measurement system of a world's space, where all recordings from other measurement systems could be mapped into this system.

For any world, the time is assumed to be strict linear ordering. There are two main reasons behind this assumption. One is that strict linear ordering fits most people's intuition about the time of the real world. The second is that this assumption makes the semantic theory simple without loss of generality. The main opposition to this assumption might be that there could be worlds having branching time or parallel histories. A cognitive model, however, regards each history as a different world, and any branching time world or parallel histories world in other theories is represented by a class of linear time worlds in a cognitive model. If it does not consider the metaphysical or notation difference, then such a representation is mathematically equivalent to other representations.

For each world, its space is supposed to be a topological space. The reason is simple: Under this assumption, it could define space regions and boundaries, and then could define processes, objects, etc. In fact, topological space is often too general to be used in everyday life or scientific research. Instead, Euclidean spaces and metric spaces are more commonly used. Of course, there is no restriction to construct worlds with other kinds of topological space. 


\subsubsection{Process and Object}

Process is the most important concept in cognitive models. Many other important concepts, such as object, action and state, are special kinds of processes. Processes are defined to characterize our intuitions of events. Generally speaking, an event is represented in cognitive models by one process, a sequence of processes or a class of sequences of processes. The definition of processes is made to be as general as possible; it implies the following theorem, showing why the concept of process is fundamental.  

\begin{theorem}[Uniqueness of Process] \label{UP}
In each cognitive model, given a world $ \hat{u}$, a time segment $[t_1, t_2]$, and a space region $S_{t}$ for each moment  $t \in [t_{1}, t_{2}]$, there is one and only one process.
\end{theorem}

\begin{corollary}
In each cognitive model, given a world $ \hat{u}$, a time moment $t$, and a space region $S_{t}$, there is one and only one state.
\end{corollary}

The above theorem and corollary summarize some of our common sense about events. Some other important features of processes are clarified as follows.  


\begin{enumerate}

\item {\bf Part of a World}: A process should be a part (subset) of a world. This requirement is mainly for clear and simplicity. If one imagines a cross-world event, then it could be represented as a class of process sequences in a cognitive model. 

\item {\bf Consecutive in Time}: A process should go through a unique segment of time. If one thinks an event goes through several disjoint segments of time, then depending on different choices, it could be represented as a class of processes, a class of process sequences, or a single process with empty space regions at some moments during the whole time segment. 

\item {\bf Consecutive in Space}: At each time moment, a process should occupy a unique space region. If one thinks an event happens in several disjoint space regions at a moment, then it could be represented as a class of processes or a class of process sequences. 

\item {\bf Complete Covering}: A process contains all information (primitive observations) that could be obtained within the time segment it goes through and within the space region it occupies at each moment. This is the crucial feature that distinguishes a process from other composite observations. This feature implies the weak objectivity of processes.

\item {\bf Weak Objectivity}: Because of the complete covering, there is no restriction to observers in primitive observations contained in a process. In other words, a process contains a primitive observation as long as it has been obtained by some observer within the time segment the process goes through and within the space region the process occupies at some moment. This removes the subjective influence of observers and establishes a weak objectivity of processes: A process is not decided by a single observer's observation. This weak objectivity is enough for people to think and talk about things in a world, not just their observations. 


\end{enumerate}




Objects are especially important in classic logic, and also important in cognitive models; besides events, objects are the most common things talked about in natural language. At first glance, it seems strange to think that an object is a process (a special event); however, if we put aside those metaphysical debates and only consider the information that could be obtained from an object, then we could find that the collection of all observations from an object just forms a process in a cognitive model. Since an object is a process, we have the following corollary.

\begin{corollary}
At any moment, an object could only exist in one world and occupy one space region. 
\end{corollary}

Not all processes are objects. Intuitions about objects tell that there are conditions satisfied by objects but not by processes in general. In other words, a process has to satisfy more requirements to become an object. The intuitions of objects, however, are often not clear enough, so it is hard to define objects by a set of necessary and sufficient conditions. The following tentative solution illustrates some conditions that distinguish an object from a process. A complete study is left to future research.

Moreover, in the framework of cognitive models, objects are defined to clarify their inner structure and contents. Therefore, mathematical objects (objects in mathematical worlds) need no special discussion because either these objects have no inner structure (just some points in the domain) or their inner structures have been clearly defined by corresponding mathematical theory. For this reason, the following discussion only considers objects in the real world, in scientific worlds, or in other imaginary worlds where objects have descriptive contents. 

\begin{enumerate}

\item {\bf Spatial Difference}: Objects should exist in some world or subworld with spatial difference. In other words, for a world or subworld to have objects, observers should be able to distinguish spatial difference from this world or subworld. This is not required for processes in general. For example, people often talk about mental processes such as feeling, reasoning and imagination; however, since human minds have no spatial difference, those mental processes are commonly not not regarded as objects. 

\vspace{5pt}

People also talk about things like mental objects; nonetheless, mental objects are objects in some imaginary worlds with spatial difference, not processes in someone's mind recognized by his self-consciousness. For example, `imagine a world that has a gold mountain'. Such an imagination is a person's mental process, and hence is not an object; however, the gold mountain is a mental object that exists in a world imagined by this person. In the first case, the person's self-consciousness is the observer; in the second case, the person's imagination is the observer. 

\item {\bf Strict Boundary}: An object usually has a clear boundary at each moment it exists. Moreover, the boundaries of an object at different moments are usually identical (stationary object), isomorphic (moving object), or changed with some regularities (deformed object). There is no such a restriction to processes in general. For this reasons, people often talk about the figure or shape of an object, but never talk about the figure or shape of a process (event). For the same reason, fluids (liquid or gas) are usually not regarded to be objects, but if the shape of fluid is fixed by some container, such as a glass of water or a balloon of gas, then it is sometimes also regarded as an object. 

\item {\bf Disjoint in Space}: At each moment, different object commonly has a disjoint space region, except that one object is a part of another. In other words, if we do not consider that some object is a part of another, then at any moment, a space region could only be occupied by at most one object. For this reason, the interior of an object is usually unreachable to other objects, unless it is broken by them and becomes an object with a different boundary. Processes generally do not have such a requirement: Different processes could be overlapped without a part-whole relation.



\item {\bf Strict Start and End Moment}: The start moment and end moment of an object is more clear and restricted than a process. There is often a dramatical change (essential difference) among observations before the start moment, during the time segment the object exists, and after the end moment. This restriction is not imposed on processes in general. For example, a person is supposed to be born at a moment and dead at a moment, any proper subset of this person is not a new person. However, a proper subset of a running process could also be a new running process. 

\item Some other restrictions.

\end{enumerate}



The above conditions obviously do not completely decide the space boundary or the start/end moment of an object. The space boundary or the start/end moment of an object might be controversial, when people's intuitions about the object are not clear enough to reach an agreement. They could also be changed when people's intuitions, common agreements, or conventions are changed. For example, when does a person start (born) or end (die)? Does a person start at the first moment when his zygote is formed, when he is a fetus with a person-like appearance, when he completely comes out of his mother's uterus, or when he is mature? Does a person end at the first moment when any part of his body is not able to move, when his brain is dead, when his body starts to decay, or when his body becomes ashes or dust, or never end since people believe that the soul would be immortal and exist eternally? There would be no unique right answer. In fact, no matter how serious the debate could be, a complete answer is probably just a common agreement or convention among the constructors of the cognitive model, usually a large community, to meet some of their practical purposes. 

\subsubsection{Relation and Operation}\label{SRO}

In classic logic, relations and operations are sets of sequences of objects \cite{Ebbinghaus1984, Enderton2001}. In cognitive models, however, they could be more general and complicated, as the reader could learn from Definition \ref{CM}. Any domain element could be an element of a sequence that belongs to a relation or an operation. Moreover, relations and operations are defined recursively, because natural language and many other languages are recursively formulated and interpreted, as the reader will see later. 

The most commonly used kind of relations in cognitive models do not consist of sequences of objects, but sequences of nonempty composite observations. This is a generalization to the traditional approach, and the following examples show the reason for such a generalization. 


Suppose that person $A$ and person $B$ are friends. Classic logic would use a pair $(A,B)$ to represent this relation. However, it is probable that the friendship between $A$ and $B$ starts from time moment $t_1$ and ends at time moment $t_2$, where the segment $[t_1,t_2]$ is much smaller than $A$'s life or $B$'s life. If we still use $(A,B)$ to represent the friendship between $A$ and $B$, then it could not answer questions like `Are $A$ and $B$ friends at time $t$?'

In classic logic, this problem could be solved by representing the friendship using a longer sequence $(A,B, t_1, t_2)$. In cognitive models, it is represented by the pair $(A',B')$ where 
\begin{equation}
A' =\{ a\in A : t_a \in [t_1,t_2] \}  \quad   B'  = \{ a\in B : t_a \in [t_1,t_2] \} 
\end{equation}

In other word, the time information is implicitly encapsulated inside the composite observations $A'$ and $B'$, which are subsets of the objects $A,B$. Then, the question `Are $A$ and $B$ friends at time $t$?' is answered by checking whether there are observations in $A'$ and $B'$ at time $t$.  

Sometimes things could become complicated and the advantages of such a representation would be more obvious. For example, the friendship between $A$ and $B$ might not last all the time from $t_1$ to $t_2$, but exists intermittently in $[t_1, t_3], [t_4, t_5], ..., [t_i, t_2]$. Moreover, during a segment such as $[t_4, t_5]$, it might be the case that $A$ regards $B$ as friend but $B$ does not regard $A$ as friend. In such a circumstance, classic logic has to employ very long sequences with indefinite length to encode all these restrictions, but in cognitive models, we could also use $(A',B')$ to represent the friendship between $A$ and $B$, where $A'\subseteq A, B'\subseteq B$ are made to include all and only the information (observations) regarded to belong to the friendship of $A,B$. 

Such a representation could be very fine-grain and elaborate. For example, during the existence of the friendship of $A,B$, it is probable that only a few actions (including mental processes) of them show that $A$ and $B$ are friends, most other actions are unrelated, so it is more convenient to represent the friendship between $A$ and $B$ by $(A',B')$ where $A'$ and $B'$ contain only those special actions. Classic logics could hardly do this job, or the representation would be far more complicated. 

If we want to further simplify the representation as much as possible, the friendship of $A,B$ could be represented by a single composite observation $C=A'\cup B'$, and such a representation could also be useful in most cases. 

A similar example is to represent the property red. In classic logic, red is represented by the set of all red objects. However, an object is probably not red all the time in its life, or not red all over its body; the red color on an object could be dark red or light red, pink red or wine red, etc; the red color could change through time on an object. In cognitive model, the property of red is represented by a set of red instances, where a red instance is a composite observation that is a small subset of an object containing primitive observations with only red obtaining result (The observer is supposed to be able to distinguish many colors, one of them is labeled as red). 




\vspace{8pt}

This section is concluded with the following axiom, which is the formulation of Leibniz's Law under the framework of cognitive models.

\begin{axiom}[Leibniz's Law]
Two things (worlds, sub-worlds, processes, states, objects, actions, etc) are identical with respect to a cognitive model if and only if they are represented by the same set in this cognitive model.
\end{axiom}

To understand this law, consider the following example. When a girl $A$ says something like `If I were a boy, then...', she imagines a world where a boy $B$ is really similar to her. According to Leibniz's Law, $A$ and $B$ could never be identical since they have different properties: $A$ is a girl and $B$ is a boy; $A$ lives in a world and $B$ lives in another world. In a cognitive model, $A$ and $B$ are two objects having different content (primitive observations), so they are not identical. For the same reason, the two worlds that $A$ or $B$ exists are also not identical. Strictly speaking, therefore, it is unreasonable for the girl to think that she and the imaginary boy are the same person, although people often thinks and speaks in this way. Even if we only consider the minds of the boy and the girl, and let the girl's mind (mental activities) to be a copy of the boy's, they are also not identical, just like two copies of a book are two different objects, not to mention that they exist in different world. Cognitive models could tell all those subtleties.

\subsection{Practical Construction of Cognitive Models}

It is obvious not an easy task to actually build a useful cognitive model. There are plenty of primitive observations, composite observations, processes, objects and other domains elements. Even a single process (object, etc) or a class of processes (class of object, etc) is not easy to construct: For a process, the constructors have to decide its start and end moment, and to decided the space region at each moment; for a class of processes, it needs to decided whether a process belongs to this class or not. How could one do such a decision or selection?  There is no general or simple answer, and the selection or decision algorithms could be very complex or still unknown to mankind. This section just provides some general ideas for doing the job. The detailed work is left to the future.


To reduce the total amount of work, a cognitive model is usually built upon composite observations, not primitive observations. One reason has been discussed before (see \ref{SCO}); another reason is that, usually, composite observations could also have a unified mathematical representation, which compresses the information it contains and extracts useful features for them to be compared. 

\begin{definition}[Representation of Composite Observations]
The mathematical representation of a composite observation under an algorithm $\Psi$ is a sequence of parameter values, where each parameter is a mathematical variable with a domain defined by $\Psi$, and the parameter values in the sequence are also decided by $\Psi$. Each parameter is called a {\bf feature} of this composite observation under $\Psi$, and all features defined by $\Psi$ are denoted by $V_{\Psi}$. A subset of the domain of a feature is called a {\bf range} of this feature. 
\end{definition}


The main idea of selecting elements in a cognitive model to form a class is to use the constancy and similarities among these elements. As an illustration, we try to define constancy and similarities for composite observations. 

\begin{definition}[Constancy and Similarities among Composite Observations]
Suppose that a set of composite observations $C_0$ are represented by a set $C$ of value sequences under the algorithm $\Psi$, and let $V_1,V_2 \subseteq V_{\Psi}$. Then, 

\begin{itemize}
\item If all sequences in $C$ have the same values for every feature in $V_1$ while have different values for every feature in $V_2$, then we say, given the algorithm $\Psi$, composite observations in $C_0$ are constant in $V_1$ under the change of $V_2$.

\item If all sequences in $C$ have values within a given range $r_e$ for each feature $e \in V_1$ while have different values for each feature in $V_2$, then we say, given the algorithm $\Psi$ and those ranges $\{r_e\ |\ e \in V_1\}$, composite observations in $C_0$ are similar in $V_1$ under the change of $V_2$.

\end{itemize}

\end{definition}



It is easy to see that constancy is a special kind of similarities, and it is similar to define constancy and similarities among primitive observations, classes of composite observations, or other elements in cognitive models. Constancy and similarities are the basis for common sense reasoning and scientific research. With the constancy and similarities, and some other conventions, it would be easier to build a cognitive model. Some general ideas are summarized as follows.

\begin{itemize}


\item {\bf Data Collection and Organization} In practice, various kinds of data could be collected to characterize a process or an object people talking about. Data could be recorded as or turned into composite observations, and stored as parts of the process or object. Processes and objects could then be classified into different categories. Other domains or domain elements are constructed similarly.

\item {\bf Hierarchical Structures}: Subworlds classes, processes classes, objects classes and many other large classes could be organized into several hierarchical structures, each of which is a directed or undirected tree where every node of the tree represents a class and every edge represents a subset relation, a part-entirety relation, or another kind of relation. Dictionaries such as WordNet have provided some of these hierarchical structures.


\item {\bf Typical Elements or Prototypes}:  A process (objects, etc) or a class of processes (objects, etc) usually consists of elements with certain similarities. Then, the set could be constructed by selecting a small subset as its typical elements or creating some prototypes, using algorithms to define features and formulate the similarities, and using these features and similarities to decide whether an arbitrary element belongs to this set or not. 

\item {\bf Operations and Relations}: An operation is a function, so all techniques of constructing functions could be used to construct operations. From a computational view, functions are realized by algorithms, and the construction of operations could be reduced to construction of algorithms. A relation could be turned into a special kind of function, and hence all techniques for constructing functions could also be applied to the construction of relations. 




\end{itemize}

Although practical construction of cognitive models are important, this article focuses on theoretical study. Therefore, the following discussion does not consider any practical construction of cognitive models, but just assume that cognitive models are defined as in Definition \ref{CM}.

\section{Interpretations on Cognitive Models}

This section discusses interpretations of symbol strings on cognitive models, including natural language expressions. Interesting enough, symbols and symbol strings could be defined as elements of cognitive models. 

\begin{definition}[Symbol Strings]
A {\bf concrete symbol} is a composite observation in some cognitive model, and an {\bf abstract symbol} is a class of concrete symbols. A {\bf concrete symbol string} or a {\bf concrete string} is a sequence of concrete symbols, and an {\bf abstract symbol string} or an {\bf abstract string} is a sequence of abstract symbols. 
\end{definition}

Since a set is regarded as an unary sequence, a single symbol is a symbol string too. The above definition only provides a necessary condition for concrete symbols under the framework of cognitive models.  Whether a composite observation is a concrete symbol depends on whether people use it to refer to other things, i.e., depends on whether it has interpretations. 


The most commonly used concrete symbols are some special objects. For example, when one writes a letter `E' on a paper, all observations of this letter form an object in some cognitive model, which is a concrete symbol. A lot of concrete symbols `E' have been or will be written, printed, spoken or imagined. All those concrete symbols `E' form the abstract symbol $E$. Processes and actions could also be  symbols, e.g., sound waves, sign language, people's facial expressions and gestures. 


A set of symbol strings is often called a {\bf language}. To simplify the notations, we say $e$ is an element of a cognitive model $\mathfrak{M}$ (written as $e\in \mathfrak{M}$) if and only if, (1) $e$ is a domain element but not a primitive observation, i.e., there is a domain $\mathbb{D}^i_j$ in $\mathfrak{M}$ such that $i\neq 0$, $j\neq 0$ and $e \in \mathbb{D}^i_j$; or (2) $e$ is a relation, i.e., there is a $\mathbb{R}_i$ in $\mathfrak{M}$ such that $e \in \mathbb{R}_i$; or (3) $e$ is an operation, i.e., there is an $\mathbb{O}_j$ in $\mathfrak{M}$ such that $e \in \mathbb{O}_j$. 


\begin{definition}[Interpretation for Symbol Strings]\label{ISS}





An interpretation $\mathcal{I}$ on a set of cognitive models $\{ \mathfrak{M}_i: i\in I\}$ for a language $\mathbb{S}$ is a function that maps each $x \in \mathbb{S}$ to a set $\mathcal{I}(x)$ where one of the following conditions holds:
\begin{itemize}
\item $\mathcal{I}(x)$ is an empty set. In this case, $x$ is said to have {\bf empty meaning}. 

\item $\mathcal{I}(x)$ contains a single element $e\in \mathfrak{M}_i$ for some $i\in I$. In this case, $x$ is called a {\bf single-meaning string} under $\mathcal{I}$. 

\item $\mathcal{I}(x)$ contains more than one element, each of which is an element of $\mathfrak{M}_i$ for some $i\in I$.  In this case, $x$ is called a {\bf multi-meaning string} under $\mathcal{I}$.  



\end{itemize}

In either case, whenever $e\in \mathcal{I}(x)$, $e$ is called a {\bf denotation} or a {\bf referent} of $x$ under $\mathcal{I}$ ($x$ denotes $e$ or refers to $e$), and the pair $(x,e)$ is called an {\bf explanation} of $x$. Each denotation or explanation of $x$ is called a {\bf meaning} of $x$ under $\mathcal{I}$.  $\mathcal{I}(x)$ is called the set of all denotations (referents) of $x$ under $\mathcal{I}$. When $e\in \mathcal{I}(x)$ and $e = \varnothing$, $e$ is called an {\bf empty denotation} of $x$ under $\mathcal{I}$. $\{ \mathfrak{M}_i: i\in I\}$ is called the {\bf underlying set of cognitive models} of $\mathcal{I}$ or the {\bf set of underlying cognitive models} of $\mathcal{I}$. 




\end{definition}


 
Empty meaning symbol strings include those pure syntactic symbols such as commas `$,$' and parentheses `$( )$'. A meaningful symbol string is a symbol string that has a denotation on some cognitive model, including an empty denotation. Meaningful symbol strings could be concrete or abstract. The common practice is that people define interpretations on abstract strings, and assume that all concrete strings contained in an abstract string are interpreted as the same. Since interpretations for concrete strings are more general, the following discussion always assumes that the symbol strings being interpreted are concrete if there is no other instruction. 

Two kinds of meaning---denotations and explanations---have been defined for symbol strings. An explanation of a symbol string tells which denotation the symbol string actually has; it is introduced because denotation alone is not enough to satisfy human intuitions about meaning in some cases. Suppose that, for example, under the interpretation $\mathcal{I}$, the symbol $=$ denotes the equivalent relation, and two different single-meaning strings $x_1,x_2$ have the same denotation $e$. Consider these two sentences $x_1=x_2$ and $x_1=x_1$. Intuitions tell that they have different meaning; however, if denotation is the only kind of meaning, $x_1,x_2$ must have the same meaning, which implies that $x_1=x_2$ and $x_1=x_1$ have the same meaning according to the general rule of replacement. This contradicts to our intuitions. In contrast, if explanation is used as meaning for $x_1,x_2$, $x_1=x_2$ and $x_1=x_1$ would have different meaning. Which kind of meaning is used often depends on the context. In fact, besides denotations and explanations, a third kind of meaning---senses---would be introduced later when interpreting a language $\mathbb{S}$ that is compositional. 

\begin{definition}[Compositional Language]
A language $\mathbb{S}$ is strictly {\bf compositional} under the interpretation $\mathcal{I}$ if there is $\mathbb{S}_0 \subseteq \mathbb{S}$ such that $\mathbb{S}_0 \neq \mathbb{S}$ and for any string $x \in \mathbb{S} - \mathbb{S}_0$, the following conditions hold.

\begin{itemize}
\item $x$ is composed by strings in $\mathbb{S}_0$ following a collection of syntactic rules;

\item The interpretation of $x$ is composed by the interpretation of strings in $\mathbb{S}_0$ following a collection of semantic rules.  

\end{itemize}
$\mathbb{S}_0$ is called the set of {\bf primitive strings} of $\mathbb{S}$.
\end{definition}

Formal languages are commonly strictly compositional \cite{Church1956, Ebbinghaus1984, Enderton2001}. Although the rules might be complicated and full of exceptions \cite{Jurafsky2008, Quirk1985}, natural language is basically compositional: most non-primitive strings and most of their meanings are compositional. For a compositional language, only primitive strings are interpreted as in Definition \ref{ISS}, other strings would be interpreted following the semantic rules. The semantic rules for natural language would be discussed later.




When speaking out or writing down a symbol string for effective communication, people usually mean only one thing; in other words, this symbol string has only one explanation (denotation) in this particular circumstance. If it is not the case, there would be ambiguities in meaning, which often make the listener confused and the communication hard to continue. To solve the ambiguities, either the speaker has to provide more information to fix the explanation, or the listener has to obtain and utilize more information to decide the explanation. Information used to reduce possible meanings of a symbol string is called {\bf context}. In principle, any interpretation under which some symbol strings are multi-meaning could be assisted by some context and becomes an interpretation under which all symbol strings are single-meaning. 

\begin{definition}[Effective Interpretation]
An effective interpretation for a language $\mathbb{S}$ is an interpretation under which every symbol string in $\mathbb{S}$ is a single-meaning string. 
\end{definition}



Effective interpretations are often obtained with the help of context, which is a collection of information. Since any piece of information could be represented by a composite observation, a context could be represented by a set of composite observations. Moreover, only actual observations could be used as context, because only them are regarded as true information from some world. 

\begin{definition}[Context]
A context $\mathbb{C}$ is a set of composite observations containing only actual observations. 
\end{definition}

It follows from assumption \ref{AOC} that every context satisfies the strong and weak observer consistency. From another perspective, using a context $\mathbb{C}$ to reduce possible meanings of a symbol string $x$ is an operation that maps $\mathbb{C}$ and the set $V$ of all possible denotations of $x$ to a subset of $V$. Such an operation is called a {\bf context operation}. Since there are various kinds of context information, there are also various kinds of context operations. Detailed study of context and context operations is beyond the scope of this article; we just simply assume their existence and use them like other operations. 

\begin{definition}[Context Operations]
A context operation is a function mapping a context $\mathbb{C}$ and a set $V$ of elements in some cognitive model to a subset of $V$.
\end{definition}

\begin{assumption}
Suppose that $\mathbb{C}$ is a context and $V$ is a set of elements in some cognitive model. Then there is a context operation $h$ such that $h(\mathbb{C}, V) \subseteq V$ is minimal. Moreover, such a minimal set is unique. 
\end{assumption}

\begin{definition}[Interpretation with Context] \label{IWC}
Suppose that $\mathbb{S}$ is a language, $\mathcal{I}$ is an interpretation, $\mathbb{C}$ is a context. Then $\mathcal{I}' = (\mathcal{I}, \mathbb{C})$ is an interpretation for $\mathbb{S}$ such that, for every $x \in \mathbb{S}$, $\mathcal{I}'(x) = h(\mathbb{C}, \mathcal{I}(x))$ is the minimal subset of $\mathcal{I}(x)$ obtained by some context operation $h$. 
\end{definition}


\begin{assumption}
Given any interpretation $\mathcal{I}$ and a language $\mathbb{S}$, there is a context $\mathbb{C}$ such that the interpretation $(\mathcal{I}, \mathbb{C})$ for $\mathbb{S}$ is an effective interpretation. 
\end{assumption}

It is often not easy to find a suitable context operation to make the interpretation of a symbol string minimal. It would be more difficult to find a context and turn an interpretation into an effective one.  These tasks belongs to the general problem of Word Sense Disambiguation (WSD) \cite{Agirre2007, Navigli2009}. Ambiguity is an important phenomenon related to interpretations. An {\bf ambiguity} of the meaning of a symbol string, generally speaking, is a state where this symbol string is supposed to have only one meaning but it could not decide what this unique meaning is. There could be many reasons to cause ambiguities, some of them are explained under the framework of cognitive models.  

The first reason is that, the interpretation is constructed only for abstract strings, but it needs to interpret a concrete string, then there could be an ambiguity of which abstract string this concrete string belongs to. Speech recognition and handwriting recognition are aimed at solving such ambiguities. The second reason is that the language is compositional, and there could be many compositional ways for the symbol string to be interpreted. For example, a sentence has many legitimate parsing trees, each of which corresponds to a different explanation. Ambiguities caused by these two reasons are called {\bf syntactical ambiguity}. 

The third reason is that people do not clearly know what the interpretation for the language is, because the underlying cognitive models or the interpretation itself are not clearly recognized or constructed. In this case and for some symbol strings, people do not know what the set of alternative explanations to do disambiguation. The forth reason is related to context: People knows all those alternative explanations of each symbol string; however, there is no adequate context information, or people do not know how to use the context information to turn the interpretation into an effective one. The ambiguities caused by the  third and forth reason are called {\bf semantic ambiguities}.  



Ambiguities could also occur because of inaccuracies, vagueness or fuzziness of meaning, but this article does not deal with such phenomenons. Similarly, though WSD is an important problem in practice, it is not a topic of this article; we just assume that it is solvable. Human experience shows that such an assumption is reasonable.

\subsection{The Interpretation of Natural Language}


The remaining pages of this article focus on the interpretation of natural language expressions---English expressions in particular---where a {\bf natural language expression} is a meaningful symbol string in natural language. The discussion is made to be as general as possible, so most results could be easily adjusted to expressions in other natural language. 

There is a significant difference between the interpretation of natural language and interpretations for brand-new symbol strings. Interpretations for completely new symbol strings are often just definitions or common agreements among the users, which could be arbitrary in principle. In contrast, natural language expressions commonly have predefined meanings to people, so in most time, their meanings are to be learned or sought, not to be defined. In history, it was often the case that a small number of people had created some expressions and defined their meanings, then those expressions and their meanings had been propagated, learned and accepted by other people. Therefore, interpreting natural language expressions is a task to seek their predefined meanings. 

Then, is the interpretation to be sought unique for a natural language such as English? Obviously it is not. Human experience and the history of language told that the interpretation of natural language had been changing through time \cite{Baugh2002}. Even at the same moment, people with different background could use very different cognitive model or have very different understanding of the meaning of the same expression; such differences would finally result in many personal cognitive models and personal interpretations for natural language. In conclusion, there are many different interpretations for natural language. Then, which interpretation is to be sought and clarified? The following assumption would simplify this problem. 

\begin{assumption}
At any fixed moment $t$, for most natural language expressions, and without considering any context (except the expression itself), there is a unique interpretation commonly accepted by most people using these expressions, which is called the {\bf standard interpretation} $\mathcal{I}_{t}$ at $t$.  
\end{assumption}

This is a reasonable assumption; otherwise, it would be hard to understand why people could communicate with each other. Then how does a standard interpretation handle problems such as: (1) Different people has a different cognitive model; (2) Different people interprets the same expression to different denotations at the same moment? Firstly, the underlying set of cognitive models of the standard interpretation $\mathcal{I}_{t}$ is supposed to include all cognitive models exist at the moment $t$. Secondly, we make the following assumption:

\begin{assumption}
Suppose that at the moment $t$, a natural language expression $x$ is interpreted to the set $A_1$ of denotations by someone $a_1$. Then $A_1 \subseteq \mathcal{I}_{t} (x)$. In other words, all denotations in $A_1$ are possible denotations of $x$ under the standard interpretation $\mathcal{I}_{t}$. 
\end{assumption}


Therefore, a standard interpretation $\mathcal{I}_{t}$ is a really huge interpretation, including all possible denotations assigned by all possible interpretors for each expression. A standard interpretation is often too large to be constructed; it is only defined for theoretical study. Indeed, it never needs to fully construct a standard interpretation, because a context commonly exists to significantly reduce the possible denotations of each expression. In communication, one almost always uses his own cognitive model to do interpretation, unless there are other instructions; so that all denotations to be considered are contained in a few cognitive models in most cases.

The main task of following pages is to seek out the {\bf current standard interpretation} $\mathcal{I}_c$, and to clarify how it is combined with contexts to interpret natural language expressions. $\mathcal{I}_c$ usually does not change very fast, and this article does not specify all its details; so there is no need to fix an accurate time moment. Various dictionaries and corpora, and a number of linguists have provided useful materials to learn $\mathcal{I}_c$. When people have conflicts over the meaning of a natural language expression, they usually appeal to those materials for solutions; however, dictionaries, corpora and linguists only provide incomplete and informal solutions. In contrast, this article seeks a formal solution to $\mathcal{I}_c$ under the framework of cognitive models. 

The solution could not be complete either, because there is too much work to do in a single article. Only some typical categories of expressions and their typical meanings will be discussed to illustrate the general ideas for doing the job. Since natural language is basically compositional, the study of the interpretation $\mathcal{I}_c$ would consists of three parts: (1) the interpretation of primitive expressions, including words and idioms; (2) the interpretation of phrases; (3) the interpretation of sentences. Discourses are left to future study. Since $\mathcal{I}_c$ considers no context, a new interpretation $(\mathcal{I}_c, \mathbb{C})$ would be constructed when a context $\mathbb{C}$ is utilized. When interpreting words and idioms, no context is used; when interpreting phrases and sentences, a context is generally assumed to exist. 

\subsection{Interpretation of Content Words}

Primitive strings of natural language include words and idioms, where an {\bf idiom} is a sequence of several words with denotations not compositional. Idioms defined here is very general, including all phrasal verbs, complex prepositions, and other word collocations with some idiomatic usages, such as `piece of cake'. No matter how complicated, an idiom could be interpreted just like a word when only considering its non-compositional meanings. When considering its compositional meanings, an idiom is interpreted just like a phrase. Therefore, idioms need no special discussion.  


Words are commonly classified into content words and function words \cite{Klammer2009}. Because of their significant difference in meaning, they are discussed separately: This section only considers the interpretation $\mathcal{I}_c$ of content words; function words are left to the next section. Depending on different {\bf parts of speech}, content words could be further classified into nouns, verbs, adjectives, and adverbs \cite{Quirk1985, Jurafsky2008}; however, auxiliary verbs such as `should' and `would', and some adverbs such as `not' and `therefore', are function words. In fact, `...our characterization of parts of speech will depend on their grammatical form and function, rather than on their semantic properties' \cite{Quirk1985}. Moreover, many English words could have different parts of speech, or be converted to words with different parts of speech using suffixes;  words in a language such as Chinese could often be flexibly used as different parts of speech but they never change in form. For these reasons, verbs, adjectives and adverbs are interpreted almost in the same way as nouns when there is no context, and therefore the following only discusses the interpretation of nouns without loss of generality. 

The simplest nouns are proper nouns (proper names), such as `Hamlet' and `Socrates'. Traditional philosophy and classic logic believed that a proper noun should denote a single object \cite{Church1956}; however, this is not true when there is no context, or the interpretation is not effective for this proper noun. 

For example, `Hamlet' could denote different object, process, or even different world in different context. It could denote the book written by Shakespeare around 1600, or one particular book published in 2000 and named `Hamlet'. The book written by Shakespeare around 1600 was obviously a different object to the book printed and published in 2000. `Hamlet' could denote the world imagined by Shakespeare, or a world imagined by a film director who is making a movie named `Hamlet'. It could also denote a particular opera actually performed by some actors in the real world, which is a process, or a particular DVD recording this opera, which is an object. Then, how should we interpret `Hamlet' without any context (under the interpretation $\mathcal{I}_c$)? This article suggests that we should put all those possible denotations of `Hamlet' together to form its interpretation. Therefore, `Hamlet' is a multi-meaning word whose interpretation $\mathcal{I}_c (Hamlet)$ contains all its possible denotations.

Other proper nouns are interpreted similarly. Consider the word `Socrates'. At first glance, `Socrates' seems to have only one denotation in most commonly encountered contexts---the ancient Greek philosopher in the real world; however, this is a misunderstanding. Consider those counterfacturals people often say, for example, `If Socrates were not living in ancient Greece, then he would have not been a philosopher.' Does the word `Socrates' in this sentence denote the ancient Greek philosopher in the real world? Obviously it does not. According to Leibniz's Law, for two things to be identical, all properties of them must be the same. If `Socrates' in the counterfactural denotes the ancient Greek philosopher in the real world, then he must be living in ancient Greece, which contradicts the premise of the counterfactural. In fact, it is commonly accepted that a counterfactural describes things in a world different to the real world \cite {Lewis1973}, and by Leibniz's Law, two things in different worlds could never be identical, no matter how similar they are. Therefore, `Socrates' could denote different person (object) in different context, and hence it is a multi-meaning word whose interpretation $\mathcal{I}_c(Socrates)$ is the set of all persons in all worlds who are named as `Socrates' (If we name other things as `Socrates' that are not persons, then those denotations should also be included). 

Then we discuss common nouns, including several sub-categories: class nouns such as bird, boat, city, piano; collective nouns such as family, school, committee, crew, army; material nouns such as water, sand, light; abstract nouns such as running, red, friendship, love, honesty, courage, beauty, truth, happiness, justice, causation \cite{Noun, Quirk1985}. Each category will be analyzed separately. 

A class noun usually denotes a class of objects in Formal Semantics \cite{Church1956, Allen1995, Jurafsky2008}, so does it in cognitive models: `bird' denotes a class of birds, `boat' denotes a class of boats, etc. Because of the vagueness of natural language, it would be controversial in some circumstances to decide whether an object belongs to an object class denoted by a class noun. This kind of vagueness is a general phenomenon in natural language. 

A collective noun usually denotes a set of object classes in classic logic. For example, the word `family' is supposed to denote the collection $A$ of all particular families, each of which is the set of all members of the family. Then, $A$ is a set of object classes. 
\begin{equation}
A =\{ B =\{b_1, b_2, ... \} : b_1, b_2, ... \textnormal{ are\ members\ of\ a\ family} \}
\end{equation}

In cognitive models, however, we could represent the concept more elaborately. For example, suppose that two persons $X,Y$ had married for one year and then divorced, without any child. Before their marriage and after their divorce, they did not form a family. Therefore, the family formed by $X,Y$ should not be represented by $\{X,Y\}$, but by their lives in the year when they were married. Suppose the start and end moment of the year is $t_1,t_2$. Then the family formed by $X,Y$ should be represented by $\{X',Y'\}$ where 
\begin{equation}
X' = \{ a\in X : t_a \in [t_1,t_2]\}  \quad Y' = \{ a\in Y : t_a \in [t_1,t_2]\} 
\end{equation}
$\{X',Y'\}$ is a process class, and the word `family' therefore denotes a set of process classes. In fact, `family' could denote a relation in some context, just like the word `friendship' analyzed in \ref{SRO}. 

The interpretation of material nouns might be a little controversial. For example, how to interpret the word `water'? This article chooses an extensional approach: A material noun is interpreted as a class of objects or processes that are concrete instances of the concept. For example, `water' is interpreted as the set of all water objects (or call them water processes if you like), where a water object is a set of observations from a particular collection of liquid water that exists within some space region: a drop of water, a glass of water, a bottle of water an ocean of water, etc. Such an interpretation might not meet all people's intuition, but it is simple and constructible. 

The interpretation of abstract nouns might be more controversial, because people usually do not have clear ideas about them, let alone have a common agreement on their meanings. In fact, the interpretation of many abstract nouns are large topics having been discussed and debated for thousand years, such as love, honesty, courage, beauty, happiness and justice. Therefore, this article could only provide some tentative solutions, which exemplify the general approach to interpret abstract nouns under the framework of cognitive models. 

The word `running' denotes the class of all running processes (events), each of which is an instance of running happening in some world. For example, suppose that the person $a$ was running from 6:00 to 6:30 on February 20th, 2016 in the real world. Then there is a running process formed by all primitive observations in the real world, within the time segment from 6:00 to 6:30 on February 20th, 2016, and within the space region occupies by $a$'s body at each moment. Such a running process is an element of the class denoted by `running'. If the constructors of the cognitive model want to include some background information during such a running process, this process could be extended to include more observations from the background.  

The abstract noun `giving' denotes a set of giving events, where a giving event is often an agent $a_1$ giving something $a_2$ to some agent $a_3$. For this reason, Formal Semantics commonly represents `giving' as a ternary relation: the set of sequences $(a_1,a_2,a_3)$ where $a_1,a_2,a_3$ are three objects \cite{Allen1995, Jurafsky2008}. A giving process, however, usually happens during a time segment $[t_1,t_2]$ that is far shorter than the whole lives of $a_1,a_2$ or $a_3$, so it should be represented by the sequence $(a'_1,a'_2,a'_3)$ where for each $i\in \{1,2,3\}$
\begin{equation}
 a'_i = \{ x\in a_i : t_x \in [t_1,t_2] \}
\end{equation}
In this case, $a'_1,a'_2,a'_3$ are three processes. A giving event could also be represented by a single process when there is no need to distinguish between these three parts. For example, a giving event could be simply represented by 
\begin{equation}
a = a'_1\cup a'_2 \cup a'_3
\end{equation}
Instead, if it is necessary, the single process $a$ could be divided into a sequence of more than three processes; for example, to include the motivation of the giver or the reaction of the receiver. How to represent a giving event in a cognitive model depends on the practical goals of the constructors. 

The word `friendship' has been analyzed in Section \ref{SRO}. It denotes a relation between objects, comprised of binary sequences of composite observations in general. Suppose that $A,B$ are friends, starting at time $t_1$ and ending at time $t_2$ without interruption. Then the sequence of processes $<A',B'>$ is an element in the relation denoted by `friendship', where $A' =\{ a\in A : t_a \in [t_1,t_2] \} $ and $B'  = \{ a\in B : t_a \in [t_1,t_2] \}$. Similar to a giving event, the friendship between $A,B$ could also be represented by a single composite observation $A' \cup B'$.  

Finally we consider the word `beauty'. Suppose that it denotes the set $B$ in a cognitive model, then what elements are supposed to be contained in $B$? Formal Semantics commonly interprets `beauty' as a property: the set of all beautiful objects \cite{Allen1995, Jurafsky2008}. Such an interpretation has many problems. The first one is: What if an object only looks beautiful during a period of its lifetime? A thing should not be contained in $B$ when it does not look beautiful. In other words, if $a$ looks beautiful only during time $[t_1,t_2]$, then it is not $a\in B$, but its subset $a'=\{ x\in a : t_x \in [t_1,t_2] \} \in B$. Since an object seldom looks beautiful all over its life, most elements in $B$ should not be objects. Indeed, it is a process that looks beautiful in general, not an object. Then, should elements in $B$ be processes? Common sense tells that whether a thing is beautiful  does not only depends on the thing itself, but also depends on the feelings of people who have observed the thing. Looking at the same object, some people feel beautiful, but other people might not. When someone observes a thing but does not feel beautiful, then his observations from this thing should not be included in the concept of beauty. This shows that, generally speaking, elements in $B$ are not processes, but composite observations: $b\in B$ only includes those primitive observations obtained by people who feel beautiful when observing the process $b'$ where $b\cap b' \subseteq b'$. Moreover, besides observations from a process when it seems beautiful to some person, this person's feelings of beauty caused by the observations of the process is better to be included. Summarizing the above discussion, this article interprets the word `beauty' as a set of composite observations, each of which consists of (1) primitive observations from a process that cause the observer to feel beautiful, and (2) the observers' feelings of beauty (observations learned by their self-consciousness) caused by those primitive observations. Other abstract nouns could be analyzed and interpreted in the same way.  

\subsection{Interpretation of Function Words}\label{IFW}


From the former section, it is not hard to see that a denotation of a content word is either a domain element or a relation. How about a function word? A natural conjecture is that a denotation of a function word is often an operation. 

\begin{conjecture}
Normally, a content word is interpreted (by $\mathcal{I}_c$) to a set of domain elements or relations, and a function word is interpreted (by $\mathcal{I}_c$) to a set of operations. 
\end{conjecture}

This conjecture will follow from the formal definition of content words and function words in the next section. Before the analysis of function words, a more fundamental problem deserves consideration: Why there needs operations to interpret natural language expressions? The main reason is that natural language is compositional: The interpretation of a non-primitive expression is composed by the interpretation of those words that comprise this expression. Besides, for effective communication, the interpretation of any expression should be effective: With enough context, when all words and idioms in an expression are single-meaning, the expression itself should also be single meaning. Such an effective composition is in fact a function (operation): Given a denotation to each word or idiom contained in an expression, the expression has at most one denotation (The expression could have no denotation because the function is probably undefined on the inputs). 

There could be many different kinds of operations, and a function word in an expression is to tell, with the help of context, what operation is to be used in the composition. In many cases, no function word is needed to  indicate the operation being used, because the operation is implicitly implied by the context and the conventions; for example, `red flowers', `smart boys', `clean sweaters', `tall healthy boys', `Tom met Marry yesterday'. Consider the phrase `red flowers'. From the discussion of content words, it is known that the word `red' has a denotation $A$ that is a property, comprised of composite observations, and the word `flower' has a denotation $B$ that is a class  of objects. What is the denotation of `red flowers'?  One intuition tells that `red flowers' denotes the set of flowers that are red at some moment. If a flower $b\in B$ is red at some moment, then there must be an $a \in A$ such that $a \cap b \neq \varnothing$, because $A$ is supposed to contain all red instances. In other words, `red flowers' denotes the set $C$ of objects obtained by the following operation:   
\begin{equation}
 C = \{ b \in B : \exists a \in A\ (a \cap b \neq \varnothing) \} 
\end{equation}
(Another intuition might tell that `red flower' denotes the set of flowers that are red all its life time, then the operation would be different.) 
 

From a logic perspective, each word in `red flowers' is a constant, and the whole phrase is a logic term: a function symbol with these two constants as arguments. However, the function symbol that is supposed to denote the above operation is omitted in this term. Instead, the operation is provided by conventions. For theoretical study, we could supplement `red flowers' with a default function word $\mathbb{F}_0$, and let $\mathbb{F}_0$ denote the above operation. After doing this, the phrase could be interpreted just like a logic term. 

Another example is `tall healthy boys'. In this phrase, the word `boys' is modified by two words `tall' and `healthy' simultaneously. In this case, an arbitrary order could be assigned to them, and the whole phrase could be interpreted in two steps. For example, we could assume that firstly the word `healthy' modifies `boys', and secondly the word `tall' modifies `healthy boys'. After the order is fixed, the operation provided by conventions could be supplemented in each step to obtain the denotation of the whole phrase, just like `red flowers'.  Other phrases without function word could be analyzed in the same way. 

Then we consider how to interpret function words. Although function words play important syntactic roles, here we just consider its semantic interpretation. The main categories of  function words include pronouns, determiners, numerals, conjunctions, prepositions, auxiliary verbs and interjections \cite{Quirk1985, Jurafsky2008}. Different category usually denotes different kinds of operations; they will be discussed separately. (Interjections are exceptions. They just denotes some mental states of the speakers.)



Pronouns include words like `I', `he', `she', `mine', `this', 'these', `which', `what', `who', `where', `when', `that', `something', `somebody', `nothing', `both', etc \cite{Lobeck2013}. Unlike content words, the denotation of a pronoun could not be decided without context. Instead, people could more or less fix a set of alternatives for a pronoun. For example, the word `he' usually denotes a single male animal; the word `which' often denotes some objects or processes mentioned before; etc. A context is used to reduce the range of alternatives, and as the context being more and more adequate, people could finally decide the denotation of the pronoun. This suggests that a pronoun should be interpreted as a context operation mapping a set of alternatives and a context to a subset of these alternatives. 


Determiners include words like `the', `this', `that', `these', `those', `his', `my', `a', `some', `every', `each', `all', `John's', `many', `most', `a few', `enough', `a number of' and `whatever' \cite{Lobeck2013}. Determiners are usually put before nouns to form noun phrases. They could be divided into two kinds. The first kind of determiners occur in phrases such as `the apple', `this book', `these tickets', `that apple', `my son' and `John's eyes'. In such a phrase, the determiner is interpreted just like a pronoun, except that the alternatives are fixed by the noun. For example, consider the phrase `the apple'. The word `apple' denotes a set of objects $A$, the set of alternatives, and `the apple' is supposed to denote a special element of $A$. Without a context, however, there is no way to figure out which element of $A$ is to be denoted by this phrase. Therefore, `the apple' should be interpreted as a context operation to map $A$ and a context $\mathbb{C}$ to a special element in $A$. Similarly, `the apples' denotes a binary operation to map $A$ and a context $\mathbb{C}$ to a special subset of $A$; the word `my' or `mine' denotes a binary operation mapping a set $D$ of alternatives and a context $\mathbb{C}$ to a subset of $D$, where $d$  is the set of all possessions of someone when $d\in D$. 

The second kind of determiners occur in phrases such as  `an apple', `some book', `every book', `all people', `most movies', `a few jobs', etc. In such a phrase, the determiner is commonly interpreted as a {\bf quantifier}. Quantifiers are a kind of operations related to propositions and truth, which will be discussed in detail later. Universal and existential quantifiers have been extensively studied in classic logic \cite {Church1956, Ebbinghaus1984, Enderton2001}. Some kinds of quantifiers, such as those ones denoted by `most', `some' and `a few', often have vague or fuzzy meanings, whose formal discussion needs special analysis that are beyond the scope of this article. A determiner of this second kind could sometimes denote different quantifier in different context. For example, the word `an' is interpreted as a universal quantifier in the sentence `An apple is a fruit', but is interpreted as an existential quantifier or an uniqueness quantifier (there exists one and only one) in `The fruit on the desk is an apple'.

Numerals have been studied by logic and set theory \cite {Church1956, Ebbinghaus1984, Boolos2007, Hrbacek1999}. In most cases, they are interpreted like determiners. Ordinal numbers like determiners of the first kind: The meaning of a phrase such as `the second apple' always needs a context to fix the background ordering, and then to fix the element being referred. Cardinal numbers are like determiners of the second kinds; in fact, there are standard translation of them to first-order logical quantifiers \cite {Ebbinghaus1984, Boolos2007, Hrbacek1999}. A cardinal number in natural language, however, could denote different quantifier in different context. For example, in the sentence `Three people have finished the work', the phrase `three people' could mean `there exists three people', but it could also mean `there are three and only three people that...'. 

Several categories of function words denote a kind of operations called {\bf connectives} in this article: (1) {\bf phrases expressing negation} such as `not', `never', `hardly' and `seldom'; (2) {\bf conjunctions} such as `and', `or', `but', `if', `so that', `because', `considering', `provided', `as if', `in order to', `though' and `whether'; (3) {\bf conjunctive adverbs} such as `besides', `hence', `however', `then', `therefore', `thus' and `meanwhile'; (4) {\bf Modal phrases} (modal auxiliary verbs and some other phrases expressing modality) \label{Mod} such as `will', `shall', `must', `may', `can', `have to', `ought to', `be possible', `be necessary' and `be probable'. Connectives are one of the main topics of logic because they are closely related to propositions and truth \cite {Church1956, Ebbinghaus1984, Blackburn2001}. Classic logic has only studied a small subset of connectives called {\bf truth-functional connectives} \cite {Church1956, Ebbinghaus1984}. Many other kinds of connectives are not truth-functional, which makes them not easy to analysis. Some kinds of non-truth-functional connectives have been studied by modal logic and philosophical logic, but many of them have not been studied yet, or there still lacks common agreement of their formal definitions \cite{Blackburn2001, Blackburn2006, Gabbay2007-7}. This article will introduce a general framework to discuss connectives when interpreting sentences; however, a detailed study will be left to the future. (Note that, in logic, quantifiers and connectives are syntactic symbols to be interpreted as semantic operations; nonetheless, in this article, they are just those operations---elements of cognitive models). 

Prepositions include words like `about',`before', `as', `like', `against', `for', `to', `with', `by', `in', `of', `on',`over', `at', `on' and `without'. They play an important role to form complicated phrases, and could denote a broad range of operations. Since the one-by-one study is too long to be covered in this article, only one example is analyzed here to illustrate the general ideas.


Consider the phrase `a man with a blue tie'. Suppose that the word `man' denotes $A$, the set of male persons; `blue' denotes the property $B$; `tie' denotes $D$, a set of objects. Assume that the context $\mathbb{C}$ interprets two determiners `a' as two existence quantifiers. Then, as discussed before, `blue tie' denotes the set $D_1$ where
\begin{equation}
D_1 = \{ d \in D : \exists b \in B\ (b \cap d \neq \varnothing) \} 
\end{equation}
and `a blue tie' denotes the operation $ \exists d \in D_1(...)$. The preposition `with' is supposed to denote an operation $F(X,Y)$, which is a little complicated. Intuition tells that $F(X,Y)$ is associated with a binary relation $R$, where $(a,b) \in R$ means $a,b$ exist simultaneously and $b$ touches or connects to $a$ in some way. In the phrase `with a blue tie', `a blue tie' modifies `with'. So `with a blue tie' denotes the unary operation $G(X)$ obtained by setting the second argument $Y$ in $F(X,Y)$ as $\exists d \in D_1(...)$
\begin{equation}
G(X)=F(X,\exists d \in D_1(...)) = \{ x\in X :  \exists d \in D_1\ \exists (a,b)\in R\ (a\cap x \neq \varnothing \wedge b\cap d \neq \varnothing ) \} 
\end{equation}
and the phrase `man with a blue tie' denotes the subset $A_1\subseteq A$, where
\begin{equation}
A_1=G(A)=F(A,\exists d \in D_1(...)) = \{ x\in A :  \exists d \in D_1\ \exists (a,b)\in R\ (a\cap x \neq \varnothing \wedge b\cap d \neq \varnothing ) \} 
\end{equation}

The discussion of this example could be generalized to most of other prepositions.

\begin{assumption}
Suppose that $\kappa$ is a preposition denoting an operation $f$. Then, there is a relation $R_f$ associated with $f$ such that the arguments of $f$ satisfy $R_f$.
\end{assumption}




Since the meaning of the word `with' is a little vague, there is in fact a kind of ambiguity (multi-meaning) when interpreting the phrase `a man with a blue tie': Is the man wearing a tie, grabbing a tie, or with it in some other way? It requires additional information in the context $\mathbb{C}$ to solve such an ambiguity. Suppose that $\mathbb{C}$ tells that the speaker only considers wearing a tie, then `with' should be interpreted as `wearing', which denotes a relation $R_1 \subset R$. What the meaning of `man wearing a blue tie'?  It just denotes $A_2$, where 
\begin{equation}
 A_2 = \{ a \in A : \exists d \in D_1\ \exists (x,y)\in R_1\ ( a\cap x \neq \varnothing \wedge d\cap y \neq \varnothing) \}
\end{equation}

From this example one could conjecture that many subordinate clauses, such as `one wearing a tie' and `the person who is wearing a tie', could be interpreted just like prepositional phrases. This would be more clear after the general discussion of the interpretation of phrases in the next section. 




\subsection{Interpretation of Phrases}

This section systematically studies the interpretation of phrases: How the meaning of a phrase is composed by the meaning of words and idioms comprising this phrase. Before this semantic study, it is necessary to define some notations and recall some general syntactic results. 


\begin{definition}[Phrases]
A natural language {\bf phrase} is a sequence of words satisfying the syntactic rules of this language, i.e., it is a syntactically well-formed sequence of words according to these rules. 
\end{definition}

The above definition has assumed the existence of a set of syntactic rules for natural language, which could be found in books such as \cite{Quirk1985, Biber1999, Downing2015, Jurafsky2008}. Unlike the common practice adopted in traditional grammar, phrases defined here are more general: Almost all natural language expressions are phrases. Phrases consist of two categories: {\bf sentences} and {\bf normal phrase}, where a normal phrase is a phrase that is not a sentence. Syntactic difference between sentences and normal phrases has been explored in books such as \cite{Quirk1985} and \cite{Biber1999}; while semantic difference between them would be clarified in the next section. In traditional grammar, phrases only means normal phrases. The reason to make the notation so general is: There is unified syntactic analysis and unified semantic analysis for all syntactically well-formed sequences of words. 

Unified syntactic analyses for phrases had been found and developed in the last century \cite{Hopcroft2007, Jurafsky2008}. No matter what method or algorithm is used, the task of syntactic analysis is to reveal a {\bf dependent structure} (or called {\bf dependent relations}) among all those words in the phrase being analyzed \cite{Quirk1985, Jurafsky2008}. Such a dependent structure is the basis for semantic interpretation. Except some rare cases, all phrases follow a context-free grammar \cite{Jurafsky2008}, and such a grammar has a Chomsky Normal Form \cite{Jurafsky2008, Hopcroft2007}. Thus, a dependent structure could be turned into a binary tree in principle; for this reason, a dependent structure of a phrase is also called a {\bf dependent tree} of this phrase. In a dependent tree, except those leaves, every node is a phrase comprised of two sub-phrases, one modifies the other. The sub-phrase being modified is usually called the {\bf head} of this phrase, and the other sub-phrase is called the {\bf modifier}.  

There are many methods and algorithms to do syntactic analysis, probably with different results  \cite{Allen1995, Manning1999, Jurafsky2008}. None of those methods is perfect, and incorrect dependent trees could often be obtained. Moreover, there could be syntactic ambiguities that will finally result in semantic ambiguities: Without enough context, more than one correct dependent tree could often be found for the same phrase, which generally makes this phrase have more than one meaning. Finally, ellipsis is a common phenomenon in natural language \cite{Quirk1985}, which increases the difficulty of analysis. This article, however, focuses on semantic analysis, and hence puts all those syntactic complications aside: We just simply assume the syntactic analysis has been done successfully.  
 
\begin{assumption}
A unique, correct and complete (without ellipsis) binary dependent tree has been attained for each phrase.
\end{assumption}

If one is familiar with classic logic, it would be helpful to make a comparison between a natural language phrase and a logic term $(f,t_1,...,t_n)$, which is formed by a function symbol $f$ and a sequence of sub-terms $t_1,...,t_n$ with a fixed order \cite {Ebbinghaus1984, Enderton2001}. There are many important differences between them. 

The first difference is: A logic term has to be complete, i.e., an $n$-ary function symbol has to be followed by $n$ terms as arguments \cite {Enderton2001}; however, a phrase could be incomplete, i.e., an $n$-ary function word or idiom could have less than $n$ sub-phrases as arguments to form a new phrase. For example, a single function word with no argument is a phrase that is incomplete, such as `in' and `at'; `in America' is a (incomplete) prepositional phrase because `in' is a binary function word ($\_\_$ in $\_\_$) with only one argument `America'. Just like function words, incomplete phrases could also have arguments to form new phrases, so they are called {\bf function phrases}. For example, `in America' could have `the person' as an argument to form a new phrase `the person in America'. Since it could have at most one argument, `in America' is an unary function phrase. In contrast, complete phrases could have no more argument, and hence they are called {\bf content phrases}. Rigorous syntactic definition of content and function phrases has to be recursive; however, such a definition is omitted, because different natural language has different syntactic rules with many irregularities. Later we will give them a simple semantic definition. 

The second difference is that, unlike function symbols in logic terms, a function word in a natural language phrase (and the operations it denotes) is at most binary, because the dependent tree of any phrase is supposed to be binary. The third difference is that, in classic logic, a sentence is usually not a term  \cite {Enderton2001}; however, a natural language sentence is a phrase by definition. Natural language sentences often have some special syntactic or semantic features that normal phrases do not have; nonetheless, as mentioned before, the syntactic analysis of sentences is the same as normal phrases \cite{Hopcroft2007, Jurafsky2008}, so would be the semantic analysis. 

The fourth difference is that natural language is not as regular as formal language in at least these two aspects. (1) The function word $\mathbb{F}$ supposed to be in a phrase is often omitted, and the operation denoted by $\mathbb{F}$ is provided by conventions; for example, `red flowers', `tall healthy boys', `Tom ran yesterday'. As analyzed before, the word `red' and `flowers' are content words, having denotations $A_1,A_2$ respectively. The phrase `red flowers' denotes a set $B$ of objects, where $B$ is obtained by an operation $f$: $B=f(A_1, A_2)$. However, there is no function word in the phrase `red flowers' to denote the operation $f$ (i.e., it should exist but is omitted), so $f$ could only be provided by conventions. (2) Unlike logic terms, the order of sub-phrases $\mathbb{F}, t_1, t_2$ in a phrase is somehow flexible, and different natural language often have different order.

The last difference is that, unlike logic terms, natural language phrases are usually multi-meaning strings, which makes the interpretation more complicated. If a phrase with a unique dependent tree contains $n$ words or idioms, each of which has $k$ denotations, then considering that some operations are provided by conventions, this phrase would have more than $k^n$ denotations when there is no other restriction. Of course, contexts usually provide enough restrictions to reduce the number of denotations of the phrase. Therefore, unlike logic terms, contexts play an important role in the interpretation of phrases. 



\vspace{8pt}

Having understood all those points, we could finally define the interpretation of phrases, which is the core of the new semantic theory. The following definition probably appears abstract and complicated at first glance; however, the remaining pages of this article would provide many examples to make it concrete. So the reader could briefly study the definition, jump to next sections, and return to it as a summarization. 


\begin{definition}[Recursive Interpretation of Phrases]\label{IP}
Suppose that $u$ is a phrase, and $\mathcal{T}_u$ is its dependent tree. Every node in $\mathcal{T}_u$ is called a {\bf sub-phrase} of $u$. Assume that $\mathbb{C}$ is a context. Then $u$ could be recursively interpreted under the interpretation $\mathcal{I} = (\mathcal{I}_c, \mathbb{C})$ as follows.

\begin{enumerate}
\item Each leaf $x$ of $\mathcal{T}_u$ is a word or an idiom. The set of all denotations of $x$ is $\mathcal{I} (x)$, defined by \ref{ISS} and \ref{IWC}. Suppose that $e \in \mathcal{I} (x)$. Then $e$ is called a {\bf denotation} of $x$ and a {\bf sense} of $x$ under $\mathcal{I}$, and the pair $(x,e)$ is called an {\bf explanation} of $x$. Moreover, we say the explanation $(x,e)$ {\bf implies} the sense $e$ and the denotation $e$, and the sense $e$ {\bf implies} the denotation $e$. 


\item Suppose that $v$ is a node but not a leaf of $\mathcal{T}_u$. Then $v$ is a phrase  comprised of two sub-phrases $x,y$. Assume that $x$ is the modifier, $y$ is the head, and $y$ is {\bf not} a modal phrase (see Page \pageref{Mod} or Section \ref{MCMC}). Suppose that $e_y$ is a denotation of $y$ implied by its sense $s_y$ and its explanation $r_y$. Assume that $e_x$ is a denotation of $x$ implied by its sense $s_x$ and its explanation $r_x$. Let $a\in \{e_x, s_x, r_x\}$. Then there are three cases.



\vspace{6px}

\begin{itemize}


\item Case I: $e_y$ is an unary operation $f_v$. If $f_v(a)$ has been defined, i.e., $a$ belongs to the domain of the function $f_v$, then $v$ has a {\bf denotation} $e_v = f_v(a)$.  

\vspace{6px}

\item Case II: $e_y$ is a binary operation $f_v$. If $f_v(a,\_\_)$ has been defined, i.e., $a$ belongs to the first domain of the function $f_v$, then $v$ has a {\bf denotation} $e_v = f_v(a,\_\_)$, which is an unary operation. 

\vspace{6px}

\item Case III: $e_y$ is a domain element or a relation. Then $e_x$ must be a domain element or a relation. Moreover, there must exist a set $Q_v$ of binary operations provided by conventions, and exist a context operation reducing $Q_v$ to a minimal subset $Q'_v\subseteq Q_v$. Assume that $f_v\in Q'_v$. If $f_v(a,e_y)$ has been defined, i.e., $a$ belongs to the first domain of the function $f_v$ and $e_y$ belongs to the second domain of the function $f_v$, then $v$ has a {\bf denotation} $e_v = f_v(a,e_y)$.

\end{itemize}

\vspace{6px}

In either case, if $e_v$ exists, (1) the phrase $v$ has a {\bf sense} $s_v$ such that,  $s_v=(f_v, s_x,s_y)$ if $a=e_x$, \ $s_v=(f_v, (s_x), s_y)$ if $a=s_x$, \ or $s_v=(f_v,r_x, s_y)$ if $a=r_x$; (2) $v$ has an {\bf explanation} $r_v=( r_x,r_y,(v, s_v))$; (3) the operation $f_v$ is called the {\bf main operation} of $s_v$ and $r_v$, and $v$ is called the {\bf target phrase} of the explanation $r_v$; (4) we say the sense $s_v$ {\bf implies} the denotation $e_v$, and the explanation $r_v$ {\bf implies} $s_v$ and $e_v$.  


\vspace{5pt}

(The case when $y$ is a modal phrase will be discussed at the end of this article. Usually, only $a=e_x$ needs to be considered. When a node $v$ consists of two sub-phrases, both of which denote operations, such as `on to and `as for', then $v$ is regarded as an idiom that denotes a single operation. When the operation $f_v$ introduces some variables, then the variables should be supplemented in suitable places. )




\end{enumerate}
When $u$ is not a leaf of $\mathcal{T}_u$, a denotation of $u$ defined above is called a {\bf composite denotation} of $u$ under the interpretation $\mathcal{I}$. Other denotations of $u$ are called {\bf non-composite denotations} of $u$ under $\mathcal{I}$. $\mathcal{I} (u)$ is the set of all denotations of $u$ under $\mathcal{I}$, which satisfies Definition \ref{ISS}. A {\bf meaning} of the phrase $u$ under the interpretation $\mathcal{I}$ is either one of its denotations, one of its senses, or one of its explanations. 

\end{definition}




\begin{corollary}\label{USD}
An explanation has a unique target-phrase and implies a unique sense. A sense implies a unique denotation. 
\end{corollary}

\begin{corollary}\label{EOCM}
Suppose that $\mathcal{I} = (\mathcal{I}_c, \mathbb{C})$ is an interpretation for a set $ \mathbb{U}$ of natural language phrases, and $u \in  \mathbb{U}$. Assume that the underlying set of cognitive models of $\mathcal{I}$ is $\Delta = \{ \mathfrak{M}_i: i\in I\}$. Then, for every $x$, if $x$ is a denotation or a sense of $u$, there is a cognitive model $\mathfrak{M} \in \Delta$ such that $x\in \mathfrak{M}$ (This notation is defined before Definition \ref{ISS}).  
\end{corollary}


\begin{definition}[Content and Function Phrases]
Suppose that $\mathcal{I} = (\mathcal{I}_c, \mathbb{C})$ is an interpretation for a set $ \mathbb{U}$ of natural language phrases, and $u \in  \mathbb{U}$. Assume that the underlying set of cognitive models of $\mathcal{I}$ is $\Delta = \{ \mathfrak{M}_i: i\in I\}$. If every denotation of $u$ is either a domain element or a relation in some $\mathfrak{M} \in \Delta$, then $u$ is called a {\bf content phrase} under $\mathcal{I}$; if every denotation of $u$ is an operation in some $\mathfrak{M} \in \Delta$, then $u$ is called a {\bf function phrase} under $\mathcal{I}$.  
\end{definition}


\begin{definition}[Effective Interpretation for Phrases]
Suppose that $\mathcal{I} = (\mathcal{I}_c, \mathbb{C})$ is an interpretation for a set $ \mathbb{U}$ of natural language phrases. Then, $\mathcal{I}$ is an {\bf effective interpretation} for $ \mathbb{U}$ if and only if (1) for each $u \in  \mathbb{U}$, $u$ has a unique explanation under $\mathcal{I}$, which implies that it has a unique sense and a unique denotation under $\mathcal{I}$; (2) the underlying set of cognitive models of $\mathcal{I}$ consists of a single cognitive model $\mathfrak{M}$, which is called {\bf the underlying cognitive model} of $\mathcal{I}$.   
\end{definition}

\begin{definition}\label{SEI}
Suppose that $\mathcal{I}$ is an interpretation for a set $ \mathbb{U}$ of natural language phrases. Then, the following set of interpretations is called the set of all effective interpretations for $ \mathbb{U}$ based on $\mathcal{I}$. 
\begin{equation}
[\mathcal{I}] = \{ (\mathcal{I}, \mathbb{C}) : \mathbb{C} \textnormal{ is a context, and } (\mathcal{I}, \mathbb{C}) \textnormal{ is an effective interpretation for }  \mathbb{U}\}
\end{equation}
\end{definition}

Three kinds of meaning have been defined for each phrase: denotations, senses and explanations. In other words, the word `meaning' itself is a multi-meaning word without context. When people understand a phrase, they are supposed to understand its explanations and senses, not just its denotations. The need for explanations has been exemplified before; however, why do we need senses? Unlike explanations, a sense generally does not contain any element in the language being interpreted; unlike denotations, a sense tells how the implied denotation is obtained through a composition procedure from other semantic elements. This special status of senses make them play a central role when interpreting sentences.


In many circumstances, senses alone could explain why different phrases with the same denotation have different meaning. For example, the word `Shakespeare' and the phrase `the man who wrote the book Hamlet around 1600' denote the same person in most contexts; however, people commonly think they have different meaning. These two phrases have different sense under Definition \ref{IP}, which explains why they have different meaning. In some circumstances, a phrase denotes nothing but has a meaning; for example, `a flying horse in the real world'. This is because such a phrase has a sense that implies an empty set of denotations.


When a formal language and its interpretation on a cognitive model are properly defined, the senses and explanations of natural language phrases could be written down as terms or formulas of the formal language, then symbolic reasoning (proof methods) could be studied. In this case, to obtain senses or explanations of phrases becomes a translation from natural language expressions to formal language expressions, which is the main topic studied by Formal Semantics \cite{Allen1995, Jurafsky2008}. Nonetheless, there is a crucial issue rarely mentioned by Formal Semantics: Since natural language expressions are commonly multi-meaning strings but formal language expressions are not, the translation is not one-to-one correspondence, but one-to-many in general. If ambiguities have not been solved, correct translation or reasoning would be hard to achieve. The complete development of a formal language or a disambiguation method, however, is too large and too complicated to be covered in this article. They are left to future research.   

\subsection{Sentences and Truth}



This section begins to discuss the interpretation of natural language sentences. There are four main categories of sentences with different {\bf speech acts}: declaratives to convey information, such as `It rained yesterday.'; interrogatives to seek information, such as `Did yesterday rain?' and `What was the weather yesterday?'; imperatives to instruct somebody to do something, such as `Go out of here!'; exclamatives to express the impression of something, such as `How beautiful the flower is!' \cite{Quirk1985, Downing2015}.  

Almost all functions of natural language are based on the conveyance of information. Moreover, to interpret sentences under the framework of cognitive models, we only consider what and how information has been conveyed. From such an informative perspective, almost every sentence could be converted into some declarative sentence that conveys the same information. For example, when one says `It rained yesterday.', this sentence conveys the information expressed by `I know it rained yesterday.'; when one says `Did it rain yesterday?', it conveys the information expressed by `I do not know if it rained yesterday and I want you to tell me (if you know)'; the imperative `Go out of here!' conveys `I know you are here, but I want you and ask you to go out of here'; `Open the door!' conveys `I think the door does not open, and I want you and ask you to open it';  `How beautiful the flower is!' conveys `I think the flower is very beautiful and I suggest you seeing it'. The conversion from an arbitrary sentence to an equivalent declarative sentence might be different in different context; however, this could be done successfully in principle. 

All those declarative sentences converted above contain a special kind of verbs called {\bf propositional verbs} or {\bf propositional attitude reports}, including words like see, hear, smell, feel, sense, percept, accept, assert, believe, know, command, consider, contest, declare, deny, doubt, enjoin, exclaim, imagine, dream, judge, want, wish, hope, expect, intend, desire, reason, infer, memorize \cite{proatti}. Propositional verbs mainly describe humans' mental activities, often connect subordinate clauses, and will be discussed in a later section. Here we just make the following (informal) assumption. 

\begin{assumption}
The meaning of a sentence with some speech act is equivalent to the meaning of a declarative sentence with some propositional verbs. 
\end{assumption}

With the above assumption, only declarative sentences need to be discussed in the new semantic theory; so for simplicity, `sentence' henceforth means `declarative sentence' if there is no other instruction. As analyzed before, a sentence is a special kind of phrase, which implies two things: (1) The dependent tree of a sentence could be obtained following the same syntactic analysis as a normal phrase \cite{Hopcroft2007, Jurafsky2008}; (2) The meaning (denotations, senses, and explanations) of a sentence could be obtained following Definition \ref{IP}. 

Sentences, however, could be easily distinguished from normal phrases. Besides those syntactic differences \cite{Quirk1985, Downing2015, Jurafsky2008}, it is generally regarded that there is an essential difference between sentences and normal phrases in their meanings: A sentence expresses a complete thought, but a normal phrase does not \cite{Quirk1985, Jurafsky2008}. Such a formulation, however, is too informal for theoretical study;  it is important to introduce a formal characterization of the semantic difference between sentences and normal phrases.

Unlike a normal phrase, when people speak out or write down a sentence in communication, they not only talk about some things, but also assert that the things being talked about are true. What a sentence talks about is characterized by its meanings. Therefore, when people speak out or write down a sentence, they make an assertion that the meaning of this sentence is true under the interpretation $\mathcal{I}$ they have adopted and the context $\mathbb{C}$ they have used. Of course, only the interpretation $\mathcal{I}_c$ will be considered here.

To formalize such an intuition, truth values have to be defined first. There are many choices, depending on different applications. From an epistemic perspective, and without considering any vagueness or degree of uncertainties, this article introduces four truth values: $Z=\{T, F, U,V\}$. $T$ represents $true$, $F$ represents $false$, $U$ represent $undecided$. These three truth values are easy to understand. When different degrees of uncertainties are considered (not in this article), truth values could be defined more complicated, such as to let $Z= [-1,1]\cup \{V\}$. 

The truth value $V$ represents $vacant$. It is introduced mainly because, unlike logical terms, a sub-phrase of a sentence could have no denotation or an empty denotation, which often makes the meaning of the whole sentence vacant. For example, consider the sentence `The flying horse in the real world is beautiful'. The phrase `the flying horse in the real world' is generally regarded to have no denotation. In this case, it seems inappropriate to say that this sentence is true or false, or its truth is undecided. Therefore, a new truth value $V$ is introduced and assigned to sentences containing such a phrase. 


After truth values are defined, the task is to seek out the truth assignment to meanings of natural language sentences under the interpretation $\mathcal{I}_c$, probably with some context $\mathbb{C}$. Three kinds of meaning are defined for each phrase: denotations, senses and explanations (see Definition \ref{IP}). Denotations and senses are pure semantic elements: If $\alpha$ is a denotation or a sense of a phrase, $\alpha \in \mathfrak{M}$ for some underlying cognitive model $\mathfrak{M}$, and $\alpha$ generally does not contain any element in the phrase being interpreted. Therefore, the truth of a denotation or a sense of a sentence is completely determined by the cognitive model it belongs to, independent of the language and the interpretation. In contrast, an explanation of a phrase $u$ consists of a sequence of pairs $(a,b)$ where $a$ is a sub-phrase of $u$, and $b$ is supposed to be the sense of $a$ under the interpretation.  Therefore, the truth of an explanation of a sentence is related to this sentence and the interpretation of this sentence.   

More specifically, senses are at the core of determining the truth assignment. If a denotation is implied by a sense, then it is completely decided by this sense. Meanwhile, a denotation could be implied by many different senses, whose truth values could be different on the same cognitive model. Therefore, it is reasonable to define the truth of a denotation on a cognitive model $\mathfrak{M}$ only when the truth values of all those senses implying the denotation are the same on $\mathfrak{M}$. 
In contrast, if a sense is implied by an explanation, the sense to be true is a necessary condition for this explanation to be true. Moreover, an explanation $\delta$ specifies the sense of each sub-phrase of its target sentence $u$; so for $\delta$ to be true, the interpretation of $u$ must be effective, and each sub-phrase $a$ of $u$ must be interpreted to the sense $b$ under this effective interpretation whenever the pair $(a,b)$ is included in $\delta$. 

The above discussion tells that: (1) truth values should be assigned first to senses of sentences with respect to cognitive models; (2) the truth of denotations and explanations of sentences could then be defined by the truth of their senses. Since sentences have not been formally defined, we will consider a more general assignment to senses of all phrases, and define sentences based no such an assignment afterwards.  






\begin{definition}[Truth Assignment of Senses] \label{TAOF}
Suppose that $\mathcal{I} = (\mathcal{I}_c, \mathbb{C})$ is an interpretation for a set $ \mathbb{U}$ of natural language phrases, and $\Delta= \{ \mathfrak{M}_i: i\in I\}$ is the underlying set of cognitive models of $\mathcal{I}$. Let $\Phi$ be the set of all senses of phrases in $ \mathbb{U}$ under $\mathcal{I}$. Let $Z$ be the set of all truth values. Then, a {\bf truth assignment} $\mathcal{A}_{\mathcal{I}}$ {\bf accompanied with} ${\mathcal{I}}$ {\bf for} $ \mathbb{U}$ is defined as: 

\begin{enumerate}
\item $\mathcal{A}_{\mathcal{I}, \mathbb{U}}: \Delta \times \Phi \rightarrow Z \cup \{ud\}$ is a function from $\Delta$ and $\Phi$ to the set $Z \cup \{ud\}$;

\item When $\mathcal{A}_{\mathcal{I}, \mathbb{U}} (\mathfrak{M}, \phi) = X \in Z$, we say $\phi$ has the truth value $X$ on $\mathfrak{M}$; when $\mathcal{A}_{\mathcal{I}, \mathbb{U}}  (\mathfrak{M}, \phi) =ud$, we say the truth of $\phi$ is undefined on $\mathfrak{M}$;
 
\item When $\phi \notin \mathfrak{M}$, $\mathcal{A}_{\mathcal{I}, \mathbb{U}} (\mathfrak{M}, \phi) =ud$. 
\end{enumerate}

\end{definition} 


\begin{assumption}
Suppose that $\mathcal{I}=(\mathcal{I}_c, \mathbb{C})$ is an interpretation for a set $ \mathbb{U}$ of natural language phrases. Let $\mathcal{I}'=(\mathcal{I}, \mathbb{C}')$ be a new interpretation with a context $\mathbb{C}'$. Let $\Delta, \Delta'$ be the underlying set of cognitive models of $\mathcal{I}, \mathcal{I}'$ respectively. Let $\Phi,\Phi'$ be the set of all senses of phrases in $ \mathbb{U}$ under $\mathcal{I}, \mathcal{I}'$ respectively. Then $\Delta' \subseteq \Delta$, $\Phi' \subseteq \Phi$, and $\mathcal{A}_{\mathcal{I}', \mathbb{U}} = \mathcal{A}_{\mathcal{I}, \mathbb{U}} |_{\Delta' \& \Phi'}$ ($\mathcal{A}_{\mathcal{I}', \mathbb{U}}$ is the restriction of $\mathcal{A}_{\mathcal{I}, \mathbb{U}}$ on $\Delta'$ and $\Phi'$),  
\end{assumption}





The above assumption holds because the context in an interpretation is just to reduce those meanings being considered. As it is analyzed before, the interpretation $\mathcal{I}_c$ is predefined for us to seek out, so does the truth assignment accompanied with it. 

\begin{assumption}
There is a unique truth assignment $\mathcal{A}_c$ accompanied with the interpretation $\mathcal{I}_c$ for the set of all natural language phrases. All other truth assignments are restrictions of $\mathcal{A}_c$. 
\end{assumption}


Sentences could then be formally defined, based on a semantic difference: For a phrase to be a sentence, each of its senses should be assigned to a truth value. 

\begin{definition}[Sentence and Normal Phrase]\label{SP}
Suppose that $\mathcal{I}=(\mathcal{I}_c, \mathbb{C})$ is an interpretation for a set $ \mathbb{U}$ of natural language phrases, and $u\in  \mathbb{U}$. Let $\Delta= \{ \mathfrak{M}_i: i\in I\}$ be the underlying set of cognitive models of $\mathcal{I}$. Let $Z$ be the set of all truth values. Let $\mathcal{A}_{\mathcal{I}, \mathbb{U}}$ be the truth assignment accompanied with $\mathcal{I}$ for $ \mathbb{U}$.  Then, $u$ is a {\bf sentence} under $\mathcal{I}$ if and only if every sense $\phi$ of $u$ under $\mathcal{I}$ is assigned to a truth value $X\in Z$ by $\mathcal{A}_{\mathcal{I}, \mathbb{U}}$ on any $\mathfrak{M} \in \Delta$ where $\phi \in \mathfrak{M}$; A phrase $u$ is a {\bf normal phrase} under $\mathcal{I}$ if and only if every sense $\phi$ of $u$ under $\mathcal{I}$ is assigned to $ud$ by $\mathcal{A}_{\mathcal{I}, \mathbb{U}}$ on any $\mathfrak{M} \in \Delta$. 
\end{definition}

Since the interpretation $\mathcal{I}_c$ is regarded to be unchanged in the discussion, it depends on the context whether a phrase is a sentence, a normal phrase, or not either one. Under an effective interpretation, however, a phrase is either a sentence or a normal phrase. Since an operation could never be a complete thought, and hence could never have a truth value, we have:  

\begin{assumption}
Under any interpretation, a sentence is a content phrase, and a function phrase is a normal phrase. 
\end{assumption}
 
Truth assignments of senses could be expanded to define the truth of denotations and explanations, which is also important in many cases.

\begin{definition}[Truth of Denotation and Explanation] \label{TDAE}
Suppose $\mathcal{I}=(\mathcal{I}_c, \mathbb{C})$ is an interpretation for a set $ \mathbb{U}$ of natural language phrases. Let $[\mathcal{I}]$ be the set of all effective interpretations for $ \mathbb{U}$ based on $\mathcal{I}$ (see \ref{SEI}). Let $Z$ be the set of all truth values.  Let $\mathcal{A}_{\mathcal{I}, \mathbb{U}}$ be the truth assignment accompanied with $\mathcal{I}$ for $ \mathbb{U}$. Let $\Phi$ be the set of all senses of phrases in $ \mathbb{U}$ under $\mathcal{I}$. Assume that $e$ is a denotation implied by some sense $\alpha \in \Phi$, $\delta$ is an explanation whose target phrase is $u\in \mathbb{U}$ (see \ref{IP}). Then,
 
\begin{itemize}
\item Suppose that $\mathfrak{M}$ belongs to the set of underlying cognitive models of $\mathcal{I}$. If $e\in \mathfrak{M}$ and if $\mathcal{A}_{\mathcal{I}, \mathbb{U}}(\mathfrak{M}, \phi)$ is the same for every sense $\phi \in \Phi$ that implies $e$, then $\mathcal{A}_{\mathcal{I}, \mathbb{U}}(\mathfrak{M}, e)=_{df} \mathcal{A}_{\mathcal{I}, \mathbb{U}}(\mathfrak{M}, \phi)$; otherwise, $\mathcal{A}_{\mathcal{I}, \mathbb{U}}(\mathfrak{M}, e)=_{df} ud$.

\item If (1) $\mathcal{I}'\in [\mathcal{I}]$ is an effective interpretation for $u$, and (2) $\delta$ is the explanation of $u$ under $\mathcal{I}'$ (whenever a pair $(a,b)$ is contained in $\delta$ and $a$ is a sub-phrase of $u$, $b$ is the sense of $a$ under $\mathcal{I}'$), then $\mathcal{A}_{\mathcal{I}, \mathbb{U}}(\mathcal{I}', \delta)=_{df} \mathcal{A}_{\mathcal{I}, \mathbb{U}}(\mathfrak{M}, \psi)$ where $\mathfrak{M}$ is the underlying cognitive model of $\mathcal{I}'$ and $\psi$ is the sense implied by $\delta$.

\vspace{5pt}


If condition (1) does not hold, let $\mathcal{A}_{\mathcal{I}, \mathbb{U}}(\mathcal{I}', \delta)=_{df} ud$. If condition (1) holds but (2) does not hold, then $\mathcal{A}_{\mathcal{I}, \mathbb{U}}(\mathcal{I}', \delta)=_{df} F$ if $\mathcal{A}_{\mathcal{I}, \mathbb{U}}(\mathfrak{M}, \psi) \in Z$, and $\mathcal{A}_{\mathcal{I}, \mathbb{U}}(\mathcal{I}', \delta)=_{df} ud$ otherwise. If $\mathcal{A}_{\mathcal{I}, \mathbb{U}}(\mathcal{I}', \delta)\in Z$, we say the explanation $\delta$ has a truth value under $\mathcal{I}'$.




\end{itemize}


\end{definition}


 In practice, people would often talk about the truth of sentences, instead of the truth of their meanings. To talk about the truth of a sentence, it requires that the interpretation for this sentence is effective; otherwise, the sentence would have at least two different senses, whose truth might be different. Therefore, when talking about the truth of a sentence, it is reasonable to make the following assumption.

\begin{assumption}
Whenever the truth of a sentence $u$ is asserted under $\mathcal{I}_c$, a context $\mathbb{C}$ always exists to make the interpretation $\mathcal{I}= (\mathcal{I}_c, \mathbb{C})$ be effective for $u$; whenever interpreting a sentence $u$ under an interpretation $\mathcal{I}= (\mathcal{I}_c, \mathbb{C})$, $\mathcal{I}$ is an effective interpretation for $u$ if there is no other instruction. 
\end{assumption}

Most of those complexities in former discussions are in fact from the consideration of multi-meaning sentences. When the interpretation is effective, things would be much simpler.



\begin{definition}[Truth of Sentence] \label{TSP}
Suppose that $\mathcal{I} = (\mathcal{I}_c, \mathbb{C})$ is an effective interpretation for the sentence $u$. Let $\delta$ be the explanation of $u$ under $\mathcal{I}$. Then $u$ {\bf has a truth value} under $\mathcal{I}$, which is identical to $\mathcal{A}_{\mathcal{I}, u}(\mathcal{I}, \delta)$: the truth value of $\delta$ under $\mathcal{I}$. 
\end{definition}

\begin{corollary}
Suppose that $\mathcal{I} = (\mathcal{I}_c, \mathbb{C})$ is an effective interpretation for the sentence $u$. Let $\mathfrak{M}$ be the underlying cognitive model of $\mathcal{I}$.  Let $\delta$ be the explanation of $u$ under $\mathcal{I}$, and let $\psi$ be the sense implied by $\delta$. Then $\psi$ is the sense of $u$ under $\mathcal{I}$, $\psi \in \mathfrak{M}$, and the truth value of $u$ under $\mathcal{I}$ is identical to $\mathcal{A}_{\mathcal{I}, u}(\mathfrak{M}, \psi)$: the truth value of $\psi$ on $\mathfrak{M}$.

\end{corollary} 


Consequently, the truth of sentences is reduced to the truth of explanations, which is further reduced to the truth of senses plus a check on the interpretation. Since a check on an interpretation is easy, the difficulty is to seek out the truth of senses---to seek out the truth assignment $\mathcal{A}_c$ for senses of sentences. Borrowing notations from other semantic theories, we call a sense of a sentence a {\bf proposition}, and call the denotation implied by a proposition the {\bf content} of this proposition. The remaining pages of this article focus on the study of various kinds of propositions and the determination of their truth. 

\subsection{Atomic Propositions}


Like propositions studied in classic logic, propositions expressed by natural language sentences are also recursively constructed (\ref{IP}), so they could be studied in a similar approach. Firstly, the simplest kind of propositions---{\bf atomic propositions}---are defined and studied, each of which is like a logic predicate without variable. Then, more complicated propositions and their truth could be reduced to atomic propositions. 


Linguists have classified sentences into {\bf simple sentences} and {\bf multiple sentences}. A simple sentence consists of a single independent clause, without any subordinate clause or coordinate clause \cite{Quirk1985}; however, not all simple sentences express atomic propositions. A simple sentence could contain phrases denoting negation or quantifiers, and from a logic perspective, negation and quantifiers make a proposition not atomic. This section, therefore, only considers simple sentences with no phrase denoting negation or quantifiers. The discussion consists of several examples. 


\subsubsection{First Example}

Consider this sentence: `Tom ran at school from 6:00 to 6:30 today'. Suppose the context tells that the world being considered is the real world. Then the phrase `from 6:00 to 6:30 today' denotes a time segment $[t_1, t_2]$ of the real world. The phrase `at school' denotes a space region $Q_t$ of the real world at each moment $t \in [t_1, t_2]$. Both phrases modify the verb `ran', which denotes a set $A$ of running processes. Since the context only considers processes in the real world, the set $A$ is restricted to $A_0$ by a context operation.
\begin{equation}
A_0 = \{a\in A: \forall x \in a\ (w_x = the\ real\ world) \}
\end{equation}
where $w_x$ is the world label of the primitive observation $x$ (see Definition \ref{EO}). Then the phrase `ran from 6:00 to 6:30 today' denotes 
\begin{equation}
A_1 = \{a\in A_0 : t_{a, min} = t_1, t_{a, max} = t_2\}
\end{equation}
where $t_{a, min} , t_{a, max}$ are the start moment and the end moment of the process $a \in A$ (see Definition \ref{CM}). The phrase `ran at school from 6:00 to 6:30 today' denotes  
\begin{equation}
A_2 =  \{a\in A_1 : \forall t \in [t_1, t_2]\ S_{a, t} = Q_t) \}
\end{equation}
where $S_{a, t} $ is the space region of the process $a$ at the moment $t$  (see Definition \ref{CM}). Finally, suppose the word `Tom' denotes the person $b$, given the context. Then what the denotation of the whole sentence `Tom ran at school from 6:00 to 6:30 today'? 

In this sentence, the noun phrase `Tom' and the verb phrase `ran at...' are both content phrases, each of which denotes a domain element or a relation in the underlying cognitive model. To obtain the denotation of the whole sentence, there needs an operation applying to the denotations of these two content phrases. Such an operation could only be provided by conventions, because no function phrase connects these two content phrases. (This belongs to the Case III in Definition \ref{IP}.) Suppose the operation provided by conventions is $f$. Common sense tells that the sentence describes a running process (event) performed by the person $b$, happening during the time segment $[t_1, t_2]$ and within space regions $Q_t$ at each $t \in [t_1, t_2]$. Therefore, the denotation of the whole sentence should be     
\begin{equation}
f(b,A_2) = \{a\in A_2: a\cap b \neq \varnothing \}
\end{equation}
By Theorem \ref{UP}, given a world, a segment of time $[t_1, t_2]$ and a space region $Q_t$ at each moment $t \in [t_1, t_2]$, there is a unique process. Therefore, $A_2$ and $f(b,A_2)$ contains at most one element. If $f(b,A_2) $ is not empty, it contains a single element of $A$, which is the running process (event) described by the whole sentence. 






The above analysis has indeed obtained the proposition expressed by the sentence, although it is a little tedious to formally write it down, and hence is omitted. The content (denotation) of this proposition is either empty, or comprised of a single process that describes an event happening in the real world. 

It is common for natural language sentences to describe events happening in some world, not necessarily in the real world. Under the framework of cognitive models, events are represented by sequences of composite observations in general; and the above example could be generalized to a large category of atomic propositions, each of which describes a single event happening in some world.

\begin{assumption}[Atomic Proposition of Type I, Necessary Condition]
The content of an atomic proposition of type I is either empty or comprised of a sequence of nonempty composite observations. 
\end{assumption}



Then, how to decide the truth of such a proposition? Firstly, suppose that it has a nonempty content. When $f(b,A_2)\neq \varnothing$, for instance, the element contained in $f(b,A_2)$ is a process representing a running event that happens in some world; however, does this event {\bf actually happen} in the world? That is not necessarily true, because the event or part of it could be pure imagination or wrong information: The speaker of the sentence might have only imagined that Tom ran at school from 6:00 to 6:30 today in the real world but lied; he might have only heard about the event but the message was distorted; he might have not seen Tom but made a wrong inference by seeing a person who looked like Tom running at school; etc.

Cognitive models represent such a difference by actual and imaginary observations (see Section \ref{OSO}). General speaking, actual observations are observations whose correctness has been established; imaginary observations are observations whose correctness is doubted, and is left to be checked or verified by actual observations. For example, when an event is represented by a process containing only actual observations, it is regarded as an event actually happening in some world; in contrast, if the process contains imaginary observations, the event being represented is regarded as containing some human imagination or some probably wrong information, instead of actual facts of the world. The new semantic theory, which is based on cognitive models, could be viewed as a theory reducing the truth of natural language sentences to the correctness of primitive observations (minimal pieces of information), since primitive observations are more accurate, elaborate (fine-grain) and fundamental, and their correctness is much more easy to be verified or established. 

When one tells others `Tom ran at school from 6:00 to 6:30 today', for example, he claims the sentence to be true: He claims that the event described by the sentence not only is a running process, but also {\bf actually} happens in the world, not his imagination or some wrong information. For a sentence to be true, therefore, it needs a verification to prove that the event described by the sentence actually happens in the world. Similarly, when one claims a sentence to be false, he claims the event described by the sentence {\bf actually does not}  happen in the world; so for a sentence to be false, it needs some kind of refutation. 

There could be many kinds of verification and refutation on cognitive models, which indeed form a large topic closely related to indirect information, uncertainty, reasoning, cognitive science and other topics of epistemology. Especially, the verification or refutation of an atomic proposition of type I is to establish a relation between actual and imaginary observations in its content or in the cognitive model. A complete exploration of verification and refutation is beyond the scope of this article; here only the simplest and probably the most strict kind is considered, which is called {\bf direct verification} and {\bf direct refutation} respectively. Direct verification or refutation might be inappropriate in many circumstances; however, they are good illustrations of this topic. 

\begin{definition}[Direct Verification and Refutation] 
\ \
\begin{itemize}
\item An imaginary observation $a$ is directly verified by an actual observation $b$ if and only if $a,b$ are the same except that (1) they have different $ac/im$ label $o[3]$; (2) they might have different observer label $o[0]$ (see Section \ref{OSO}).    

\item An imaginary observation $a$ is directly refuted by an actual observation $b$ if and only if $a,b$ are the same except that (1) they have different $ac/im$ label $o[3]$; (2) they might have different observer label $o[0]$; (3) they have different obtaining result $re[0]$ (see Section \ref{OSO} and Definition \ref{PO}). 

\item A composite observation $A$ has been directly verified (refuted) on a cognitive model $\mathfrak{M}$ if and only if every imaginary observation contained in $A$ is directly verified (refuted) by some actual observation in $\mathfrak{M}$. 

\item A sequence of composite observations has been directly verified (refuted) on a cognitive model $\mathfrak{M}$ if and only if every composite observation in this sequence has been directly verified (refuted) on $\mathfrak{M}$.
 
\end{itemize}
\end{definition}

\begin{corollary}
When a cognitive model is given, the truth of a sequence of nonempty composite observations is determined, no matter which proposition it is the content of. 
\end{corollary}

\begin{assumption}[Truth of Atomic Proposition of Type I]\label{TAP}
Suppose that $\phi$ is an atomic proposition of type I whose content consists of a sequence $a$ of nonempty composite observations. Then  %



\begin{enumerate}

\item If $a$ is directly verified on a cognitive model $\mathfrak{M}$, then $\phi$ is true on $\mathfrak{M}$; 

\item If $a$ is directly refuted on a cognitive model $\mathfrak{M}$, then $\phi$ is false on $\mathfrak{M}$; 

\item If $a$ is neither directly verified nor directly refuted on a cognitive model $\mathfrak{M}$, then $\phi$ is undecided on $\mathfrak{M}$. 


\end{enumerate}


\end{assumption}

\begin{assumption}[Empty Content]
When an atomic proposition $\phi$ of type I has an empty content, $\phi$ has the truth value $V$ on a cognitive model it belongs to.  
\end{assumption}
 
As illustrated before, the proposition expressed by `The flying horse in the real world is beautiful' often has an empty content, and hence has the truth value $V$ on most cognitive models, since its sub-phrase `the flying horse in the real world' usually has no denotation in the model. When one has imagined a flying horse in the real world, however, the phrase `the flying horse in the real world' would have a denotation, and the sentence probably has a nonempty content. If the sentence is `the flying horse is beautiful', and if the context considers a world that is not the real world, the sentence is probably true. 
 
When a proposition is assigned $V$ on a cognitive model under an interpretation, it usually indicates that the cognitive model should be extended to include more elements, and the interpretation should be modified to make the content of the proposition nonempty. This is achieved through a learning procedure. Learning on cognitive models is an important topic of future study. Finally, it follows from above definitions plus Assumption \ref{AOC} and \ref{SOCCM} that 
 
\begin{theorem}[Consistency of Atomic Propositions of Type I]
An atomic proposition of Type I is assigned to one and only one truth value on a given cognitive model.
\end{theorem}

\subsubsection{Second Example}

Consider another sentence: `Tom gave me the book'. The word `give' denotes a class of giving events. Each giving event could be represented by a single process, then the interpretation of the sentence is almost the same as the first example; however, this sentence could also be analyzed in a little different way. Similar to what has done in Formal Semantics \cite{Allen1995, Jurafsky2008}, a giving event could be divided into three disjoint parts, and be represented by a ternary sequence of processes $(a_1,a_2,a_3)$: $a_1$ is a giving action of the giver, $a_2$ is an action of the object being given, and $a_3$ is an action of the receiver; all of them are assumed to have the same start and end moments for simplicity. Then, the word `give' denotes a ternary relation $G$ where $(a_1,a_2,a_3) \in G$ represents a giving event. Moreover, suppose the context tells that `me' denotes the person $b$, `Tom' denotes the person $a$, `the book' denotes the object $c$, the world being considered is the real world, the time segment is $[t_1,t_2]$, and for each $t\in [t_1,t_2]$, the space region $Q_{t}$ consists of three disjoint regions $Q_{1,t}, Q_{2,t}, Q_{3,t}$ for objects $a,b,c$ respectively. Then, the relation $G$ could first be restricted by several steps of context operations.
\begin{align}
G_1  & = \{(a_1,a_2,a_3)\in G : t_{a_1, min} =t_{a_2, min} =t_{a_3, min}= t_1, t_{a_1, max} = t_{a_2, max}= t_{a_3, max}= t_2\}  \\
G_2  & =  \{(a_1,a_2,a_3)\in G_1 : \forall t \in [t_1, t_2]\ (S_{a_1, t} = Q_{1,t} \wedge S_{a_2, t} = Q_{2,t} \wedge S_{a_3, t} = Q_{3,t}  ) \} \\
G_3 & = \{(a_1,a_2,a_3)\in G_2: \forall x \in a_1\ \forall y \in a_2\ \forall z \in a_3\ (w_x = w_y = w_z = the\ real\ world) \}
\end{align}
Then, `give me' denotes
\begin{equation}
G_4 = \{(a_1,a_2,a_3)\in G_3: a_2 \cap b \neq \varnothing \}
\end{equation}
and `gave me the book' denotes
\begin{equation}
G_5 = \{(a_1,a_2,a_3)\in G_4: a_3 \cap c \neq \varnothing \}
\end{equation}
Finally, the denotation of `Tom gave me a book' is 
\begin{equation}
G_6 = \{(a_1,a_2,a_3)\in G_5: a_1 \cap a \neq \varnothing \}
\end{equation}
where $G_6$ contains at most one sequence of composite observations. Therefore, this sentence expresses an atomic proposition of type I. 

If a relation in a proposition (such as $G$ in above example) is not unary, and if it has to be restricted by some operation (such as to obtain $G_4, G_5$ from $G_3$), the operation should decide which element in the sequence to be restricted (such as whether to restrict $a_1$, $a_2$, or $a_3$). Variables are then used to represent the position of sequences in a relation when the proposition has to be written down in a formal language. For example, the word `give' denotes a ternary relation, so it is represented by $G(x_1,x_2,x_3)$: a relation symbol $G$ with three variables. Then, the denotation of the phrase `give me' is  
\begin{equation}
f_1(b(x_2), G(x_1,x_2,x_3)) = \{(a_1,a_2,a_3)\in G_3: a_2 \cap b \neq \varnothing \} 
\end{equation}
where $f_1$ is an operation provided by conventions (Case III in Definition \ref{IP}), $b$ is the person denoted by `me', and $b(x_2)$ is to indicate that $b$ modifies or restricts the second element of sequences in $G$. The usefulness of variables would be more clear when discussing quantifiers in the next section. 

In fact, a giving event could be represented by a sequence of more than three elements. For example, it could include a new process $a_4 \subseteq a_1$ to represent the purpose of the giving action of the giver; it could include a process $a_5 \subseteq a_3$ to represent the final influence or result of this giving event; etc. There is no restriction on adding new elements into the sequence to make the representation more refinement, except that the representation would become more complicated. The simplest way, of course,  is to represent an event by a single process or composite observation, only if the representation is adequate for applications. 




\subsubsection{Third Example}


The third example is `Tom and Mary are friends'. This sentence is not equal to `Tom is a friend and Mary is a friend'; instead, Tom and Mary together form a single friendship. Therefore, unlike `Tom and Mary are human beings', the word `and' in this example does not connect two coordinate clauses, but connect two words to form a single phrase. Suppose that `Tom' denotes the person $a$, `Mary' denotes the person $b$. There are two approaches to analyze the sentence (context operations are omitted). The first one is, the phrase `Tom and Mary' denotes a pair $(a,b)$, and `friend' denotes a binary relation $R(x_1,x_2)$. Then, the denotation of the whole sentence could be obtained by an operation $g$:
\begin{equation}
g(\{(a,b)\}(x_1,x_2),R(x_1,x_2)) = \{ (c,d)\in R : (a\cap c \neq \varnothing ) \wedge (b\cap d \neq \varnothing ) \}
\end{equation}

The operation $g$ could be divided into two operations $g^1$ and $g^2$:
\begin{align}
g^1(\{a\}(x_1),R(x_1,x_2)) \   &=\{ (c,d)\in R : a\cap c \neq \varnothing  \} = R_1  \\
g^2(\{b\}(x_2),R_1(x_1,x_2))  &= \{ (c,d)\in R_1 : b\cap d \neq \varnothing \} 
\end{align}

The second approach is to let the phrase `Tom and Mary' denotes the set $A=\{a,b\}$, and to let the word `friend' denote a property $C$ where 
\begin{equation}
C = \{ c: c= a \cup b \wedge (a,b)\in R\}
\end{equation}

Then, the denotation of the sentence `Tom and Mary are friends' is 
\begin{equation}
f(A(x_1), C(x_1)) = \{c\in C: \forall x\in A\ (x \cap c \neq \varnothing ) \} \end{equation}

In this example, $g^1, g^2$ and $ f$ could be classified as the same kind of operation. Similar sentences are like `These three persons own the company together' and `The family own the large house (together)'. In these sentences, a group of things do something or are in a relationship together, not separately. 

\subsubsection{Fourth Example}

The forth example is `Tom finished the homework'. Suppose that `Tom' denotes the person $a$, `the homework' denotes a process $b$. The word `finish' denotes a binary relation $R(x_1,x_2)$. It seems that the denotation of the whole sentence could be obtained by two steps:   
\begin{align}
h^1(\{a\}(x_1),R(x_1,x_2)) \  &=\{ (c,d)\in R : a\cap c \neq \varnothing  \} = R_1  \\
h^2(\{b\}(x_2),R_1(x_1,x_2)) &= \{ (c,d)\in R_1 : b\cap d \neq \varnothing \} 
\end{align}

Nonetheless, the second step is inappropriate. The formula $b\cap d \neq \varnothing $ means the having-been-finished task $d$  overlaps with the homework $b$, which would be true when a part of the homework $b$ has been finished, not all of it; however, common sense tells that the sentence means the person $a$ has finished the whole homework $b$, not just a part of it. Therefore, instead of using $b\cap d \neq \varnothing $ in the operation $h^2$, we should use $b=d$ or $b \cap d =b$. In other words, we should let
\begin{align}
h^2(\{b\}(x_2),R_1(x_1,x_2)) &= \{ (c,d)\in R_1 : b= d \}  \quad or   \\
h^2(\{b\}(x_2),R_1(x_1,x_2)) &= \{ (c,d)\in R_1 : b\cap d =b \} 
\end{align}

Generally speaking, the condition $x= y$ and $x \cap y =x$ are stronger than $x \cap y \neq \varnothing$. In this article, $x= y$ is called an {\bf exact match}, $x \cap y =x$ a {\bf strong match}, and $x \cap y \neq \varnothing$ a {\bf weak match}. There could be other kinds of match. Which kind of match is used in the operation depends on the relations being operated and depends on the representation method. Usually, strong match and weak match only apply to composite observations. 


\subsubsection{Basic Operations and Atomic Propositions}
 
All propositions discussed in former examples are formed by the same kind of operations: {\bf basic operations}.

\begin{definition}[Basic Operation]\label{BO}
A basic operation is a binary operation $f$ with two arguments, $A(x_i)$ and $R(x_1,..., x_k)$, where $A$ is a set of domain elements, $R$ is a $k$-ary relation, and $x_1,..., x_k$ are variables indicating the position of sequences in $R$. $A$ is called the range of the variable $x_i$ and the domain of the operation $f$. Moreover, depending on $A$ and $R$ (only illustrating the three most commonly used kinds of match),
\begin{align}
f(A(x_i), R(x_1,..., x_k)) &= \{(b_1,...,b_k)\in R: \forall a\in A\ (a \cap b_i \neq \varnothing) \} \quad or \\
f(A(x_i), R(x_1,..., x_k)) &= \{(b_1,...,b_k)\in R: \forall a\in A\ (a \cap b_i = a) \} \quad \ or \\
f(A(x_i), R(x_1,..., x_k)) &= \{(b_1,...,b_k)\in R: A = b_i \}
\end{align}
When the set $A =\{a\}$ containing only one element, $A$ could be simply written as $a$. The variable $x_i$ is said to be {\bf bounded} (not free) by $f$ in the sense $(f, A(x_i), ...)$. 
\end{definition}


\begin{assumption}
Basic operations are elements of every cognitive model.
\end{assumption}

\begin{definition}[Atomic Propositions]
A sense $\phi$ is an atomic proposition if and only if (1) the denotation implied by $\phi$ is either empty or comprised of a single element of a relation $R$, and (2) if $R$ is a $k$-ary relation $R(x_1,..., x_k)$ ($k>0$), then each $x_i$ ($i\leq k$) is bounded by a basic operation in $\phi$. 
\end{definition}


\begin{corollary}
There is no free variable in an atomic proposition.
\end{corollary}

It is not hard to see that any atomic proposition $\phi$ could be written as $\phi = (f, \alpha, \beta)$, where $f$ is a basic operation, $\alpha, \beta$ are two senses of some phrases. The content of $\phi$ is simply $f(\alpha, \beta)$. If only basic operations are considered, an atomic proposition $\phi$ could be regarded as a $k$-ary relation $R(x_1,..., x_k)$ being operated by $k$ basic operations $f^1, ..., f^k$ with domains $A_1, ..., A_k$ respectively. Thus, $\phi$ could be written as $R(A_1,..., A_k)$:
\begin{equation}
\phi = R(A_1,..., A_k) = (f^1, A_1(x_1), (f^2, A_2(x_2), ..., (f^k, A_k(x_k), R(x_1,..., x_k)\dots)
\end{equation}
The content of $\phi$ could be written as
\begin{equation}
|R(A_1,..., A_k) | = f^1(A_1(x_1), f^2(A_2(x_2), ..., f^k (A_k(x_k), R(x_1,..., x_k)\dots)
\end{equation}

\begin{definition}[[Atomic Proposition of Type I]
An atomic proposition of type I is an atomic proposition whose content is either empty or comprised of a single sequence of nonempty composite observations, and whose truth is decided by the method presented in Assumption \ref{TAP}.
\end{definition}

\subsubsection{Fifth Example}

There are other types of atomic propositions. Consider the sentence `Tom is Mike'. Common sense tells that this sentence means these two names `Tom' and `Mike' denote the same person. Suppose that the word `is' denotes the equivalent relation $\equiv$, `Tom' denotes the person $a$, `Mike' denotes the person $b$. Then, without considering any context operations, the sense expressed by this sentence is 
\begin{equation}
\beta=(f^1,a(x_1),(f^2,b(x_2), \equiv (x_1,x_2)))
\end{equation}
where $f^1,f^2$ are two basic operations:
\begin{align}
f^2(b(x_2), \equiv (x_1,x_2)) & = \{(b_1,b_2)\in \ \equiv: b = b_2 \} = R \\
f^1(a(x_1), R (x_1,x_2)) \ & = \{(b_1,b_2)\in R: a = b_1 \} 
\end{align}
If $(a,b) \in \ \equiv$, the denotation of this sense is $\{(a,b)\}$; otherwise it is empty. Therefore, $\beta$ is an atomic proposition. In classic logic, this proposition is simply written as $a\equiv b$. 

The crucial problem is how to decide the truth of $\beta$. It appears that the method presented in Assumption \ref{TAP} is inappropriate: Whether $a\equiv b$ on a cognitive model is completely determined by whether they are the same set, unrelated to whether the primitive observations contained in them are imaginary or actual. In other words, no matter $a,b$ are actual persons or imaginary persons, if $a, b$ are the same set, then $\beta$ is true. This is in fact the truth definition adopted by classic logic \cite{Church1956, Ebbinghaus1984, Enderton2001}, although the formulation is a little different.

\begin{assumption}[Truth of Atomic Proposition of Type II]\label{TPMR}
Suppose that $\mathfrak{M}$ is a cognitive model, and $\phi \in \mathfrak{M}$ is an atomic proposition of type II whose content is $A$. Assume that $\phi$ has the form $(f, \alpha, \beta) $, where $f$ is a basic operation, $\alpha, \beta$ are two senses. Let $B,C$ be the denotations implied by $\alpha, \beta$ respectively. Then, 

\begin{enumerate}

\item If $A \neq \varnothing$, then $\phi$ is true on $\mathfrak{M}$; 

\item If $A = \varnothing$ but $B, C \neq \varnothing$, then $\phi$ is false on $\mathfrak{M}$; 

\item If $B = \varnothing$ or $C = \varnothing$ (which implies $A = \varnothing$), then $\phi$ is vacant on $\mathfrak{M}$.

\end{enumerate}

\end{assumption}

\begin{definition}[Atomic Proposition of Type II]
An atomic proposition of type II is an atomic proposition whose truth is decided by the method presented in Assumption \ref{TPMR}.
\end{definition}

Unlike type I, the truth of atomic propositions of type II is completely decided by the belong-to relation between elements in the cognitive model. Since the belong-to relation in a cognitive model is constructed without uncertainty---an element either belongs to a set or not---no proposition of this type has the $undecided$ truth value.  Finally, we have the following rules for propositions in general. 

\begin{assumption}[Vacant of Propositions] \label{VOP}
Let $f$ be an operation in the proposition $\phi\in \mathfrak{M}$, and let $\alpha$ be an argument of $f$. Suppose that (1) $\alpha$ is the sense of some normal phrase and $\alpha$ implies an empty denotation, or (2) $\alpha$ is a proposition having the truth value $vacant$ on $\mathfrak{M}$. Then $\phi$ is assigned to the truth value $vacant$. 


\end{assumption}

\begin{assumption}
If a proposition $\phi$ is assigned to the truth value $vacant$, $\phi$ implies an empty denotation. The converse does not hold in general. 
\end{assumption}

\subsection{Propositions with Quantifiers}



After the discussion of atomic propositions, the study on more complicated propositions is a routine work that parallels to what has done in classic logic: More complicated propositions could be recursively reduced to atomic propositions \cite{Church1956, Ebbinghaus1984, Enderton2001}. Complicated propositions are formed by various kinds of operations, where two kinds are especially important: {\bf quantifiers} and {\bf connectives}. Classic logic has studied truth-functional connectives, existential and universal quantifiers \cite{Church1956,  Ebbinghaus1984, Enderton2001}; philosophical logic has studied some categories of non-truth functional connectives \cite{Blackburn2001, Blackburn2006, Gabbay2007-7}. This article aims at providing a general and unified framework based on cognitive models for the study of quantifiers and connectives, leaving the complete study to future. This section focuses on quantifiers, which are commonly denoted by determiners in noun phrases. 
 


Consider the sentence `All trees turned green'. The main verb `turn' in this sentence describes a set $G$ of changing events. Each changing event could be represented by a single process, or by a sequence of two states where the second state is the result of the change of the first state. We adopt the two-states approach. To represent the event of a tree turning green, for example, the first state is a state of the tree at some moment when it is not green, and the second state is a state of the tree at a later moment of the same year when it is green. Therefore, $G$ is a binary relation written as $G(x_1,x_2)$. Moreover, assume the context tells that the world considered is still the real world, all first states in $G$ happen at some moment $t \in [t_1, t'_1]$, within space region $Q_1$, and all second states in $G$ happen at $t \in [t_2, t'_2]$, within space region $Q_2$ (Since the sentence is talking about many events simultaneously, the context only provides the range of the time and space). Using context operations, $G$ is restricted to the set $G_1$ of changing events:
\begin{equation}
\begin{split}
G_1  & = \{(a_1,a_2)\in G : t_{a_1}\in [t_1, t'_1], t_{a_2} \in [t_2, t'_2], S_{a_1, t_1} \subseteq Q_{1}, \\ 
& \quad \quad S_{a_2, t_2} \subseteq Q_{2}, \forall x \in a_1\ \forall y \in a_2\ (w_x = w_y =  the\ real\ world)\}  
\end{split}
\end{equation}
Unlike atomic propositions, $G_1$ usually contains more than one element. Suppose that `green' denotes the property $A$. Then the phrase `turned green' denotes the relation
\begin{align}
G_2 &= \{(b,c)\in G_1: \exists a\in A\ (a \cap c \neq \varnothing) \} \quad or \\
G_2 &= \{(b,c)\in G_1: \exists a\in A\ (a \cap c =c) \}
\end{align}
The first equation means some parts of the final state are green, and the second means all parts are green. Which equation is used depends on the context. 

No matter which equation is used, the set $G_2$ could be viewed as being obtained by a binary operation $f_0$ such that $G_2=f_0(A(x_2), G_1(x_1,x_2))$, where $f_0$ is provided by conventions since there is no phrase in the sentence denoting this operation (Case III in Definition \ref{IP}). The sense of `turned green' could be written as $(f_0, A(x_2), G_1(x_1,x_2))$. As the reader will see, $f_0$ is in fact the existential quantifier $f_{\exists}$ defined later. 



Then, suppose that `trees' denotes a class $B$ of objects. Common sense tells that the word `all' denotes the universal quantifier, written as $f_{\forall}$ in this article. The universal quantifier $f_{\forall}$ is a binary operation having two arguments $B$ and $G_2$. If analyzed step by step, the phrase `all trees' denotes an unary operation $f_{\forall} (B, \_\_)$  (Case II in Definition \ref{IP}), and the whole sentence denotes (Case I in Definition \ref{IP}):
\begin{equation}
f_{\forall} (B(x_1), G_2(x_1,x_2)) =  \{(a,c)\in G_2: \exists b\in B\ (a \cap b \neq \varnothing) \}
\end{equation}

If context operations are omitted, the proposition expressed by the sentence could be written as 
\begin{equation}
\beta=(f_{\forall} ,B(x_1), (f_{\exists}, A(x_2), G_1(x_1,x_2)))
\end{equation}

The above example shows how to obtain the denotation and the sense of a phrase where the sense contains quantifiers. A proposition with quantifiers commonly talks about many things (such as many events) simultaneously, and its denotation is supposed to contain all those things being talked about.  This consideration generates the following definition. 

\begin{definition}[Quantifier]
A quantifier is a binary operation $f$ with two arguments, $A(x_i)$ and $R(x_1,..., x_k)$, where $A$ is a set of domain elements, $R$ is a $k$-ary relation, $x_1,..., x_k$ are variables indicating the position of sequences in $R$. $A$ is called the range of the variable $x_i$ and the domain of the quantifier $f$. Moreover, depending on $A$ and $R$ (only illustrating the three most commonly used kinds of match),
\begin{align}
f(A(x_i), R(x_1,..., x_k)) &= \{(b_1,...,b_k)\in R: \exists a\in A\ (a \cap b_i \neq \varnothing) \} \quad or \\
f(A(x_i), R(x_1,..., x_k)) &= \{(b_1,...,b_k)\in R: \exists a\in A\ (a \cap b_i = a) \} \quad or \\
f(A(x_i), R(x_1,..., x_k)) &= \{(b_1,...,b_k)\in R: \exists a\in A\ (a = b_i) \}
\end{align}
When the set $A =\{a\}$ containing only one element, $A$ could be simply written as $a$. The variable $x_i$ is said to be {\bf bounded} (not free) by $f$ in the sense $(f, A(x_i), ...)$.  
\end{definition}


Then, how to decide the truth of propositions with quantifiers? The approach is almost the same as what has done in classic logic \cite{Enderton2001}. Some notations have to be defined first.

\begin{definition}[Formula]
A formula is obtained from a proposition deleting some quantifier symbols and their domains, and making the correspondent variables free. When a formula $\alpha$ has free variables $x_1,...,x_m$, it is written as $\alpha[x_1,...,x_m]$. Note that propositions have no free variable. 
\end{definition}

For example, consider the proposition $\beta= (f_{\forall} ,B(x_1), (f_{\exists}, A(x_2), G_1(x_1,x_2)))$. If we delete the quantifier symbol $f_{\forall}$ and its domain $B(x_1)$ in $\beta$, we obtain the formulas $ (f_{\exists}, A(x_2), G_1(x_1,x_2))$; if we delete $f_{\exists}$ and $A(x_2)$, we obtain the formula $(f_{\forall} ,B(x_1), G_1(x_1,x_2))$; If we delete all these four elements, we obtain the formula $ G_1(x_1,x_2)$. 

\begin{definition}[Value Assignment]
A value assignment $\sigma$ is a function mapping each variable to an element in some cognitive model.  
\end{definition}

For simplicity, the following only considers formulas with only one free variable. The more general case could be handled by a routine work. 

\begin{definition}[Truth under Value Assignment]
Let $\sigma$ be a value assignment. let $\alpha[x_i]$ be a formula with a single free variable $x_i$. Suppose $\mathfrak{M}$ is a cognitive model, $\alpha[x_i] \in \mathfrak{M}$, $\sigma(x_i) \in \mathfrak{M}$ and $\sigma(x_i) \neq \varnothing$. Then the truth of the formula $\alpha[x_i]$ on $ \mathfrak{M}$ and $\sigma$ is identical to the truth of the proposition $ (f, \sigma(x_i), \alpha[x_i])$ on $\mathfrak{M}$, where $f \in \mathfrak{M}$ is a {\bf basic operation} (see Definition \ref{BO}).  
\end{definition}

Then, the truth of propositions with quantifiers could be defined. Different quantifier would have different rules to decide the truth.

\begin{definition}[Truth Definition of Universal Quantifier]
Let $f_{\forall}$ be the universal quantifier, and let $\alpha[x_i]$ be a formula with a single free variable $x_i$. Then the proposition $\beta = (f_{\forall}, A(x_i), \alpha[x_i])$ is true on the cognitive model $ \mathfrak{M}$ if and only if $A\neq \varnothing$ and {\bf for every} value assignment $\sigma$ where $\sigma(x_i) \in A$, the formula $\alpha[x_i]$ is true on $ \mathfrak{M}$ and $\sigma$.  
\end{definition}

\begin{definition}[Truth Definition of Existential Quantifier]
Let $f_{\exists}$ be the existential quantifier, and let $\alpha[x_i]$ be a formula with a single free variable $x_i$. Then the proposition $\beta = (f_{\exists}, A(x_i), \alpha[x_i])$ is true on the cognitive model $ \mathfrak{M}$ if and only if {\bf there is} a value assignment $\sigma$ such that $\sigma(x_i) \in A$ and the formula $\alpha[x_i]$ is true on $ \mathfrak{M}$ and $\sigma$. 
\end{definition}

The above discussion could be easily generalized to other quantifiers.  For example, consider the sentence `Most trees turned green'. Suppose the word `most' denotes the quantifier $f_{most}$. Then the proposition expressed by this sentence is 
\begin{equation}
(f_{most} ,B(x_1), (f_{\exists}, A(x_2), G_1(x_1,x_2)))
\end{equation}

Common sense tells that the following definition is reasonable. 

\begin{itemize}
\item The proposition $\beta = (f_{most}, A(x_i), \alpha[x_i])$ is true on $ \mathfrak{M}$  if and only if $A\neq \varnothing$ and there is an $A' \subseteq A$ such that $A'$ contains {\bf most} elements of $A$ and for every value assignment $\sigma$ where $\sigma(x_i) \in A'$, the formula $\alpha[x_i]$ is true on $ \mathfrak{M}$ and $\sigma$. 
\end{itemize}

Of course, the above definition does not provide an exact rule for people to decide the truth of propositions with the quantifier $f_{most}$, because the quantifier $f_{most}$ (the meaning of the word `most') is vague intrinsically. How to handle vague meaning or vague operations is beyond this article. Sometimes, we could give a simple accurate characterization of `most'. For example, in some context, `most' means `more than 90\%'. Then the above definition turns to be exact: 

\begin{itemize}
\item The proposition $\beta = (f_{most}, A(x_i), \alpha[x_i])$ is true on $ \mathfrak{M}$  if and only if $A\neq \varnothing$ and there is an $A' \subseteq A$ such that $A'$ contains {\bf more than 90\%} elements of $A$ and for every value assignment $\sigma$ where $\sigma(x_i) \in A'$, the formula $\alpha[x_i]$ is true on $ \mathfrak{M}$ and $\sigma$. 
 
\end{itemize}


For a better understanding of quantifiers defined in this article, it is useful to make a comparison between them and the quantifiers defined in classic logic. In classic logic, quantifiers are syntactic symbols in the formal language; in this article, quantifiers are operations in a semantic model (a cognitive model). Of course, some formal symbols have to be introduced to denote those operations; however, those symbols belong to the metalanguage, not the objective language being studied. Moreover, unlike a formal language used by maths or science, it is rare to find words as variables in natural language. Instead, variables belongs to the metalanguage, only introduced to interpret natural language phrases. 

The second difference is: A quantifier in classic logic does not need to specify a domain for it, since the semantic model generally has only one domain; in contrast, a quantifier denoted by a natural language phrase commonly needs a set of elements to be its domain, because cognitive models are multiple-domain models. The domain of a quantifier is supposed to be one of its argument.  Although multiple-domain models could be reduced to single-domain models in principle \cite{Enderton2001}, they are more natural and convenient to be used when interpreting natural language expressions. 


The third difference is: There are only two kinds of quantifiers in classic logic ($\exists$, $\forall$); however, many other kinds of quantifiers could be denoted by natural language phrases. These quantifiers are often denoted by determiners or cardinal numbers such as `a/an', `three', `these', `all', `some', `a few', `a number of', `thousands of', `many', `most' and `almost all'. Some of those quantifiers could be defined by the universal and existential quantifier, but some could not; some have precise meaning, but some are inaccurate or vague. Vague quantifiers are denoted by phrases such as `a few', `many', `a number of', `most' and `almost all'. Moreover, a function phrase could denote different quantifier in different context. For example, `a/an' could denote the universal quantifier or the existential quantifier, or means `there exists one and only one...'. Similarly, a cardinal number such as `three' could mean `there exists three things in the domain such that...', or `there exists three and only three things ...'. 


The last difference is: Ellipsis is a general phenomenon in natural language; so the phrase supposed to denote a quantifier or the domain of a quantifier is omitted sometimes, which have to be supplemented by conventions or contexts. One example has been illustrated before (`turn green'). There is another example: `Prices rose these days'. In this sentence, `prices' has no determiner before it; nonetheless, common sense tells that a quantifier is needed there because `prices' is a plural that talks about more than one price simultaneously. To interpret such a sentence, human conventions often provide a set of possible quantifiers: It could be `all prices', `some price', `most prices', or others. Then, the context selects the most probable one (Case III in Definition \ref{IP}). When the context fails to determine the quantifier, there would be an ambiguity: Whether the speaker means all prices rose, most prices rose, some prices rose, or something else? The phrase supposed to denote the domain of a quantifier could also be omitted; for example, `There are a lot of flowers on the tree. All are red'. In this example, the word `all' is a determiner denoting the universal quantifier; however, there is no phrase denoting the domain of this quantifier. It is the context that tells the domain of the quantifier is the set of flowers on the tree. 

\subsection{Multiple Clauses I}\label{MCI}



A natural language sentence could have more than one clause. Multiple clauses in a sentence are often connected by function words, which could be omitted sometimes. When two clauses being connected are at the same level, they are called {\bf coordinate clauses}. Coordinate clauses are usually connected by words or phrases like `and', `or', `so', `but', `both..., and...', `neither..., nor...', and `either..., or...'. When one clause forms a component of another clause, it is called a {\bf subordinate clause}. Subordinate clauses are often connected by function phrases like `when', `where', `what', `that', `if...then', and `so that'. Subordinate clauses could be subject, object, complement, adverbial, or other components in another clause. When two clauses are connected by a function word $ \xi$ to form a new clause $u$, $\xi$ is called a {\bf connector} or a {\bf clause linker}.  \cite{Quirk1985} 

The study on multiple clauses in this article is not based on their syntactic difference, but mainly on their semantic distinction. From a semantic perspective, no matter how complicated the syntactic structure of a multiple sentence could be, there are only two cases.  Suppose that a sentence $u$ is formed by two clauses that are connected by a clause linker $\xi$. Under the interpretation $\mathcal{I}$, $u$ expresses the proposition $\phi$ and $\xi$ denotes the operation $f_{\xi}$ (If $\xi$ is omitted, $f_{\xi}$ is provided by conventions and context). Then one of the following cases happens.

\begin{enumerate}
\item Case I: $f_{\xi}$ is the main operation of $\phi$; for example, `Tom went to the theater and Mary went too', `When he came back, I had finished my homework'  

\item Case II: $f_{\xi}$ is not the main operation of $\phi$; for example, `The movie I saw was terrific', `I know that he is lying'.  
\end{enumerate}

This and the next two sections focus on the second case, where $u$ is comprised of a main clause and a subordinate clause. This section studies a special sub-case where the subordinate clause could be interpreted just like a normal phrase. The discussion consists of several examples. 


\subsubsection{First Example}

Consider the sentence `Those mountains that are located in California are high'. Suppose the idiom `be located in' denotes a binary relation $R$. Common sense tells that $R$ is a relation between an object and the space regions it occupies. Formally speaking, 
\begin{equation}
R=\{ (x,y) : \textnormal{$x$ is an object, $y$ is a space region, and } \forall t \in [t_{x,min}, t_{x,max}]\ S_{x,t} \subseteq y\} 
\end{equation}
where $t_{x,min}$ is the start moment of the object $x$, $t_{x,max}$ is the end moment of $x$, and $S_{x,t}$ is the space region occupied by $x$ at the moment $t$ (see Definition \ref{CM}). Suppose that the context tells that the time being considered is $[t_1, t_2]$, the world is the real world. Then $R$ has to be restricted to
\begin{equation}
 R_1 = \{ (x,y)\in R :  t_{x,min} \geq t_1, t_{x,max} \leq t_2, w_x =  the\ real\ world\}
 \end{equation}
Assume that `California' denotes the space region $S_0$ in the real world. Then `are located in California' denotes the relation 
\begin{equation}
R_2 = \{ (x,y)\in R_1 :  y \subseteq S_0 \} 
\end{equation}
Suppose that `mountains' denotes a class $A$ of objects. The word `that' in the phrase `mountains that are located in California' is a clause linker, which is supposed to denote a binary operation $f$. Then, `that are located in California' denoted the unary operation $f(R_2(x_1,x_2),\_\_)$, and `mountains that are located in California' denotes
\begin{equation}
A_1 =f(R_2(x_1,x_2),A(x_1))= \{a \in A : \exists (x,y)\in R_2\ (a \cap x \neq \varnothing) \}
\end{equation}


As analyzed before, the determiner `those' denotes a context operation, and `those mountains that are located in California' denotes a set $A_2 \subseteq A_1$ fixed by the context. Finally, suppose that the word `high' denotes the property $H$, and assume that the main operation of the whole proposition is a universal quantifier, provided by conventions and fixed by the context. In other words, the whole sentence is equal to `All those mountains that are located in California are high' in the given context. Then, the denotation of the whole sentence is
\begin{equation}
H_1 = f_{\forall} (A_2(x_1),H(x_1))= \{ b\in H : \exists a\in A\ (a\cap b \neq \varnothing \}
\end{equation}

 It is easy to see that the above analysis also applies to the sentence `All those mountains located in California are high', although in this sentence the operation $f$ is provided by conventions, not denoted by the word `that'. Other sentences with nonfinite clauses could be interpreted in the same way.
 



\subsubsection{Second Example}


Consider the sentence `How the prisoner escaped is a mystery' or `The mystery is how the prisoner escaped'. It is easy to see that the clause `the prisoner escaped' expresses an atomic proposition of type I. Assume that its content is $A$ that is either empty or contains a single escaping process (action). The word `How' in this sentence is an adverb denoting a set $B$ of methods, ways, manners, or approaches to complete some actions. Then `How the prisoner escaped' denotes 
\begin{equation}
B_1= \{ b\in B: a\in A\ (a \cap b \neq \varnothing\}
\end{equation}
where $B_1$ is also either empty or comprised of a single composite observation that represents the method of the escaping action contained in $A$. Assume that the word `is' denotes the equivalent relation $\equiv$, the word `a' denotes an existential quantifier, and `mystery' denotes the property $C$. Then, the whole sentence denotes:
\begin{equation}
E=f(B_1(x_1), f_{\exists} (C(x_2), \equiv (x_1,x_2)))
\end{equation}
where $f$ is a basic operation with exact match.

\subsubsection{Third Example}

Consider the sentence: `I know what you want'. Suppose that the context tells that `I' denotes the person $a$, `you' denotes the person $b$. Assume that `know' denotes a binary relation $R$, and `want' denotes a binary relation $Q$. Suppose that the context operations (time, space, world, etc) restrict the relation $Q$ to $Q_1$, and $R$ to $R_1$. Then the phrase `you want' denotes 
\begin{equation}
Q_2  = \{ (x,y)\in Q_1: x \cap b \neq \varnothing\}
\end{equation}
The word `what' is a pronoun in the sentence, and the clause `what you want' denotes the set of things that $b$ want:
\begin{equation}
Q'= \{ y : (x,y)\in Q_2\}
\end{equation}
Then, `know what you want' denotes
\begin{equation}
R_2 = \{ (x,y)\in R_1: y =q \wedge q\in Q'\}
\end{equation}
and the whole sentence denotes
\begin{equation}
R_3 = \{ (x,y)\in R_2: x \cap a \neq \varnothing\}
\end{equation}


\subsection{Multiple Clauses II: Reporting Verbs}

All former examples considers only composite denotations of sentences (see Definition \ref{IP}); nonetheless, sentences and normal phrases could also have non-composite denotations, which are commonly used in quotations.

For instance, `Marry said, `Tom went to school yesterday.''. The clause `Tom went to school yesterday' describes an event; however, common sense tells that what Marry said was not the event described by the clause, but the speech referred by this clause. In other words, when interpreting the whole sentence, the clause `Tom went to school yesterday' should denote a speech said by Marry. A speech is a concrete symbol string, not a meaning composed by the meaning of words in some clause. This becomes more obvious by the use of quotation marks. A clause used in this way is often called a {\bf direct speech} or {\bf quoted speech} \cite{Downing2015}. Generally speaking, a direct speech is a concrete symbol string used to denote another concrete symbol string such that, they belong to the same abstract symbol string. 

\begin{assumption}
Suppose that $ \mathfrak{M}$ is the underlying cognitive model of the interpretation for a concrete symbol string $u$, and $u$ is a direct speech. If $u$ belongs to an abstract symbol string $A\in \mathfrak{M}$, then $u$ denotes $A$ or a concrete symbol string $v$ such that $v \in A$. Such a denotation is a non-composite denotation of $u$. 
\end{assumption}


The difference between composite and non-composite denotations is obvious. Suppose that yesterday is May 1st. Then, `Tom went to school yesterday' and `Tom went to school on May 1st' describe the same event, i.e., they have the same composite denotation. In contrast, `Marry said, `Tom went to school yesterday.'' and `Marry said, `Tom went to school on May 1st.'' have different meaning, because the speech that Marry said was different.


Quotations marks could also be used on normal phrases to indicates the usage of non-composite denotations. For example, in the sentence ``The person' is a noun phrase', the concrete symbol string ``the person'' denotes itself or the abstract symbol string it belongs to, not a person in some world. The quotation marks, however, are often omitted in practical usages without ambiguity. For example, in the sentence `pineapple is a noun', the word `pineapple' denotes a symbol string, not a class of fruits, because only a symbol string could be a noun. 
  
Verbs that could report direct speeches are called {\bf reporting verbs}, such as `say', `tell', `ask', `declare', `remark', `reply', `think', `write', and `hear'. Reporting verbs could also have {\bf indirect speeches} \cite{Downing2015}; for example, `Tom asked whether Marry understood what he meant', `Tom told us that we should go fast', `Tom said that the man had come at six'. There is often a clear syntactic difference between direct speeches and indirect speeches in English \cite{Downing2015}; however, how to interpret indirect speeches? There are two cases.

In the first case, sentences with indirect speeches could be easily converted to sentences with direct speeches, and the conversion does not change the meaning of the sentence \cite{Biber1999, Downing2015}. For example, `Tom said that the man had come at six' often has the same meaning as `Tom said, `The man came at six.'' In this case, the direct speech after the conversion has the same meaning as the indirect speech. 

When people indirectly report what someone has spoken or thought, however, they usually do not exactly report the speech being spoken or thought by him. Instead, they just report the ideas of the speech. This forms the second case of the usage of indirect speeches. For example, suppose that the person Tom has another name Jack, and Marry said, `Tom went to school yesterday.' Then, people could indirectly but correctly report that, Marry said Jack had gone to school yesterday. In this example, the two clauses `Tom went to school yesterday' and `Jack had gone to school yesterday' have the same denotation (describe the same event), but are not the same speech (They belong to different abstract symbol string). In fact, when someone says `Marry said that Jack had gone to school yesterday', he usually means Marry said some speech whose meaning (denotation) and truth are the same as the clause `Jack had gone to school yesterday' under the interpretation. Generally speaking, the following assumption holds in practice.


\begin{assumption}
When a sentence $u$ is an indirect speech, it denotes another sentence $v$ such that, under the interpretation for ${u,v}$, $u,v$ have the same denotation, and the propositions expressed by $u,v$ have the same truth value (They do not necessarily express the same proposition). 
\end{assumption}


Some reporting verbs could connect a normal phrase or a subordinate clause that is neither a direct speech nor an indirect speech; for example, `I heard a rumor', `I heard he died', `I heard what he said'. In this case, we should use composite denotations of the phrase to interpret the whole sentence. 

\subsection{Multiple Clauses III: Propositional Verbs}


Reporting verbs belong to a more general category of words: {\bf propositional verbs} ({\bf propositional attitude reports}). Propositional verbs are verbs that could connect subordinate clauses and describe actions of human or human-like agents. Besides reporting verbs, typical propositional verbs include see, know, believe, think, consider, deny, memorize, imagine, dream, feel, enjoy, hate, want, hope, expect, intend, desire, etc \cite{proatti}. 
 
 
Actions of human-like agents are generally classified into body actions  and mental activities, both of them could be described by propositional verbs. Body actions include one's speaking actions, facial expressions, and brain activities observed by scientific apparatus; mental activities are activities of one's mind, usually observed by his self-consciousness. Although body actions and mental activities are closely related, the description of them should be treated as different meaning of propositional verbs; otherwise, confusions or ambiguities would occur. Consider the sentence `Tom saw Marry yesterday'. This sentence could mean that Tom stood or sat in a special position, towards a special angle where Marry could be observed by his eyes; it could also mean that Tom had perceived some images by his eyes, then realized that he was seeing the girl Marry (recognized these images as images of Marry). Although it is common that Tom's special body action of seeing would result in his mental perception and recognition, it is not necessarily true: Tom could have extremely concentrated on something else, and did not realize that he had seen the girl called Marry, even though his body had the action of seeing the girl; or Tom could have just saw some deceptive images of Marry and made a wrong inference, but in fact Marry had never been there. 

Suppose that we only consider actual observations of the real world. As analyzed before, body actions and mental activities are observed by totally different kind of observers: sensation and self-consciousness. This leads to a crucial difference between them: Body actions could be observed by many other actual observers, and hence could be directly verified or refuted by their observations; however, the only actual observer of a person's mental activities is his self-consciousness thus far, because no scientific apparatus exists to directly observe one's mental activities. Therefore, when a proposition is about one's body actions, the verification and refutation are just like other events; when a proposition is about one's pure mental activities, only observations obtained by his self-consciousness could be used for direct verification or refutation. 



In a cognitive model, a mental activity of an agent $A$ is commonly represented by a pair $(P,P_d)$ where $P$ is a mental process of $A$ and $P_d$ is called the  {\bf product} of $P$. If $A$ is a real person, each mental activity of $A$ could only be directly and actually observed by $A$'s self-consciousness, and hence every actual observation in $P$ has $A$'s self-consciousness as the observer label. Different propositional verbs often talk about different kinds of mental activities, which are distinguished by one's resolution power of self-consciousness. In other words, a person's self-consciousness distinguishes whether his mental activity is a dream, a belief, a piece of knowledge, an intention, a feeling, or of another kind. Moreover, the contents of a mental activity of an agent $A$---what the agent $A$ dreams, believes, knows, desires, intends, etc---are represented by the obtaining results of observations in a mental process $P$ of $A$, obtaining results which constitute the product $P_d$. For example, when an agent $A$ believes that a person lives on the moon, this believing action is a mental activity of $A$, and is represented by a mental process $P$, where the product $P_d$ is the belief `a person lives on the moon'. The product of a mental activity could be very complicated; for instance, people could imagine a whole world during a single activity of imagination.   



If the product of a mental process is described by a few sentences, the product is the meaning of these sentences. A sentence has three kinds of meaning (denotations, senses and explanations), and it depends on the context that which kind of meaning is the product of the mental process. There is an example: consider these two sentences `Tom believe $u_1=u_1$' and `Tom believe $u_1=u_2$', where $u_1,u_2$ are two symbols having the same denotation under the interpretation $\mathcal{I}$. As analyzed before, the two clauses $u_1=u_1$ and $u_1=u_2$ have the same denotation and the same sense, but have different explanation under $\mathcal{I}$. If the denotation or the sense is the product of Tom's believing activity, then these two sentences describe the same mental activity of Tom, and hence have the same meaning. Nonetheless, people generally think these two sentences have different meaning, which could be explained if the explanations of $u_1=u_1$ and $u_1=u_2$ are used to be the product of the believing activity of Tom.  

Suppose that $\kappa$ is a propositional verb and $u$ is a sentence. Since $\kappa$ is the head of the phrase $\kappa u$, the denotation of $\kappa u$ is commonly composed by the denotation of $\kappa$ and the meaning of $u$. There are three kinds of meaning of $u$: denotation, sense and explanation; so we have to make a decision on which kind is used to obtain the denotation of $\kappa u$. In other words, our choices are (see Definition \ref{IP}):

\begin{enumerate}
\item  a non-composite denotation of $u$. 

\item a composite denotation of $u$.

\item a sense of $u$, i.e., a proposition $\phi$ expressed by $u$.


\item an explanation of $u$.
\end{enumerate}

For simplicity, we always assume in this section that the interpretation $\mathcal{I}$ is effective for $u$ and $\kappa u$. Then, there are three choices in all. The following example shows the difference between those three choices in a formal approach. Consider the sentence: `Tom knows $u_1 = u_2$', where $u_1,u_2$ are two symbols. Let $u$ be an abbreviation of the clause $u_1 = u_2$. Suppose that both $u_1,u_2$ denote $a$ under an interpretation $\mathcal{I}$, and the symbol $=$ denotes the equivalent relation $\equiv$. Then, the denotation of the clause $u$ is the pair $e_u=(a,a)$, which belongs to $\equiv$, and the sense of $u$ is 
\begin{equation}
s_{u}=(f_2, a(x_1), (f_1, a(x_2), \equiv (x_1,x_2)))
\end{equation}
where $f_1,f_2$ are two basic operations. The explanation of $u$ is 
\begin{equation}
r_{u}= (r_{u_1}, r_{ = u_2}, (u, s_{u}))
\end{equation}
where $r_{u_1}$ is $(u_1,a)$, and $r_{=u_2}$ is $((=,\equiv), (u_2,a), (=u_2, (f_1, a(x_2), \equiv (x_1,x_2))))$.  Suppose that `Tom' denotes the person $b$. Then, what the denotation of the word `know'? There are three cases:
\begin{enumerate}
\item The denotation of the clause $u$ is used to interpret the whole sentence. Then, the word `know' denotes a binary relation $R_1$ such that $(c_1,c_2)\in R_1$ implies $c_2$ is a denotation of some phrases; and the denotation of the whole sentence is 
\begin{equation}
\{(c_1,c_2)\in R_1: c_1\cap b \neq \varnothing \wedge c_2 = e_u \}
\end{equation}

\item The sense of the clause $u$ is used to interpret the whole sentence. Then the word `know' denotes a binary relation $R_2$ such that $(c_1,c_2)\in R_2$ implies $c_2$ is a sense of some phrases; and the denotation of the whole sentence is 
\begin{equation}
\{(c_1,c_2)\in R_2: c_1\cap b \neq \varnothing \wedge c_2 = s_{u}\}
\end{equation}

\item The explanation of the clause $u_1=u_2$ is used to interpret the whole sentence. Then, the word `know' denotes a binary relation $R_3$ such that $(c_1,c_2)\in R_3$ implies $c_2$ is an explanation of some phrases; and the denotation of the whole sentence is 
\begin{equation}
\{(c_1,c_2)\in R_3: c_1\cap b \neq \varnothing \wedge c_2 = r_{u}\}
\end{equation}
\end{enumerate}

In every case, the first element $c_1$ of a sequence $ (c_1,c_2)\in R_i$ $(1 \leq i \leq 3)$ is someone's mental states of knowing something, and $c_2$ is the product of $c_1$. Different kind of products $c_2$ indicates knowing different kind of things, and $R_1, R_2, R_3$ are therefore different denotations of the word `know', which implies `know' is a multi-meaning word in general. When one speaks out a sentence like `$a$ knows $u$', he should be clear about whether the agent $a$ knows the denotation, knows the sense, or knows the explanation of $u$; otherwise, there would be some confusions or ambiguities.

What is the conceptual difference between relations $R_1, R_2, R_3$? In other words, what is the conceptual difference between knowing a denotation of a sentence $u$, knowing a sense of $u$, and knowing an explanation of $u$? Suppose that $u$ describes an event $P$ under the interpretation $\mathcal{I}$. Then, generally speaking, when one knows the denotation of $u$, he only knows the event $P$; when one knows the sense of $u$, he further knows how the event $P$ is extracted by a composite procedure from other objects, processes, and so on; when one knows the explanation of $u$, he further knows how each sub-phrase of $u$ is interpreted under $\mathcal{I}$. 

Sometimes, it is easy to make the choice, since other choices will not make sense. For example, the sentence `I know what Tom said' means I know something said by Tom; so the composite denotation of the clause `what Tom said' is used to interpret the whole sentence. In contrast, when interpreting the sentence `He knows Shakespeare was the writer of Hamlet', the sense or the explanation of the clause `Shakespeare was the writer of Hamlet' has to be used, because in most contexts the denotation of `Shakespeare was the writer of Hamlet' is the same as the denotation of `Shakespeare was Shakespeare'. When interpreting the sentence `He knows Kelley is Marry', however, only the explanation of the clause `Kelley is Marry' could be used, because `Kelley is Marry' and `Kelley is Kelley' have the same denotation and the same sense under an interpretation where `Kelley' and `Marry' denote the same person. 

In many cases, however, it is not so easy to make the decision, and additional context information is needed. Consider the sentence `Tom knew that Kelley had come yesterday'. Suppose that `Kelley' denotes the person $b$ under the interpretation $\mathcal{I}$. Did Tom know that `Kelley' denotes $b$? Without special context information, it is possible that Tom only knew the event that the person $b$ had come yesterday, but did not know the word `Kelley' denotes $b$. In this case, it is incorrect to use the explanation of the clause `Kelley had come yesterday' to interpret the whole sentence. Similarly, the sentence `Tom knew that the girl with a red coat had come yesterday' does not imply that Tom knew the sense of the phrase `the girl with a red coat': It is possible that the speaker of the sentence saw the girl wearing a red coat, but Tom was not there and had never seen or imagined this before. In this cases, only the denotation of the clause could be used to interpret the whole sentences. 







The reason for such a complication is that, when someone says `Tom knew that Kelley had come yesterday', the speaker and the listener use an interpretation $\mathcal{I}$ to interpret this sentence; however, $\mathcal{I}$ is not necessarily known by the person Tom, and hence Tom does not necessarily know the sense or the explanation of the clause, unless a special context shows he does. 

Sentences with other propositional verbs could be analyzed similarly. The above discussion only tells how to decide the denotation of a sentence with a propositional verb; however, how to decide its truth? When a sentence with a propositional verb describes some agents' body actions, the truth of the sentence could be reduced to atomic propositions of Type I, because body actions of some agents are just like other events. Therefore, it only needs to consider the case when a sentence with a propositional verb describing some agents' pure mental activities. In this case, the propositional verb denotes a relation that represents pure mental activities of some agents, and such a relation is called an {\bf $M$-relation}. 

\begin{definition}[$M$-relation]
An $M$-relation is a binary relation $R$ such that, for each $(c_1,c_2)\in R$, $c_1$ is a mental process of some agent, $c_2$ is the product of $c_1$. 
\end{definition}
 
Propositions with $M$-relations are called $M$-propositions, which could be simple or complicated. Nonetheless, we only need to define the simplest kind of $M$-propositions and decide their truth; more complicated $M$-propositions could be reduced to them, because propositions are recursively constructed in general.


\begin{definition}[Atomic $M$-Proposition]\label{APIII}
$\phi$ is an atomic $M$-proposition if and only if (1) the content of $\phi$ is either empty or comprised of a single sequence belonging to an $M$-relation, and (2) its truth is decided by Assumption \ref{TAPIII}. 
\end{definition}


The decision of the truth of an atomic $M$-proposition $\phi$ is a little complicated. Suppose the content of $\phi$ consists of a sequence $(c_1,c_2) \in R$ where $R$ is an $M$-relation. The product $c_2$ of the mental process $c_1$ is often a denotation, a sense, or an explanation of some clause; so $c_2$ usually has a truth value under the interpretation. Then, for $\phi$ to be true, does it require $c_2$ to be true? Common sense tells that this depends on whether $c_2$ is supposed to be knowledge. For instance, `Tom believes that he could fly' is true does not require the clause`he could fly' to be true; in contrast, `Tom knows that he could fly' is true requires that `he could fly' is true. Most propositional verbs are in the same category of `believe', such as `imagine', `desire' and `expect''. A few propositional verbs are in the category of `know', such as `learn', `see' and `notice'. 

For $\phi$ to be true, however, it always requires that the mental process $c_1$ {\bf actually happen} in the world being considered. For example, `Tom believes that he could fly' is true requires that Tom actually has such a belief in the world he live; `Tom imagines that he could fly' is true requires Tom actually has such an imagination; `Tom knows that he could fly' is true requires Tom actually has such a knowledge. This implies that $c_1$ needs a verification or refutation, just like the content of an atomic proposition of type I.

\begin{assumption}[Truth of Atomic $M$-Proposition] \label{TAPIII}
Let $ \mathcal{I} = ( \mathcal{I}_c,  \mathbb{C})$ be an effective interpretation whose underlying cognitive model is $\mathfrak{M}$.  Suppose that the sentence $u$ expresses the proposition $\phi$ under $ \mathcal{I}$, where $\phi$ is an atomic $M$-proposition whose content consists of a sequence $(c_1,c_2) \in R$ such that, $R$ is an $M$-relation and $c_2$ is the product of $c_1$. Then
\begin{itemize}
\item If the product $c_2$ is not supposed to be knowledge, then (1) $\phi$ is true on $\mathfrak{M}$ if $c_1$ is directly verified on $\mathfrak{M}$; (2) $\phi$ is false on $\mathfrak{M}$ if $c_1$ is directly refuted on $\mathfrak{M}$; (3) $\phi$ is undecided on $\mathfrak{M}$ if $c_1$ is neither directly verified nor directly refuted on $\mathfrak{M}$. 

\item If the product $c_2$ is supposed to be knowledge, then (1) $\phi$ is true on $\mathfrak{M}$ if $c_1$ is directly verified on $\mathfrak{M}$, and $c_2$ is true on $\mathfrak{M}$ or with respect to $ \mathcal{I} $; (2) $\phi$ is false if $c_1$ is directly refuted on $\mathfrak{M}$, or $c_2$ is false on $\mathfrak{M}$ or with respect to $ \mathcal{I} $; (3) $\phi$ is undecided in other cases. (When $c_2$ is a denotation or a sense of a clause, it has truth on a cognitive model; when $c_2$ is an explanation of a clause, it has truth with respect to an effective interpretation (see \ref{TAOF} and \ref{TDAE}.) 
\end{itemize}

\end{assumption}


\begin{assumption}
Suppose that $\phi \in \mathfrak{M}$ is an atomic $M$-proposition whose content is empty. If $\phi$ is not assigned to the truth value $V$ by Assumption \ref{VOP}, then $\phi$ is assigned to the truth value $F$ on $\mathfrak{M}$. 
\end{assumption}

A practical difficulty could arise when a real person wants to verify or refute a mental process that is not his own (i.e., to check if other people actually has some mental activity or not). The reason has been mentioned before: Mental activities of a real person could only be directly known by the self-consciousness of his own; other people could not directly verify or refute observations in such a process; instead, they could only provide indirect verification or refutation. In practice, people has summarized some general rules and methods to indirectly check if other person actually has some mental activity or not, usually by reasoning from his body actions; however, those rules or methods are usually fallible, with more or less uncertainties.  Reasoning, uncertainties, and indirect verification or refutation are topics beyond the scope of this article. 

\subsection{Proposition with Connectives}



In last three sections, when a multiple sentence $u$ expresses a proposition $\phi$, the operation $f_{\xi}$ denoted by the main clause linker $\xi$ of $u$ (or provided by conventions) is not the main operation of $\phi$ (see Section \ref{MCI}). The next two sections discuss the opposite case: $f_{\xi}$ is the main operation of $\phi$. In this case, $f_{\xi}$ is called a {\bf connective}. Connectives could be denoted by four kinds of function phrases:

\begin{enumerate}  
\item {\bf Phrases expressing negation}: `no', `not', `never', `hardly', `seldom', `scarcely', most words with a prefix `un-' or `dis-', etc;  

\item {\bf Coordinate connectors}: function phrases that connect coordinate clauses, e.g., `and', `or', `but', `both..., and...', `neither..., nor...', `either..., or...', etc \cite{Quirk1985};

\item {\bf Adverbial clause linkers}: function phrases that connect adverbial clauses, e.g., `if', `then', `unless', `assuming', `because', `since', `hence', `thus' `when', `after', `as', `where', `so that', `that', `though', `even if', `while', `whether', etc \cite{Quirk1985};

\item {\bf Modal phrases}: modal auxiliaries verbs and some other phrases expressing modality, e.g., `can', `could', `may', `might', `must', `shall', `should', `will', `would', `be possible', `be probable', `be necessary', `dare', `need', `ought to', `have to', `be able to', etc \cite{Downing2015}.

\end{enumerate}



This section focuses on connectives denoted by the first three kinds of phrases; the next section discusses modality: connectives denoted by modal phrases. There are two things worth noting before the discussion of connectives. The first thing is, just like quantifiers, connectives are operations in cognitive models, not syntactic symbols in a formal language: They are pure semantic elements. The second thing is that function phrases that denote connectives are often multi-meaning under $ \mathcal{I}_c$. For example, the word `and' could mean the logic connective $\wedge$, or have meanings similar to words like `therefore', `then', `yet', `in contrast', `but', `will', `also' and `similarly' \cite{Quirk1985}, which often denote different connectives. Moreover, a word like `and' does not always denote a connective. For example, the sentence `Tom and Marry are friends' does not mean Tom is a friend and Marry is a friends. To decide the connective denoted by a function phrase in a context is an important but difficult problem, belonging to the general problem of word sense disambiguation, which is not a topic of this article. The simplest connectives are truth-functional connectives.


\begin{definition}[Truth-functional Connectives]
Suppose that $\alpha, \beta$ are two propositions. A connective $f$ is a {\bf truth-functional connective} if and only if the truth of the propositions $(f, \alpha)$ (unary) or $(f, \alpha, \beta)$ (binary) are completely decided by the truth of $\alpha$ and $\beta$. In other words, their truth could be decided by truth tables or truth functions. 
\end{definition}

Note that $(f, \alpha)$ and $(f, \alpha, \beta)$ are simplified notations of propositions. A symbol $\odot$ is often introduced to denote the connective $f_{\odot}$, and propositions are often written as sentences in a formal language: $ \odot \ \alpha$ and $\alpha \odot \beta$. Five truth-functional connectives have been extensively studied in classic logic, denoted by symbols $\neg$, $\rightarrow$, $\wedge$, $\vee$, $\leftrightarrow$ respectively \cite{Enderton2001}; however, unlike in classic logic, there are four truth values defined in this article, so the truth tables are a little more complicated. Suppose that $\alpha, \beta$ are propositions and $\odot \in \{\rightarrow,\wedge,\vee,\leftrightarrow\}$, then the general rule are 

\begin{itemize}
\item $\neg \alpha$ is assigned to $V$ if and only if $\alpha$ is assigned to $V$.

\item $\alpha \odot \beta$ is assigned to $V$ if and only if $\alpha$ or $\beta$ is assigned to $V$.

\item Depending on different context, other three truth values $\{T,F,U\}$ follow Kleene's or Lukasiewicz's Three-valued Logic \cite{Gabbay2007}. 
\end{itemize}


There are other truth-functional connectives denoted by words in natural language. For example, the conjunction `or' could mean that, one of those two propositions is true but never both are true. Then a new truth-functional connective $\curlyvee$ could be defined to represent such a meaning: 
\begin{equation}
\alpha \curlyvee \beta =_{df} (\alpha \vee \beta) \wedge \neg (\alpha \wedge \beta) 
\end{equation}

The implication defined above is often called {\bf material implication}: when the premise is false or the conclusion is true, the whole proposition is true; when the premise is true and the conclusion is false, the whole proposition is false. According to different intuitions, however, there are at least two different kinds of material implication: one is defined by Kleene's Three-valued Logic, and the other is defined by Lukasiewicz's \cite{Gabbay2007}. 

Implication is often denoted by phrases like `if...then', `since' and `because'. Besides several kinds of material implication, there are other kinds of implication, which are often not truth-functional connectives. For example, for the sentence `The water is boiling because it is heated by a fire' to be true, it requires a causal relation between the two events described by `The water is boiling' and `it is heated by a fire'. This requirement makes the implication denoted by the word `because' not truth-functional. 


There are many other non-truth-functional connectives. Consider the sentence: `When he came back, I was doing my homework.' In this sentence, the conjunction `when' denotes a connective: a binary operation $h(a,b)$ such that the arguments $a,b$ of $h$ should be two events happening at the same moment. In other words, it requires that $a,b$ satisfy a special relation $R_h$ such that, for each sequence $(a,b)\in R_h$, $a,b$ are states happening at the same time moments ($t_a=t_b$ for every $(a,b)\in R_h$). The relation $R_h$ is called the relation associated with $h$. 

Suppose that the clause `he came back' expresses a proposition $\alpha$ whose content is $e_{\alpha}$, and suppose that `I was doing my homework' expresses a proposition $\beta$ whose content is  $e_{\beta}$. Then the whole sentence expresses the proposition $\gamma=(h, \alpha, \beta)$ whose content is $e_{\gamma}$, where
\begin{equation}
e_{\gamma} = h(e_{\alpha}, e_{\beta}) = \{(x,y)\in R_h : x=e_{\alpha}, y=e_{\beta}\}
\end{equation}
In other words, the content of $\gamma$ is just the sequence formed by $e_{\alpha}$ and $e_{\alpha}$ only if it belongs to the relation $R_h$. This example could be generalized to almost all non-truth-functional connectives, except those ones denoted by modal phrases. 

\begin{assumption}
Suppose that  $h$ is a connective, probably denoted by a modal phrase. Assume that $h$ is the main operation of proposition $\phi$. Then, $h$ is associated with a relation $R_h$, and the content of $ \phi$ is either empty or a single element belonging to $R_h$. 
\end{assumption}

The relation associated with a connective is crucial to this connective. In fact, a sentence with a phrase denoting a connective could often be turned into a sentence with the same denotation and the same truth but without the phrase. For example, suppose that $u,v$ are two sentences, then `If $u$, then $v$' is equivalent to `$u$ implies $v$', `$u$ causes $v$' or `$u$ explains $v$' in most cases. 

A sentence has three kinds of meaning (denotation, sense and explanation), and a connective could operate not only on denotations but also on other kinds of meaning of sentences. In other words, the relation associated with a connective could be a relation of denotations, a relation of propositions, a relation of explanations, or a mixture. When the associated relation is a time relation or a causal relation, it is a relation between events of worlds, which are commonly denotations of sentences; in contrast, when the associated relation is a relation of implication, it is often a relation between propositions or explanations.

Then, how to decide the truth of a proposition whose main operation is a non-truth-functional connective? Different connective would have different rules to decide the truth, related to the associated relation, the content of the proposition, and the truth of the clauses (the sub-propositions). Since there are so many different kinds of connectives, the detailed study could not be covered in this article; only a necessary condition is summarized as follows.  

\begin{assumption}
Suppose that $\phi$ is a proposition whose main operation is a connective, probably denoted by a modal phrase. Then $\phi$ is true implies that the content of $\phi$ is nonempty. 
\end{assumption}

Finally, a few words about discourses could be said after the discussion of connectives. A discourse is comprised of several sentences in a single topic, sentences which are supposed to have some relations between them. Some function phrases often exists to link those sentences in a discourse, which are called {\bf sentential linkers}; or if there are not, some connectives would be provided by conventions to connect the propositions expressed by those sentences. From this point of view, a discourse could be regarded as a huge sentence with a structure made by those connectives, denoted by sentential linkers or provided by conventions.

\subsection{Modal Connectives}\label{MCMC}


This section discusses the interpretation of modal phrases. Sentences with a modal phrase often express a kind of {\bf modality}, which is a special idea like possibility, probability, necessity, prediction, obligation, permission, volition, ability, or natural tendency \cite{Downing2015}. Modality is classified into three main categories: epistemic modality, deontic modality and dynamic modality, where the last category is considered less central \cite{Downing2015}. 

Modality is commonly related to things in more than one world. The standard approach to analyze modality is based on Possible World Semantics \cite{Blackburn2001, Blackburn2006, Gabbay2007-7}. Possible World Semantics believes that every kind of modality expresses a kind of possibility or necessity that could be analyzed on models comprised of a set of possible worlds and some access relations: It is possible (necessary) that a proposition $\phi$ is true on a word $w$ if and only if there exists a world (for every word) $w'$ reached by $w$ where $\phi$ is true. For instance, the sentence `It is necessary that Socrates is a philosopher' is true on the real world if and only if the proposition expressed by `Socrates is not a philosopher' is true on every world $w'$ reached by the real world.  


Possible World Semantics was a great achievement to understand modality; however, there is a fundamental problem left unsolved: In the above example, does the word `Socrates' denote the same person in every world, or does it denote different person in different world?  If `Socrates' denotes different person in different world, then the clause `Socrates is a philosopher' must express different proposition in different world, because the sentence talks about different object in different world. If `Socrates' denotes the same person $a$ in every world, then how could $a$, which is often a physical object in the real world, cross to a different world and probably live a different life? Therefore, it appears that either approach has its problem, and the second one---the word `Socrates' denotes the same person in every world--- requires far stronger justification than the first one. From this consideration, the new semantic theory chooses the second approach---the word `Socrates' denotes different person in different world, and modifies the Possible World Semantic to solve its problems.

If the word `Socrates' denotes different person in different world, then it is a multi-meaning word; more important, the clause `Socrates is a philosopher' is a multi-meaning expression: It expresses different proposition in different world. Therefore, to assert `It is necessary that Socrates is a philosopher', it is not to assert that the same proposition expressed by the clause `Socrates is a philosopher' is true on every world reached by the real world, but to assert that all propositions expressed by this clause in the context are true. This idea is formally developed as follows. 

Suppose that $\lambda u$ represents a natural language sentence where $\lambda$ is a modal phrase and $u$ is a clause. A modal phrase, as exemplified before, is a phrase denoting some unary operation called {\bf modal operation} or {\bf modal connective} that is supposed to formally characterize a kind of modality expressed by the phrase. Most modal phrases are multi-meaning function phrases, probably denoting different modal connective in different context to talk about different kind of modality. Some modal phrases could be content phrases sometimes, such as `need' and `dare'. Therefore, it is important to make clear the denotation of a modal phrase in a given context before interpreting the whole sentence. 

Suppose that $\mathcal{I}=(\mathcal{I}_c, \mathbb{C})$ is an effective interpretation for $\lambda u$. Then, the modal phrase $\lambda$ denotes a unique modal connective $f_{\lambda}$. The essential point to understand modality in this new semantic theory is: Although $\mathcal{I}$ is effective for the sentence $\lambda u$, it is often not effective for the clause $u$. In other words, since $\lambda$ expresses a kind of modality, the clause $u$ is often a multi-meaning phrase under $\mathcal{I}$, even if $\mathcal{I}$ is effective for the sentence $\lambda u$. The context $\mathbb{C}$ does not select a single denotation for each word or idiom contained in $u$, but determines a range of worlds, a range of time, a range of space, a range of observers, and a range of other elements being considered,  and therefore decides the set of denotations for each sub-phrase of $u$. 


The clause $u$ is a multi-meaning phrase under $\mathcal{I}$ implies that $u$ has a set $\mathbb{A}_u$ of denotations, a set $\mathbb{B}_u$ of senses, and a set $\mathbb{E}_u$ of explanations under $\mathcal{I}$. The operation $f_{\lambda}$ therefore operates on $\mathbb{A}_u$, $\mathbb{B}_u$, or $\mathbb{E}_u$. In other words, $\mathbb{A}_u$, $\mathbb{B}_u$, or $\mathbb{E}_u$ is the argument of the unary operation $f_{\lambda}$, and the denotation of the sentence $\lambda u$ would be $f_{\lambda} (\mathbb{A}_u)$, $f_{\lambda} (\mathbb{B}_u)$, or $f_{\lambda} (\mathbb{E}_u)$. Like other connectives, $f_{\lambda}$ is associated with a (unary) relation $P_{f_{\lambda}}$; to obtain the denotation of $\lambda u$, $f_{\lambda}$ will decide whether $\mathbb{A}_u$, $\mathbb{B}_u$, or $\mathbb{E}_u$ satisfies $P_{f_{\lambda}}$ or not. The above discussion generates the following definition. 

\begin{definition}[Interpretation of Modality]
Suppose that $\lambda u$ is a natural language sentence where $\lambda$ is a modal phrase and $u$ is a clause. Let $\mathcal{I}= (\mathcal{I}_c, \mathbb{C})$ be an effective interpretation for $\lambda u$. Let $f_{\lambda}$ be the denotation of $\lambda$ under $\mathcal{I}$, which is a modal connective associated with a property $P_{f_{\lambda}}$. Suppose that, under $\mathcal{I}$, $\mathbb{A}_u$ is the set of all denotations of $u$,  $\mathbb{B}_u$ is the set of all senses of $u$, and $\mathbb{E}_u$ is the set of all explanations of $u$. Then, depending on the context $\mathbb{C}$, 


\begin{itemize}
\item The {\bf denotation} of the sentence $\lambda u$ is $f_{\lambda} (\mathbb{A}_u)$, $f_{\lambda} (\mathbb{B}_u)$, or $f_{\lambda} (\mathbb{E}_u)$, where $f_{\lambda} (\mathbb{X}_u)= \{a\in P_{f_{\lambda}}: a=\mathbb{X}_u\}$  ($\mathbb{X}_u$ is $\mathbb{A}_u$, $\mathbb{B}_u$, or $\mathbb{E}_u$).

\item The {\bf sense} of $\lambda u$ is $(f_{\lambda}, s_{\lambda}, \mathbb{B}_u)$, $(f_{\lambda}, s_{\lambda}, (\mathbb{B}_u))$, or $(f_{\lambda}, s_{\lambda}, \mathbb{E}_u)$, where $s_{\lambda}$ is the sense of $\lambda$, $(f_{\lambda}, s_{\lambda}, \mathbb{B}_u)$ implies the denotation $f_{\lambda} (\mathbb{A}_u)$, $(f_{\lambda}, s_{\lambda}, (\mathbb{B}_u))$ implies $f_{\lambda} (\mathbb{B}_u)$, and $(f_{\lambda}, s_{\lambda}, \mathbb{E}_u)$ implies $f_{\lambda} (\mathbb{E}_u)$.

(If $f_{\lambda}=s_{\lambda}$, they are simply written as $(f_{\lambda},\mathbb{B}_u)$, $(f_{\lambda},  (\mathbb{B}_u))$, and $(f_{\lambda}, \mathbb{E}_u)$.)
 
\item The {\bf explanation} of $\lambda u$ is $(r_{\lambda}, (u, \mathbb{E}_u), (\lambda u, s_{\lambda u}))$, where $r_{\lambda}$ is the explanation of $\lambda$, and $s_{\lambda u}$ is the sense of $\lambda u$. 
\end{itemize}


\end{definition}


If $\mathcal{I}$ is not an effective interpretation for $\lambda u$, things would be more complicated: There would be several sets of denotations, senses, explanations of $u$ to be argument of a modal connective $f \in \mathcal{I}(\lambda)$.

\vspace{6pt}

Then, how to decide the truth of a proposition whose main operation is a modal connective? There are many different kinds of modality, and hence many different kinds of modal connectives. Each kind of modal connective would have its own rules to decide the truth, and most of them are related to the characterization of uncertainty. Therefore, a complete study could not be covered in this article; instead, we only discuss the two simplest modal connectives---possibility and necessity---as an illustration of the general ideas. 

Suppose the sentence `It is necessary (possible) that $u$' is interpreted under an effective interpretation $\mathcal{I}= (\mathcal{I}_c, \mathbb{C})$, where $u$ is a sentence. The common understanding of necessity or possibility, and the approach adopted in this article together generate the following truth definition: `It is necessary (possible) that $u$' is true if and only if every (some) meaning of $u$ is true. Since there are three kinds of meaning, there are three cases: every (some) denotation of $u$ is true, every (some) sense of $u$ is true, or every (some) explanation of $u$ is true. Moreover, denotations and senses have truth on cognitive models, but explanations have truth under effective interpretations. Therefore, the formal definition is a little complicated.

\begin{assumption}[Truth of Possibility/Necessity]
Suppose that $u$ is a sentence, and the sentence `It is necessary (possible) that $u$' is interpreted under an effective interpretation $\mathcal{I}=(\mathcal{I}_c, \mathbb{C})$. Let $f_{\square}$ be the {\bf necessity modal connective}, denoted by the idiom `It is necessary that'. Let $f_{\diamondsuit}$ be the {\bf possibility modal connective}, denoted by the idiom `It is possible that'. Let $\mathfrak{M}$ be the underlying cognitive model of $\mathcal{I}$, and let $[\mathcal{I}]$ be the set of all effective interpretations for $u$ based on $\mathcal{I}$ (see Definition \ref{SEI}). Assume that, the clause $u$ has a set $\mathbb{A}_u$ of denotations,  a set $\mathbb{B}_u$ of senses, and a set $\mathbb{E}_u$ of explanations under $\mathcal{I}$. Then, depending on the context $\mathbb{C}$, the sentence `It is necessary that $u$' expresses a proposition $(f_{\square}, \mathbb{B}_u)$, $(f_{\square}, (\mathbb{B}_u))$, or $(f_{\square}, \mathbb{E}_u)$, and

\begin{itemize}

\item $(f_{\square}, \mathbb{B}_u)$ is true if and only if every element in $\mathbb{A}_u$ is true on $\mathfrak{M}$.

\item $(f_{\square}, (\mathbb{B}_u))$ is true if and only if every element in $\mathbb{B}_u$ is true on $\mathfrak{M}$.

\item $(f_{\square}, \mathbb{E}_u)$ is true if and only if for every $\delta \in \mathbb{E}_u$, $\delta$ is true under the interpretation $\mathcal{I}' \in [\mathcal{I}]$ where $\delta$ is the explanation of $u$ (see Definition \ref{TDAE}).


\end{itemize}

When considering the sentence `It is possible that $u$', we just change each $f_{\square}$ to $f_{\diamondsuit}$, and change each word `every' to `some' in above clauses.

\end{assumption}

A practical difficulty is to decide the context or the set $\mathbb{X}_u$ ($\mathbb{X}_u$ is $\mathbb{A}_u$, $\mathbb{B}_u$, or $\mathbb{E}_u$). This difficulty is consistent with our common sense: The truth of sentences with modal phrases are often vague and controversial, because the context could be easily changed in the conversation and there is often no definite rules to decide the set $\mathbb{X}_u$ in a given context. For example, people might claim that `It is necessary that the sun rises every day'. When one claim this sentence, he probably only considers those worlds where a solar system exists, the earth spins and rotates around the sun; however, another person who heard this sentence might respond that `No, it is not necessary because you could imagine a world where the earth does not spin'. Both claims could be true under the corresponding interpretation, because what the second person says changes the context to consider worlds where the earth does not spin. 

Using the theory presented in this section, it is easy to understand why these two sentences `It is necessary that $u=u$' and `It is necessary that $u=v$' usually have different meaning and truth. The crucial point is still that $u=u$ and $u=v$ are multi-meaning clauses under an interpretation for the whole sentence. In a usual context, the symbol $=$ is interpreted as the equivalent relation, and two concrete symbols belonging to the same abstract symbol are interpreted to the same denotation. In such a context, the proposition expressed by $u=u$ is always true, no matter what the denotation of $u$ is; however, the proposition expressed by $u=v$ could be false, because $u,v$ belong to different abstract symbol and hence could have different denotation. 

\appendix





\bibliography{dissertation}

\begin{thebibliography}{10}

\bibitem{Noun}
{Noun}.
\newblock https://en.wikipedia.org/wiki/Noun.

\bibitem{Agirre2007}
Eneko Agirre and Philip Edmonds.
\newblock {\em Word Sense Disambiguation: Algorithms and Applications}.
\newblock Springer, 2007.

\bibitem{Allen1995}
James Allen.
\newblock {\em Natural Language Understanding, 2nd Edition}.
\newblock The Benjamin/Cummings Publishing Company, 1995.

\bibitem{Aristotle1991}
Aristotle.
\newblock {\em The Complete Works of Aristotle}.
\newblock Princeton University Press, 1991.

\bibitem{Baugh2002}
Albert~C. Baugh and Thomas Cable.
\newblock {\em A History of the English Language, Fifth Edition}.
\newblock Routledge, 2002.

\bibitem{Beltagy2016}
I.~Beltagy, Stephen Roller, Pengxiang Cheng, Katrin Erk, and Raymond~J. Mooney.
\newblock Representing meaning with a combination of logical and distributional
  models.
\newblock {\em Computational Linguistics December 2016, Vol. 42, No. 4:
  763-808.}, 2016.

\bibitem{Biber1999}
Douglas Biber, Stig Johansson, Geoffrey Leech, Susan Conrad, and Edward
  Finegan.
\newblock {\em Longman Grammar of Spoken and Written English}.
\newblock Pearson Education Limited, 1999.

\bibitem{Blackburn2001}
Patrick Blackburn, Maarten de~Rijke, and Yde Venema.
\newblock {\em Modal Logic}.
\newblock Cambridge University Press, 2001.

\bibitem{Blackburn2006}
Patrick Blackburn, Johan van Benthem, and Frank Wolter.
\newblock {\em Handbook of Modal Logic}.
\newblock Elsevier Science, 2006.

\bibitem{Boleda2016}
Gemma Boleda and Aurelie Herbelot.
\newblock Formal distributional semantics: Introduction to the special issue.
\newblock {\em Computational Linguistics December 2016, Vol. 42, No. 4:
  619-635.}, 2016.

\bibitem{Boolos2007}
George~S. Boolos, John~P. Burgess, and Richard~C. Jeffrey.
\newblock {\em Computability and Logic, 5th}.
\newblock Cambridge University Press, 2007.

\bibitem{Church1956}
Alonzo Church.
\newblock {\em Introduction to Mathematical Logic}.
\newblock Princeton University Press, 1956.

\bibitem{Coon2010}
Dennis Coon and John~O. Mitterer.
\newblock {\em Introduction to Psychology, twelfth edition}.
\newblock Wadsworth, Cengage Learning, 2010.

\bibitem{Croft2004}
William Croft and D.~Alan Cruse.
\newblock {\em Cognitive Linguistics}.
\newblock Cambridge University Press, 2004.

\bibitem{Downing2015}
Angela Downing.
\newblock {\em English Grammar, A University Course, 3rd}.
\newblock Routledge, 2015.

\bibitem{Ebbinghaus1984}
H.~D. Ebbinghaus, J.~Flum, and W.~Thomas.
\newblock {\em Mathematical Logic, 2nd}.
\newblock Springer-Verlag, 1984.

\bibitem{Enderton2001}
Herbert~B. Enderton.
\newblock {\em A Mathematical Introduction to Logic, 2nd}.
\newblock Harcourt/Academy Press, 2001.

\bibitem{Evans2006}
Vyvyan Evans and Melanie Green.
\newblock {\em Cognitive Linguistics: An Introduction}.
\newblock Edinburgh University Press, 2006.

\bibitem{Frege1960}
Gottlob Frege.
\newblock {\em Translations from the Philosophical Writings of Gottlob Frege}.
\newblock Basil Blackwell, 1960.

\bibitem{Gabbay2007-7}
Dov~M. Gabbay and John Woods.
\newblock {\em Handbook of the History of Logic, Volume 7}.
\newblock Elsevier, 2007.

\bibitem{Gabbay2007}
Dov~M. Gabbay and John Woods.
\newblock {\em Handbook of the History of Logic, Volume 8}.
\newblock Elsevier, 2007.

\bibitem{Geeraerts2007}
Dirk Geeraerts and Hubert Cuyckens.
\newblock {\em The Oxford Handbook of Cognitive Linguistics}.
\newblock Oxford University Press, 2007.

\bibitem{Hopcroft2007}
John~E. Hopcroft.
\newblock {\em Introduction to Automata Theory, Languages, and Computation}.
\newblock Pearson Education, 2007.

\bibitem{Hrbacek1999}
Karel Hrbacek and Thomas Jech.
\newblock {\em Introduction to Set Theory, 3rd Edition}.
\newblock Marcel Dekker, 1999.

\bibitem{Jurafsky2008}
Dan Jurafsky and James~H. Martin.
\newblock {\em Speech and Language Processing: An Introduction to Natural
  Language Processing}.
\newblock Prentice Hall, 2008.

\bibitem{Kandel2013}
Eric~R. Kandel, James~H. Schwartz, Thomas~M. Jessell, Steven~A. Siegelbaum, and
  A.~J. Hudspeth.
\newblock {\em Principles of Neural Science, fifth edition}.
\newblock McGraw Hill Company, Inc, 2013.

\bibitem{Klammer2009}
Thomas~P. Klammer, Muriel~R. Schulz, and Angela~Della Volpe.
\newblock {\em Analyzing English Grammar, sixth Edition}.
\newblock Pearson, 2009.

\bibitem{Kripke1959}
S.~Kripke.
\newblock A completeness analysis of modal logic.
\newblock {\em Journal of Symbolic Logic, 24:1-14.}, 1959.

\bibitem{Lewis1973}
David Lewis.
\newblock {\em Counterfactuals}.
\newblock Blackwell, 1973.

\bibitem{Lobeck2013}
Anne Lobeck and Kristin Denham.
\newblock {\em Navigating English Grammar: A Guide to Analyzing Real Language}.
\newblock Wiley-Blackwell, 2013.

\bibitem{Manning1999}
Christopher~D. Manning and Hinrich Schutze.
\newblock {\em Foundations of Statistical Natural Language Processing}.
\newblock The MIT Press, 1999.

\bibitem{McELIECE2004}
R.~J. McELIECE.
\newblock {\em The Theory of Information and Coding}.
\newblock Cambridge University Press, 2004.

\bibitem{proatti}
Thomas McKay and Michael Nelson.
\newblock Propositional attitude reports.
\newblock In Edward~N. Zalta, editor, {\em The Stanford Encyclopedia of
  Philosophy}. Metaphysics Research Lab, Stanford University, spring 2014
  edition, 2014.

\bibitem{Montague1974}
Richard Montague.
\newblock The proper treatment of quantification in ordinary english.
\newblock {\em Formal Philosophy, Yale University Press, New Haven, CT, page
  247-270.}, 1974.

\bibitem{Munkres2000}
James~R. Munkres.
\newblock {\em Topology}.
\newblock Prentice Hail, Inc., 2000.

\bibitem{Navigli2009}
Roberto Navigli.
\newblock Word sense disambiguation: a survey.
\newblock {\em ACM Computing Surveys, Vol. 41, No. 2, Article 10}, 2009.

\bibitem{Partee2008}
Barbara~H. Partee.
\newblock {\em Compositionality in Formal Semantics: Selected Papers by Barbara
  H. Partee}.
\newblock John Wiley \& Sons, 2008.

\bibitem{Plato1997}
Plato.
\newblock {\em Plato: Complete Works}.
\newblock Hackett Publishing Company, 1997.

\bibitem{Quirk1985}
Randolph Quirk, Sidney Greenbaum, Geoffrey Leech, and Jan Svartvik.
\newblock {\em A Comprehensive Grammar of the English Language}.
\newblock Longman Group Limited, 1985.

\bibitem{Russell2009}
Bertrand Russell.
\newblock {\em The Basic Writings of Bertrand Russell}.
\newblock Taylor \& Francis e-Library, 2009.

\bibitem{sep-meaning}
Jeff Speaks.
\newblock Theories of meaning.
\newblock In Edward~N. Zalta, editor, {\em The Stanford Encyclopedia of
  Philosophy}. Spring 2016 edition, 2016.

\end{thebibliography}

\bibliographystyle{plain}

\end{document}